\documentclass[10pt,twocolumn,letterpaper]{article}

\usepackage[pagenumbers]{cvpr} 

\usepackage{color}
\usepackage[table]{xcolor}

\definecolor{cvprblue}{rgb}{0.21,0.49,0.74}
\usepackage[pagebackref,breaklinks,colorlinks,citecolor=cvprblue]{hyperref}

\def\paperID{13154} 
\def\confName{CVPR}
\def\confYear{2024}

\usepackage{times}
\usepackage{epsfig}
\usepackage{graphicx}
\usepackage{amsmath}
\usepackage{amssymb}
\usepackage{colortbl}
\usepackage{caption}
\usepackage{microtype}

\usepackage[export]{adjustbox}

\colorlet{tablerowcolor}{gray!10}
\newcommand{\rowcolorize}{\rowcolor{tablerowcolor}}
\definecolor{forestgreen}{RGB}{34, 139, 34}

\title{Distributed Global Structure-from-Motion with a Deep Front-End}

\author{Ayush Baid $^{* \dag}$
\and
John Lambert$^{* \dag}$
\and
Travis Driver$^{* }$
\and
Akshay Krishnan$^{* }$
\and
Hayk Stepanyan
\and
Frank Dellaert\\
\hspace{-48mm}Georgia Tech
}

\begin{document}
\maketitle

\def\thefootnote{}\footnotetext{*Equal contribution.}\def\thefootnote{\arabic{footnote}}

\def\thefootnote{}\footnotetext{$^\dag$Work completed while at Georgia Tech.}\def\thefootnote{\arabic{footnote}}

\newcommand{\mbf}[1]{\mathbf{#1}}
\newcommand{\AK}[1]{\textbf{\color{red} [AK:{#1}]}}
\newcommand{\AY}[1]{\textbf{\color{blue} [AY:{#1}]}}
\newcommand{\JL}[1]{\textbf{\color{cyan} [JL:{#1}]}}

\begin{abstract}

While initial approaches to Structure-from-Motion (SfM) revolved around both global and incremental methods, most recent applications rely on incremental systems to estimate camera poses due to their superior robustness. Though there has been tremendous progress in SfM `front-ends' powered by deep models learned from data, the state-of-the-art (incremental) SfM pipelines still rely on classical SIFT features, developed in 2004. In this work, we investigate whether leveraging the developments in feature extraction and matching helps global SfM perform on par with the SOTA incremental SfM approach (COLMAP). To do so, we design a modular SfM framework that allows us to easily combine developments in different stages of the SfM pipeline. Our experiments show that while developments in deep-learning based two-view correspondence estimation do translate to improvements in point density for scenes reconstructed with global SfM, none of them outperform SIFT when comparing with incremental SfM results on a range of datasets. Our SfM system is designed from the ground up to leverage distributed computation, enabling us to parallelize computation on multiple machines and scale to large scenes. Our code is publicly available at \href{https://github.com/borglab/gtsfm}{github.com/borglab/gtsfm}.
    
\end{abstract}


\section{Introduction}


\begin{figure}
    \centering
    \includegraphics[width=\linewidth]{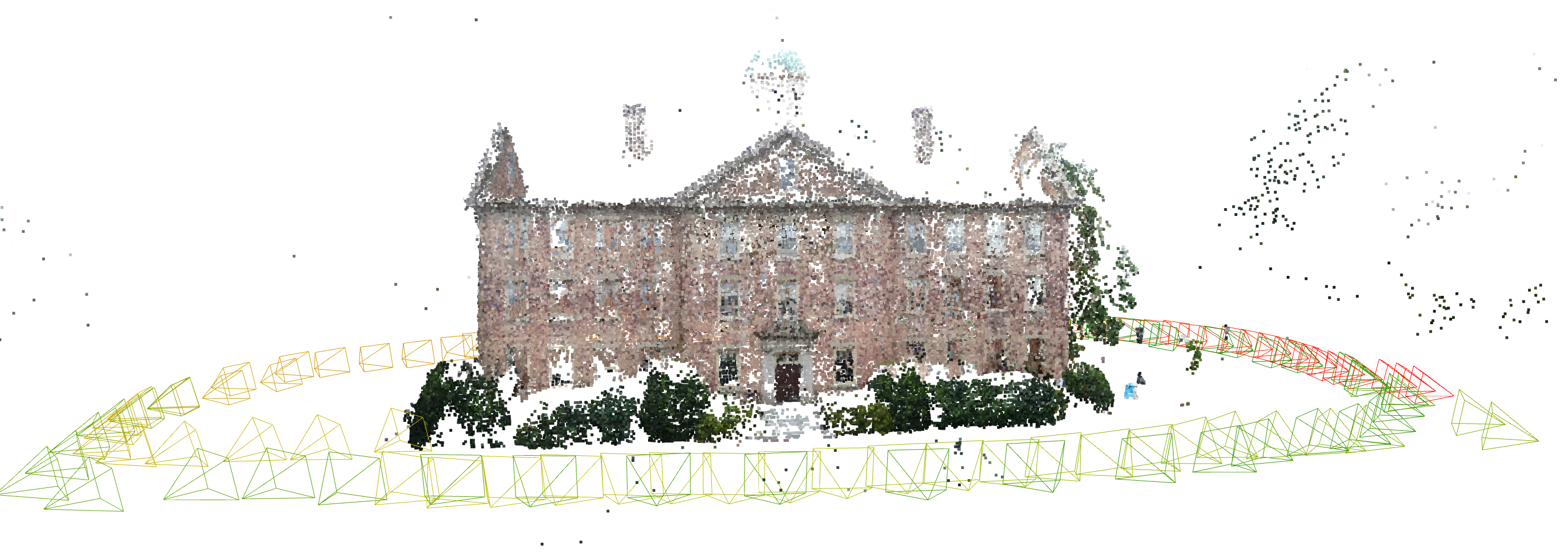}
    \includegraphics[width=0.4\linewidth]{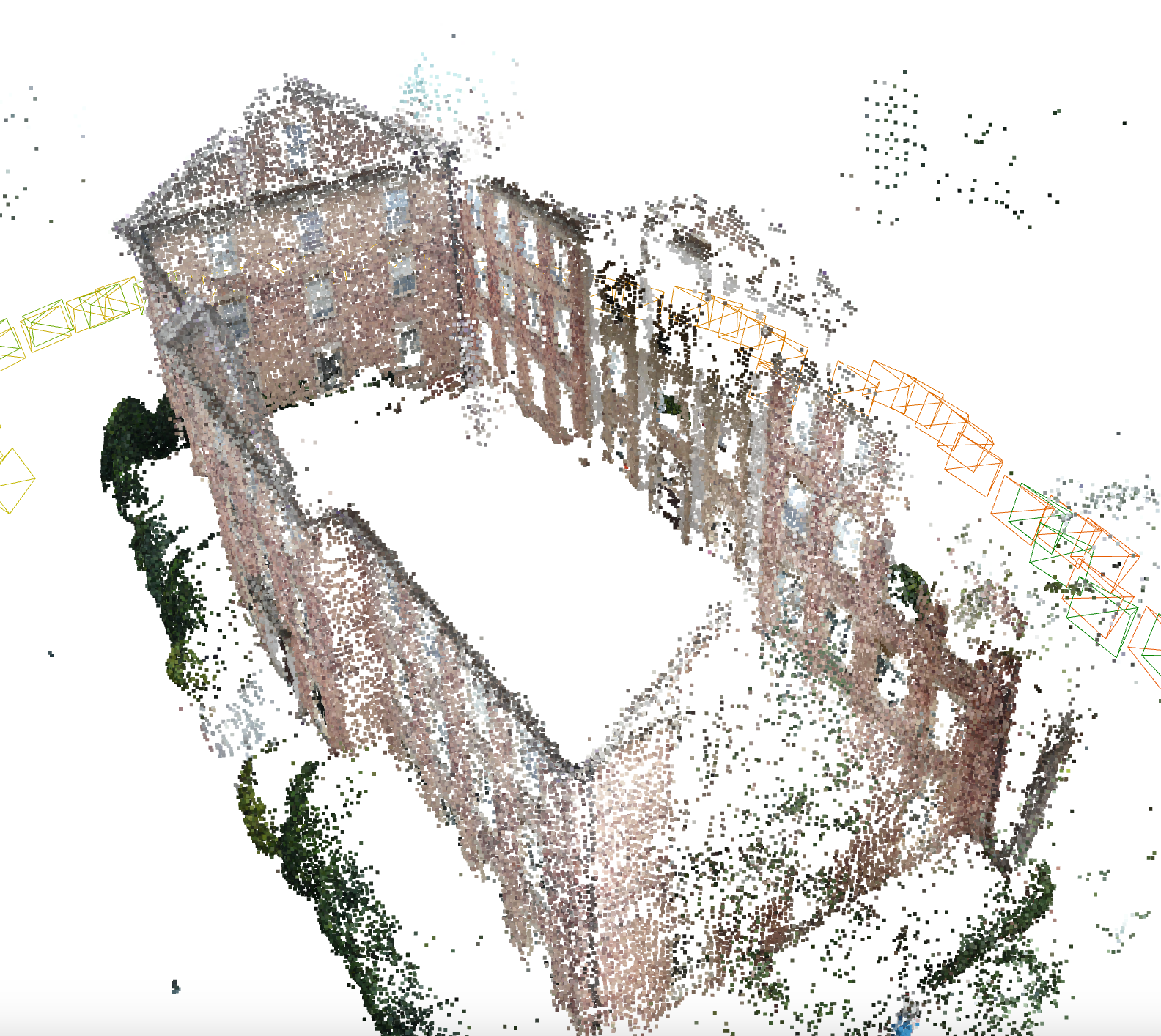}
    \includegraphics[width=0.4\linewidth]{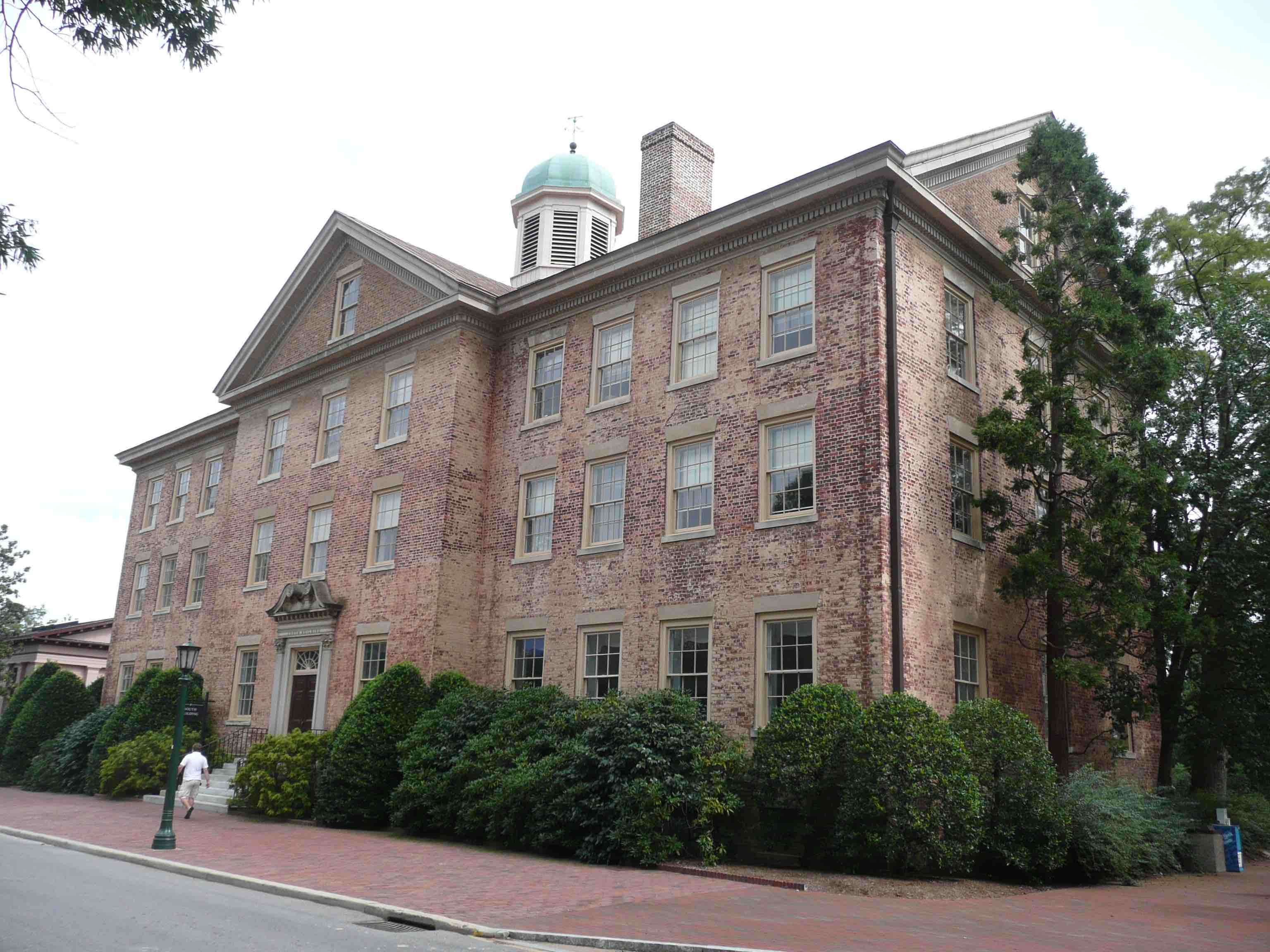}
    \caption{A sparse reconstruction of the UNC South Building using GTSfM with a deep LoFTR-based \cite{Sun21cvpr_LoFTR} front-end, with an example image input. Multi-view stereo is not used.}
    \label{fig:teaser-fig-south-bldg}
\end{figure}

Building accurate maps of the world is essential for spatial artificial intelligence (AI), with applications from autonomous robots to AR/VR. Structure-from-Motion (SfM) and multi-view stereo (MVS) have proven to be effective methods for creating maps with vision-only inputs. More broadly, SfM is a fundamental building block for 3d computer vision.
For certain types of scenes with simple to medium complexity, e.g.  datasets with $\sim$100 object-facing images, high-fidelity world models can be easily extracted with tools such as COLMAP \cite{Schonberger16cvpr_COLMAP}. 
These models and associated registered camera poses have enabled new breakthroughs in machine learning, through methods such as NeRF \cite{Mildenhall20eccv_NERF}, its variants \cite{Park21iccv_Nerfies, Reizenstein21iccv_CO3D}, Gaussian Splatting \cite{Kerbl23acmtog_GaussianSplat}, accurate monocular depth predictions for humans \cite{Li19cvpr_DepthsFrozenPeople}, and more.

Incremental SfM is the dominant paradigm, as global SfM suffers from a lack of accuracy, largely due to difficulty in reasoning about outliers globally in a single pass. 
However, to our knowledge, almost all global SfM systems today use classical frontends, reliant on feature matching with handcrafted descriptors, and the past decade has seen a flurry of work towards a \textit{deep front-end} for SfM. 
In this work, we analyze whether leveraging deep front-ends leads to an improvement in global SfM over classical front-ends.

In the modern AI era, computation on clusters with 1000's of GPUs or TPUs has become common \cite{Narayanan21ichpc_MegatronLM, Alayrac22neurips_Flamingo, Chen22arxiv_PALI, Chowdhery22arxiv_Palm, Touvron23arxiv_Llama, Cherti23cvpr_ReproducibleScalingLaws}, yet existing open-source SfM systems are not designed for and do not support multi-node distributed computation. 
Moreover, state-of-the-art SfM techniques are incremental, which makes them slow on very large datasets (e.g. with more than 500 images). 
Incremental SfM begins by finding a good first image pair, then triangulating 3D points from two-views, then adding one additional image pair at a time, registering it to the 3d points, then performing bundle adjustment, removing outliers, and continuing until all possible image pairs have been registered. 
This is certainly not the only possible approach;  \emph{global} SfM methods have also been explored for some time \cite{Govindu01cvpr_TwoViewConstraintsMotionEstimation,Govindu04cvpr_LieAlgebraicAveraging, Govindu06accv_RobustnessInMotionAveraging,Crandall11cvpr_DISCO,Enqvist11iccvw_NonsequentialSfM,Moulon13iccv_GlobalFusionSfM,Sweeney15iccv_OptimizingViewingGraph}. 
They avoid the need to do incremental pose estimation and refinement, but are known to suffer from poor accuracy \cite{Knapitsch17acmtg_TanksTemples}.

Why is global SfM not sufficiently accurate \cite{Knapitsch17acmtg_TanksTemples}? 
One way to think about the SfM problem is to divide SfM into a front-end that generates image correspondences, and a back-end (optimization).  
Without noise from `front-end' measurements, we find global SfM is close to exact. However, a single false positive can degrade performance.  
A key problem is that reasoning about outliers is challenging. Techniques from sequential methods, such as filtering out measurements inconsistent with the current model at each step, are not directly applicable in a global setting. 
It is harder to reason independently about which measurements are unreliable and, therefore, the most challenging aspect of SfM is correspondence, and when to trust correspondences, and of all the places where deep learning can be injected into the geometric modeling of SfM, feature matching is the most apparent part.
In this work, we aim to investigate whether injecting deep learning into the SfM front-end can rectify these accuracy shortcomings. 

\noindent Our contributions are as follows:
\begin{itemize}
    \item we provide an open-source \textit{global} SfM framework that is natively parallelizable and distributable on clusters, available as a Python package with no compilation required;
    \item we are among the first to analyze different deep front-ends in the context of global SfM;
    \item we demonstrate significant runtime decreases with respect to a state-of-the-art \textit{incremental} SfM pipeline.
\end{itemize}


\section{Related Work}

\subsection{Classical and Deep Front-Ends for SfM}
Traditional SfM systems compute keypoints, descriptors, matches, and verify correspondences
\cite{Snavely06siggraph_PhotoTourism, Snavely08ijcv_ModelingTheWorld, Schonberger16cvpr_COLMAP}. Surprisingly, little of 20 years of research towards using machine learning for the SfM front-end has been incorporated upstream into open-source libraries today, from COLMAP \cite{Schonberger16cvpr_COLMAP}, to OpenSfM \cite{Moulon16iwrrpr_OpenMVG} to OpenMVG \cite{Moulon16iwrrpr_OpenMVG} to Theia \cite{Sweeney15acmicm_TheiaSfM} (although some extensions are available \cite{Lindenberger21iccv_PixelPerfectSfM}). Early SfM systems were created before deep learning began to show promise in this domain, thus all components are hand-crafted. Furthermore, end-to-end methods for SfM aren’t accurate enough. Accordingly, we utilize local features and well-modeled geometry in the back-end. The literature on deep learning for correspondence estimation is vast; we refer the reader to  surveys  \cite{Jin21ijcv_PaperToPractice,Lambert22thesis}.

\subsection{Incremental SfM}
Incremental SfM traditionally uses point correspondences to iteratively establish camera poses and global structure. Pollefeys \emph{et al.} \cite{Pollefeys04ijcv_VisualModeling} introduced some of the modern framework for incremental SfM, which was expanded to massive datasets in Bundler \cite{Snavely08ijcv_ModelingTheWorld}, VisualSfM \cite{Wu133dv_IncrementalSfM}, and COLMAP \cite{Schonberger16cvpr_COLMAP}. The Tanks and Temples benchmark \cite{Knapitsch17acmtg_TanksTemples} indicates that COLMAP represents the state-of-the-art over both incremental and global SfM, but COLMAP can be slow in practice. COLMAP has been extended in many ways, such as the use of feature volumes to refine track measurements \cite{Lindenberger21iccv_PixelPerfectSfM}.

\begin{table}[]
\caption{A selection of SfM systems in the literature with open-source code and the capabilities they natively provide. Abbreviations: `Incr.' (Incremental), `Glob.' (Global), `Distr.' (Distributed).}
\vspace{-3mm}
\centering
\begin{adjustbox}{width=\linewidth}
\begingroup
\begin{tabular}{l cccccc}
\toprule
 \rowcolorize \textsc{\textbf{SfM System}}  & \textsc{\textbf{Incr.}} & \textsc{\textbf{Glob.}}    & \textsc{\textbf{Classic.}}  & \textsc{\textbf{Deep}} & \textsc{\textbf{Multi-}}  & \textsc{\textbf{Multi-}}  \\
 \rowcolorize    &  &     & \textsc{\textbf{Front}} & \textsc{\textbf{Front}} & \textsc{\textbf{Worker}} & \textsc{\textbf{Machine}}  \\
\rowcolorize & & & \textsc{\textbf{End}} & \textsc{\textbf{End}} & \textsc{\textbf{(Parallel)}} & \textsc{\textbf{(Distr.)}} \\
\midrule
\textsc{Bundler} \cite{Snavely06siggraph_PhotoTourism, Snavely08ijcv_ModelingTheWorld}          & \checkmark   &           &  \checkmark                   &                & \checkmark                        &                             \\
\rowcolorize \textsc{Theia} \cite{Sweeney15acmicm_TheiaSfM}             &             & \checkmark & \checkmark                 &                & \checkmark                     &                             \\
\textsc{OpenMVG} \cite{Moulon16iwrrpr_OpenMVG}          & \checkmark   & \checkmark & \checkmark                 &                &                         &                             \\
\rowcolorize \textsc{COLMAP} \cite{Schonberger16cvpr_COLMAP}           & \checkmark   &           &                     &                & \checkmark                     &                             \\
\textsc{OpenSfM} \cite{Gargallo16github_OpenSfM}          & \checkmark   &           & \checkmark                    &                &                         &                             \\
\rowcolorize \textsc{DagSfM}            & \checkmark            & \checkmark          &  \checkmark                   &                &  \checkmark                       &  \checkmark                           \\
\textsc{Pix-Perf. SfM} \cite{Lindenberger21iccv_PixelPerfectSfM} &          \checkmark   &           &                     &  \checkmark              &  \checkmark                       &                             \\
\rowcolorize \textsc{CReTA} \cite{Manam22eccv_CReTA}            &             &  \checkmark         &   \checkmark                  &                &                         &                             \\
 \textsc{Zhang et al.} \cite{Zhang23cvpr_RotAvgUncertaintyRobustLosses} & & \checkmark & \checkmark & \checkmark & & \\
\rowcolorize \textsc{GTSfM (Ours)}            &             & \checkmark       & \checkmark                 & \checkmark            & \checkmark                     & \checkmark                        \\
\bottomrule
\end{tabular}
\endgroup
\end{adjustbox}
\vspace{-5mm}
\end{table}

\subsection{Global SfM}
In Global SfM, also known as \emph{non-sequential} SfM or \emph{batch} SfM, one matches all possible image pairs, obtains a large number of two-view pose constraints, synchronizes all of these binary rotation measurements with some form of least squares, then estimates the camera positions, triangulates 3D points, and performs a single global bundle adjustment to refine points and poses. Both incremental and global SfM are subject to a feature matching stage with $O(n^2)$ complexity for $n$ images. Global SfM is not new -- Govindu introduced formulations for it two decades ago \cite{Govindu01cvpr_TwoViewConstraintsMotionEstimation,Govindu04cvpr_LieAlgebraicAveraging, Govindu06accv_RobustnessInMotionAveraging}.


An advantage of Global SfM is its ability to exploit redundancy. For a viewgraph $\mathcal{G} = (\mathcal{V}, \mathcal{E})$ with nodes as camera poses, and edges as two-view pose measurements, we can exploit all of the links in a graph to average out noise and distribute error evenly across the entire graph. For a dataset of $N$ images, there can be up to $\frac{N(N-1)}{2}$ pairs for which the relative motions can be estimated, potentially providing a highly redundant set of observations that can be efficiently averaged \cite{Govindu04cvpr_LieAlgebraicAveraging}. However, the community has yet to find techniques to use this redundancy to an advantage in accuracy.



Most global SfM systems rely upon rotation averaging \cite{Hartley13ijcv_RotationAveraging,Tron16cvprw_SurveyRotOptSfM} and subsequent translation averaging \cite{Olsson11_StableSfmUnordered,Jiang13iccv_GlobalLinearMethodCamPoseReg,Wilson14eccv_1DSfM,Cui15bmvc_LinearGlobTransEstFeatTracks,Goldstein16eccv_Shapefit} for accurate bundle adjustment initialization, although other formulations exists, such as discrete MRF-based methods \cite{Crandall11cvpr_DISCO} or hierarchical SfM methods that  merge camera clusters formed via incremental SfM \cite{Chen20pr_GraphBasedParallelSfM}. For example, Google's city-scale Street View SfM \cite{Klingner13iccv_StreetViewSfM} combined clusters of 1500 cameras. OpenMVG \cite{Moulon16iwrrpr_OpenMVG} uses a least-squares rotation averaging \cite{Martinec07cvpr_RotationTranslationAveraging} technique. Other rotation averaging methods have since been proposed, such as Shonan  \cite{Dellaert20eccv_Shonan} or RCD \cite{Parra21cvpr_RotationCoordinateDescent}. Concurrent work weights two-view rotation measurements by two-view bundle adjustment uncertainties \cite{Zhang23cvpr_RotAvgUncertaintyRobustLosses} in rotation averaging, but we did not find this to yield accuracy gains in our experiments. CReTA \cite{Manam22eccv_CReTA} accounts for outliers in translation averaging by iteratively reweighting point correspondences and thus translation measurements.


\subsection{Outlier Rejection for SfM} \label{ss:outlier_rejection}
Outlier rejection is critical to successful SfM. Not only is it very difficult to triangulate points from inexact camera positions, but bundle adjustment with Gaussian noise models cannot deal with outliers. While incremental systems can reject outliers at each registration stage via reprojection error, global SfM does not enjoy this privilege, and its performance is heavily reliant upon low outlier rates. Global SfM systems instead utilize a number of carefully-crafted outlier rejection techniques to eliminate noisy measurements to prevent them from playing a role in joint optimization.

\noindent \textbf{Relative Pose Consistency} The most common outlier rejection approaches rely upon cycle consistency \cite{Sharp01icra_MultiviewCycleBasis} of relative measurements within triplets. For example, the deviation from identity of composed relative rotations in a cycle strongly suggests the magnitude of relative rotation errors \cite{Enqvist11iccvw_NonsequentialSfM,Moulon13iccv_GlobalFusionSfM,Moulon16iwrrpr_OpenMVG}, as used in OpenMVG and Theia. By accumulating these deviations over a large set of loops one can obtain the statistics needed to infer the set of false positives \cite{Zach10cvpr_LoopConstraints}. 
 Others generate random spanning trees from relative poses in a RANSAC-like scheme \cite{Olsson11_StableSfmUnordered} for estimating global camera poses, such that ${}^w\mathbf{R}_i =  {}^w\mathbf{R}_j \big({}^j \mathbf{R}_i\big)$ roughly holds for as many relative rotations as possible.
Theia \cite{Sweeney15acmicm_TheiaSfM} also uses filtering based on global-to-relative agreement heuristics.  1dSfM \cite{Wilson14eccv_1DSfM}
rejects outlier translation directions based on consistent ordering on 1d projections. Instead of using hand-crafted heuristics, Phillips \cite{Phillips19arxiv_GraphsLearningCycleConsistent} uses graph neural networks (GNNs) to introduce learning-based cycle consistency on the keypoint match graph, instead of relative pose graph. 

\noindent \textbf{Learned Matchability Classifiers} Other methods such as SALVe \cite{Lambert22eccv_salve} and  Doppelgangers \cite{Cai23iccv_Doppelgangers} align views and predict a matchability confidence for each putative image pair with a ResNet CNN. However, we find the recall of the Doppelgangers pretrained classifier to be too low for use in practice.  



\begin{figure}[t!]
    \centering
    \includegraphics[width=\linewidth]{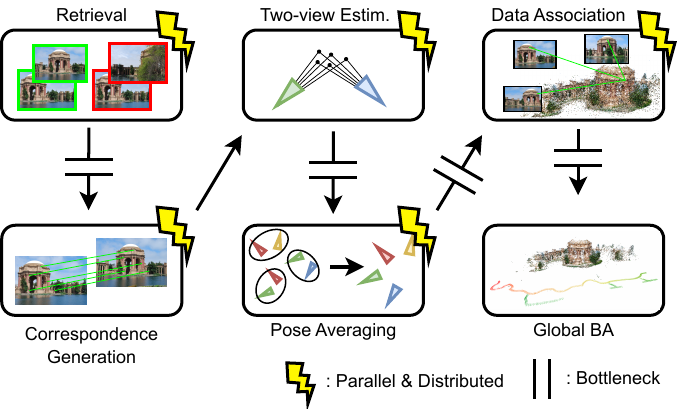}
    \vspace{-3mm}
    \caption{GTSfM system overview. A `bottleneck' indicates that all parallelized tasks from the previous module must be completed before proceeding to the next.}
    \label{fig:system-overview}
\end{figure}

\section{Approach}

Given a collection of images, our global SfM system, GTSfM, executes the following stages: retrieval of image-pairs based on view similarity, correspondence generation from retrieved image pairs, geometric verification and two-view pose estimation, rotation and translation averaging, data association, and global bundle adjustment.
We group the image-pair retrieval, correspondence generation, and two-view pose estimation into the `front-end', and the rest of the stages into the `back-end.' The front-end produces a view graph which defines a relative pose (rotation and translation up to a scale factor) for a subset of the image-pairs retrieved. The back-end estimates the camera pose for all cameras in the view-graph computed by the front-end. 

As shown in Figure \ref{fig:system-overview}, we modularize our global SfM system in a way that allows us to easily evaluate Global SfM's end-to-end performance using state-of-the-art techniques for different components, in ways which have not been performed previously. Moreover, our global SfM framework allows us to fully parallelize and distribute several steps of the SfM system: image pair retrieval, correspondence generation, relative translation and rotation outlier rejection, and data association.
We therefore design our library to support parallelization across multiple machines in a cluster via Dask \cite{Rocklin15scipy_Dask} -- enabling SfM to scale to larger scenes with more compute. 
We briefly describe the different techniques that GTSfM uses for each component below, with emphasis on changes that were made to benefit end-to-end performance. 

\subsection{Retrieval}
Given a collection of $N$ images, we retrieve image pairs with indices $(i,j)$ using various methods. 
For unordered image sets, matching images with little to no overlap could result in spurious image pairs being inserted into the graph, thus degrading the solution.
Moreover, for datasets with sequential image ordering, retrieval matches are necessary to provide loop closures and provide sufficient pose estimation accuracy. 
Accordingly, we compute \textit{global} image descriptors using a deep image-retrieval network, NetVLAD~\cite{Arandjelovic16cvpr_NetVLAD}, and only match the top-$k$ image pairs according their similarity scores.
We leverage a combination of sequential retrieval (i.e. based on a fixed look-ahead horizon) and image similarity-based retrieval for ordered datasets.

\subsection{Correspondence Generation}

\noindent\textbf{Front-end} We explore a number of different front-ends in order to obtain point correspondences. 
The front-end consists of correspondence generation, then relative pose estimation and epipolar geometry-based verification, followed by two-view bundle adjustment (as recommended by \cite{Julia17psivt_CriticalReviewTrifocalTensor}), and pair rejection based on the amount of support (i.e. inliers). 
Two types of correspondence generation are possible: sparse feature detection and matching (e.g. SIFT \cite{Lowe04ijcv_SIFT} + mutual nearest neighbor matcher), or direct dense image matching (e.g. LoFTR \cite{Sun21cvpr_LoFTR}). 
In the literature, to our knowledge, only sparse feature detection and matching has been used in Global SfM. Each front-end that we analyze includes Essential matrix estimation with RANSAC, using Nister's 5-Point algorithm \cite{Nister04tpami_FivePointRelativePose}. 
We assume good prior knowledge of the intrinsics (e.g., provided in EXIF), as this is a common use case for modern drone or single-camera (e.g. NeRF) applications.

\noindent\textbf{View Graph Generation} Let $\mathcal{G} = (\mathcal{V}, \mathcal{E})$ be the input viewgraph, where $\mathcal{V}$ and $\mathcal{E}$ denotes
the set of vertices and edges in $\mathcal{G}$ respectively. To each edge $(i,j) = e_k \in \mathcal{E}$, we assign a 3D rotation $\mathbf{R}_k \in \mathbb{SO}(3)$ and translation $\mathbf{t}_k \in \mathbb{R}^3$, equivalent to $({}^j\mathbf{R}_i , {}^j\mathbf{t}_i )$ between camera vertices $i$ and $j$. In the subsequent stage, our objective is to identify $\mathcal{G}^\prime$ which is noise-free and self-consistent.






\subsection{Rotation Averaging}

\begin{figure}
    \centering
    \includegraphics[width=0.3\linewidth]{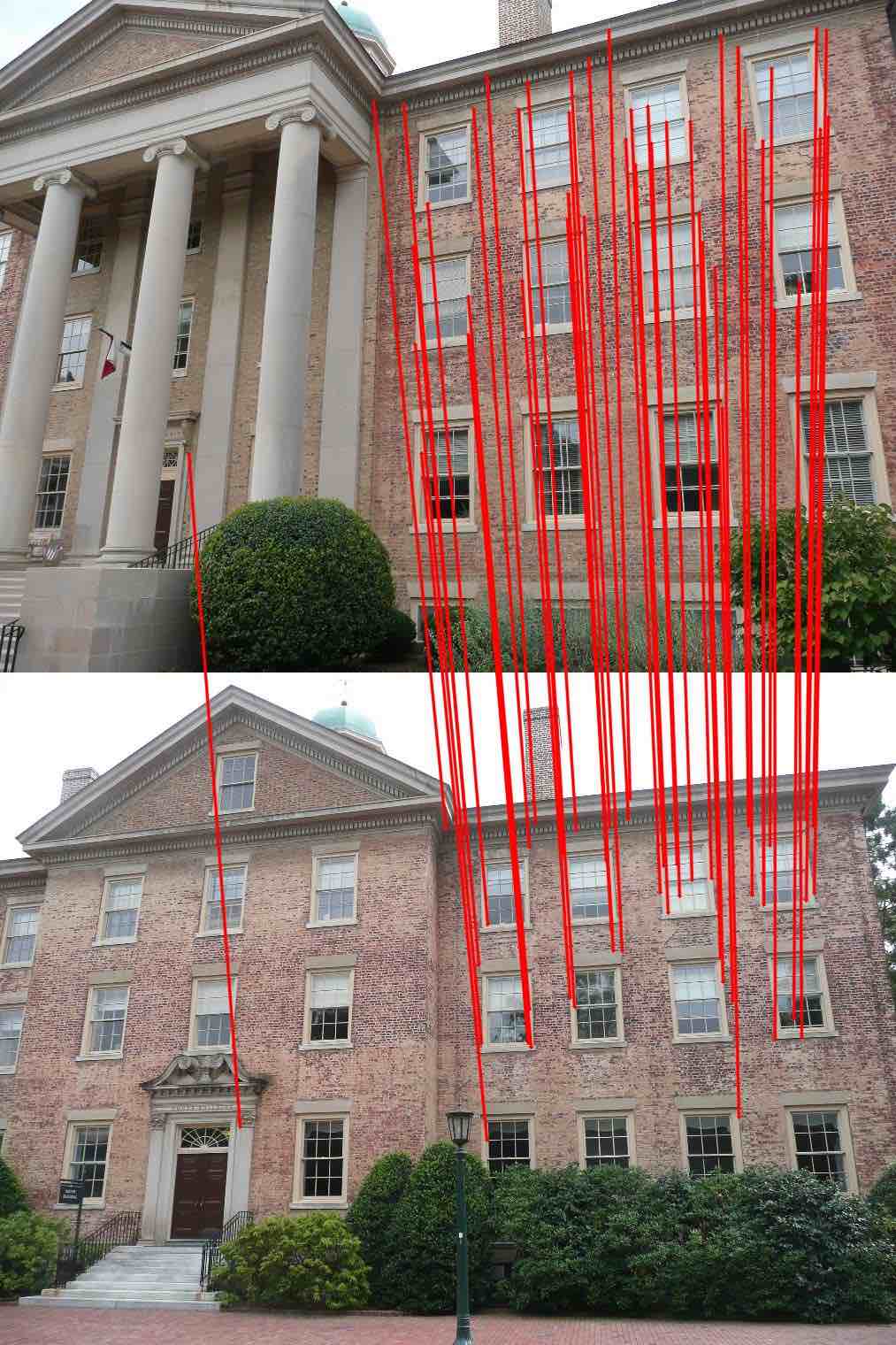}
    \includegraphics[width=0.3\linewidth]{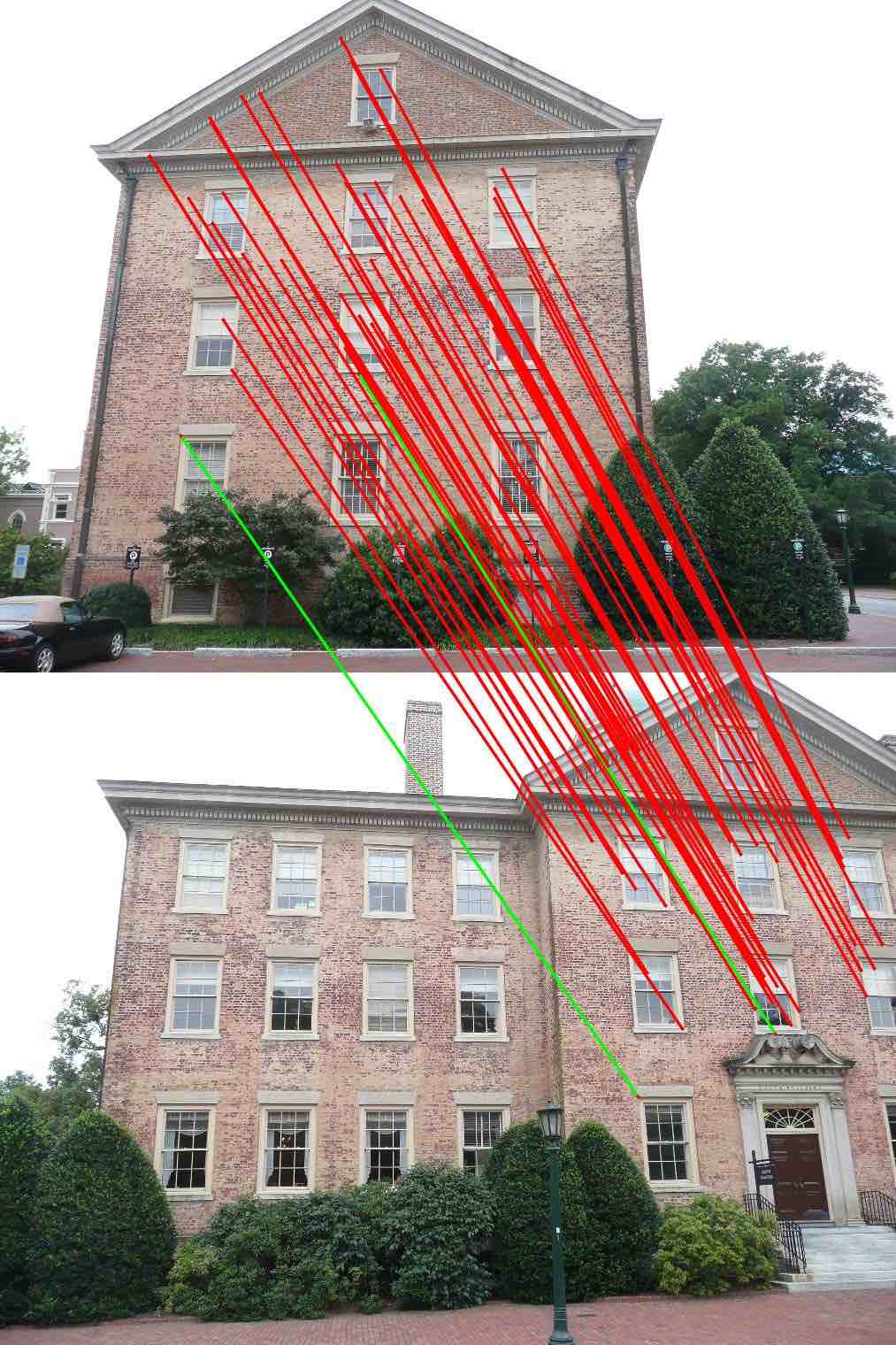} 
    \caption{Doppelgangers due to repetitive structures lead to wildly incorrect relative motion estimates when generating correspondences using the LightGlue \cite{Lindenberger23iccv_LightGlue} matcher on the South Building dataset.}
    \label{fig:doppelganger-examples}
\end{figure}

Two-view poses estimated from the front-end are noisy in most practical settings, creating the greatest challenge for a global SfM system. The failure of feature matchers to disambiguate repetitive structures, e.g. doppelgangers \cite{Cai23iccv_Doppelgangers} (see Figure \ref{fig:doppelganger-examples}), requires global reasoning about outlier relative rotations.
Techniques that estimate the initial global camera pose, e.g. rotation and translation averaging, are especially sensitive to noise in relative poses. We therefore use outlier rejection techniques to reduce the noise levels present in the input to these components.

\noindent \textbf{Rotation Cycle Consistency} In a noise-free setting, we have a consistency criterion over length-3 loops $L$ of image pairs $e_k$ \cite{Olsson11_StableSfmUnordered}, i.e. $|L|=3$ \cite{Zach10cvpr_LoopConstraints, Olsson11_StableSfmUnordered}:
\begin{equation}
\begin{array}{ll}
\mathbf{R}_L = \mathbf{R}_{e_3} \cdot \mathbf{R}_{e_2} \cdot \mathbf{R}_{e_1} = I,  & e_k \in L.
\end{array}
\end{equation}
%

\noindent However, in a setting with noisy measurements $\mathbf{R}_{e_k}$, we can find all loops $L \in \mathcal{L}_k$ that an image pair $e_k$ participates in, i.e. $\mathcal{L}_k = \{ L \mid e_k \in L \}$ and accept those with a cycle error summary statistic below a certain threshold $\epsilon_{cycle}$. More formally, we assign a label $\{y_k\} \in \{0,1\}$ for each image pair edge $e_k$ according to:
\begin{equation}
    y_k = \underset{L \in \mathcal{L}_k }{\mbox{min}} \Big\{ \| \log(\mathbf{R}_L)^\vee \|_2 \mid e_k \in L \Big\} < \epsilon_{cycle}
\end{equation}
\noindent where $\log$ is the logarithmic map and $\vee$ is the $vee$ operator (see \cite{chirikjian2011}). We reject all image pairs that fall into $\mathcal{E}_{reject} = \{e_k \mid y_k = 0 \}$ to form $\mathcal{G}^\prime$. In practice, we find that the \texttt{median} operator alone suffers from poor enough recall that loop closures may be omitted altogether. Accordingly, we first use the \texttt{min} operator to maximize recall, and then afterwards filter again according to \texttt{median} operator, which allows some false positives. 

\noindent \textbf{Rotation Averaging Optimization} We solve for the rotations of global camera poses via multiple rotation averaging \cite{Hartley13ijcv_RotationAveraging}. We use only the relative rotation measurements corresponding to image pair edges in the largest connected component of $\mathcal{G}^\prime$. The problem can be defined as:
\begin{equation}
\underset{\mathbf{R_1}, ... \mathbf{R_n} \in \mathbb{SO}(3)}{\mbox{argmin}} \sum\limits_{(i,j) \in \mathcal{E}} d({}^i\hat{\mathbf{R}}_{j}, {}^i\mathbf{R}_w  {}^w\mathbf{R}_j),
\end{equation}
where $\hat{\mathbf{R}}$ denotes a rotation \textit{measurement}. 
We use Shonan Averaging \cite{Dellaert20eccv_Shonan}, which uses a convex relaxation of the following maximum likelihood problem:
\begin{equation}
\underset{\mathbf{R} \in \mathbb{SO}(d)^n}{\max} \sum\limits_{(i,j) \in \mathcal{E}} \kappa_{ij} \mbox{tr} ( {}^w\mathbf{R}_i {}^i\hat{\mathbf{R}}_{j} {}^j\mathbf{R}_w),
\end{equation}
%
%
where $\kappa_{ij} \geq 0$ are concentration parameters for an assumed Langevin noise model. While Shonan is proven to converge to globally optimal solutions under very mild noise levels, we find that real-world SfM noise distributions significantly exceed such noise levels, making the cycle-consistency outlier rejection stage (view graph estimation) critical.
%

\subsection{Translation Averaging} 
Given estimated camera rotations $\{\mathbf{R}_i\}_{i=1}^N$ in a global frame, along with pairwise relative translation directions,  we recover the position $\{\mathbf{t}_i\}_{i=1}^N$ of each camera in a global frame using 1dSfM-based \cite{Wilson14eccv_1DSfM} translation averaging. We transform the relative unit translations to a world frame using the estimated rotations, project them (and landmark rays) along several random directions, and find measurements consistent with the minimum-feedback arc set (MFAS) problem  \cite{Wilson14eccv_1DSfM}, and remove the inconsistent edges. We then optimize a chordal error using a Huber cost function:
\begin{equation}
    E_{ch}(\cdot) = \sum\limits_{(i,j) \in \mathcal{E}} d_{ch} \Big( \mathbf{\hat{t}}_{ij}, \frac{ \mathbf{t}_j - \mathbf{t}_i  }{ \| \mathbf{t}_j - \mathbf{t}_i \| } \Big)^2
\end{equation}
where $d_{ch}$ is defined as $d_{ch}(\mathbf{u}, \mathbf{v}) = \| \mathbf{u} - \mathbf{v} \|_2$.


\subsection{Data Association }
We obtain 2D tracks for points detected by the front-end using only the measurements in the cameras for which averaging succeeded, using a disjoint-set forest algorithm. We triangulate the 2D measurements using Direct Linear Transform (DLT) (along with RANSAC to filter outlier measurements) to obtain initial 3D positions for the tracked features.  

\subsection{Bundle Adjustment}
We refine the initial camera poses and triangulated point cloud using bundle adjustment.  The bundle adjustment optimization problem can be formally defined as follows: Given sparse points $\{ \mathbf{P}_j \in \mathbb{R}^3\}$, intrinsic parameters $\{ \mathbf{C}_i\}$ of the cameras, and initial camera poses $\{({}^w\mathbf{R}_i, {}^w\mathbf{t}_i) \in \mathrm{SE}(3)\}_{i=1}^N$ for each image \cite{Lindenberger21iccv_PixelPerfectSfM}:
\begin{equation}
    E_{BA}(\cdot) = \sum\limits_j \sum\limits_{(i,u) \in \mathcal{T}(j)} \| \Pi({}^i\mathbf{R}_w \mathbf{P}_j + {}^i\mathbf{t}_w ; \mathbf{C}_i)  - \mathbf{p}_u \|_{\gamma}
\label{eqn:ba}
\end{equation}
where $\mathcal{T}(j)$ is the set of images $i$ and keypoints $u$ in track $j$, $\Pi(\cdot)$ projects to the image plane, and $\|\cdot\|_{\gamma}$ is a robust norm. We use the Bundler \cite{Snavely08ijcv_ModelingTheWorld} camera model for calibration, with a single focal length $f$, two radial distortion coefficients $\kappa_1,\kappa_2$ (quadratic and quartic), and a image center $(u_0,v_0)$ (principal point) in pixels. We initialize the camera poses from the global camera rotations estimated from Shonan and the global camera positions estimated from 1dSfM, and refine points and camera parameters using the Levenberg-Marquardt algorithm \cite{Levenberg44qam_MethodNonlinearLeastSquares, Marquardt63siam_AlgorithmLeastSquaresNonlinear}. We use multiple rounds of bundle adjustment, filtering tracks at increasingly tight track reprojection error thresholds.
\section{Experimental Results}

\subsection{Implementation Details}

\noindent \textbf{Cluster Details} We execute all GTSfM variants on a cluster of 4 nodes, each equipped with 1 NVIDIA GeForce RTX 3080 GPU with 12 GB RAM and 12th Gen Intel Core i5-12600K with 16 cores and 32 GB of CPU RAM, running Ubuntu 20.04.4.

\noindent \textbf{Image Pair Retrieval} For each baseline, we include sequential image pairs with a maximum lookahead horizon of 10 frames. We augment these pairs by retrieving additional image pairs using NetVLAD \cite{Arandjelovic16cvpr_NetVLAD}, with up to 5 retrieval matches per image frame (for datasets with less than 500 images, and 15 retrieval matches otherwise. We compute dot products between global image descriptors in parallel, with each worker processing $50 \times 50$ sub-blocks of the similarity matrix, with a minimum required similarity score of 0.3.

\subsubsection{Feature Matching Baselines}
 A maximum of 5000 keypoints are used for each of the following front-ends, with images up to 760 px in resolution.

\noindent \textbf{SIFT Front-end} We use SIFT keypoints \cite{Lowe04ijcv_SIFT}, with mutual nearest neighbor matching and a ratio check.

\noindent \textbf{SuperGlue Front-end}  Uses SuperPoint \cite{Detone18cvprw_SuperPoint} and a GNN-based SuperGlue  \cite{Sarlin20cvpr_SuperGlue} matcher. SuperPoint was trained by the original authors on MS-COCO \cite{Lin14eccv_MSCOCO}.

\noindent \textbf{LightGlue Front-end}  Uses SuperPoint \cite{Detone18cvprw_SuperPoint} and a Transformer\cite{Vaswani17neurips_AttentionIsAllYouNeed}-based LightGlue  \cite{Lindenberger23iccv_LightGlue} matcher. Lightglue is fine-tuned by the original authors on MegaDepth \cite{Li18cvpr_MegaDepth}.

\noindent \textbf{LoFTR Front-end} \cite{Sun21cvpr_LoFTR} Unlike previous image matchers, LoFTR (trained on \cite{Li18cvpr_MegaDepth}) introduced a paradigm of correspondence generation without a common set of keypoints that would be shared when matching. 
We find that naively using the union of all keypoints across all image pairs creates 100s of thousands of 2-view tracks. 
Accordingly, we perform non-maximum suppression (NMS) by merging keypoint coordinates within the same image, with a NMS radius of 3 pixels, similar to \cite{He21tr_LoftrImwChallenge}, and unlike \cite{He23arxiv_DetectorFreeSfm} that quantizes LoFTR keypoints onto a grid. 
In effect, we merge correspondences (tracks), and suppress duplicate keypoints. 

\begin{figure*}
	\centering
	\small
	\begin{tabular}{@{\hspace{0.0mm}}c@{\hspace{1.0mm}}c@{\hspace{1.0mm}}c@{\hspace{2.5mm}}c@{\hspace{1.0mm}}c@{\hspace{1.0mm}}c@{\hspace{0.0mm}}}
		Input & Ground Truth & SIFT & SuperGlue & LightGlue & LoFTR\\
		\includegraphics[width=0.16\linewidth]{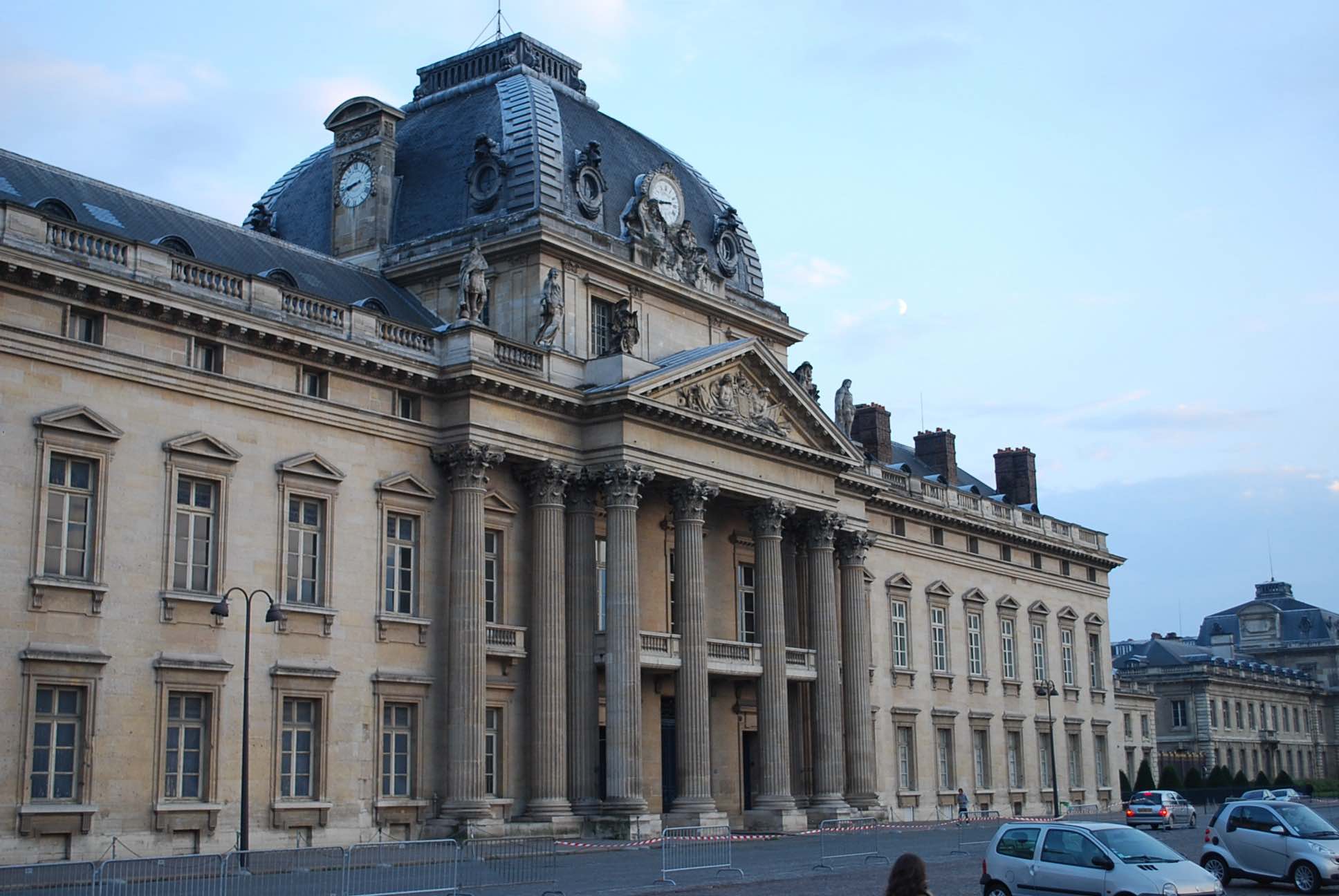}&
		\includegraphics[width=0.16\linewidth]{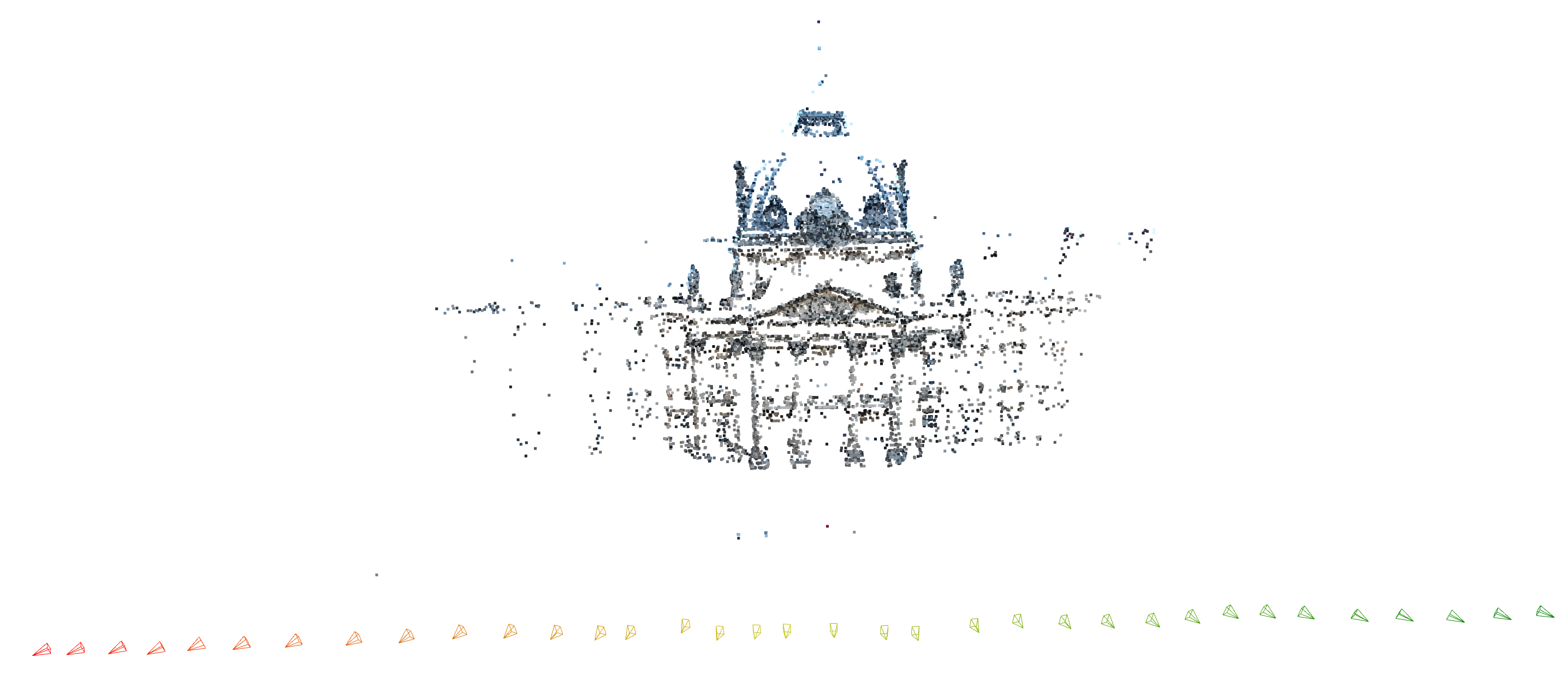}&
  		\includegraphics[width=0.16\linewidth]{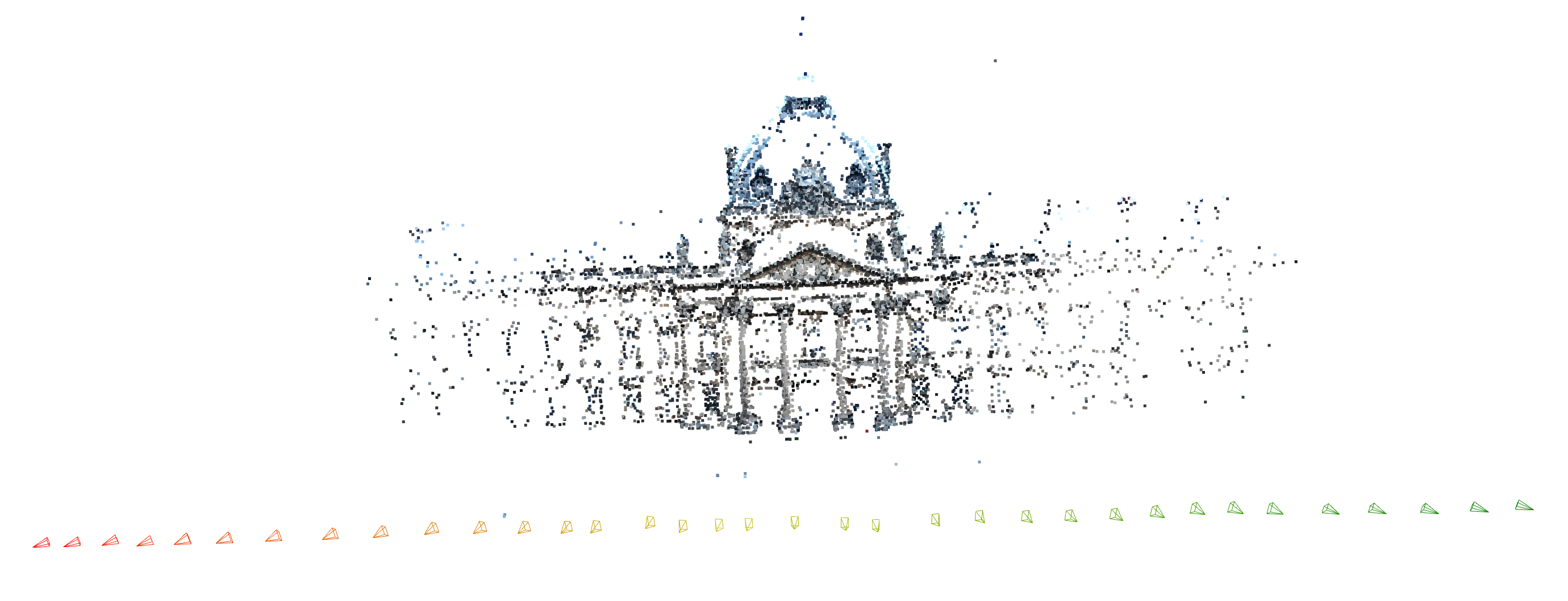}&
    	\includegraphics[width=0.16\linewidth]{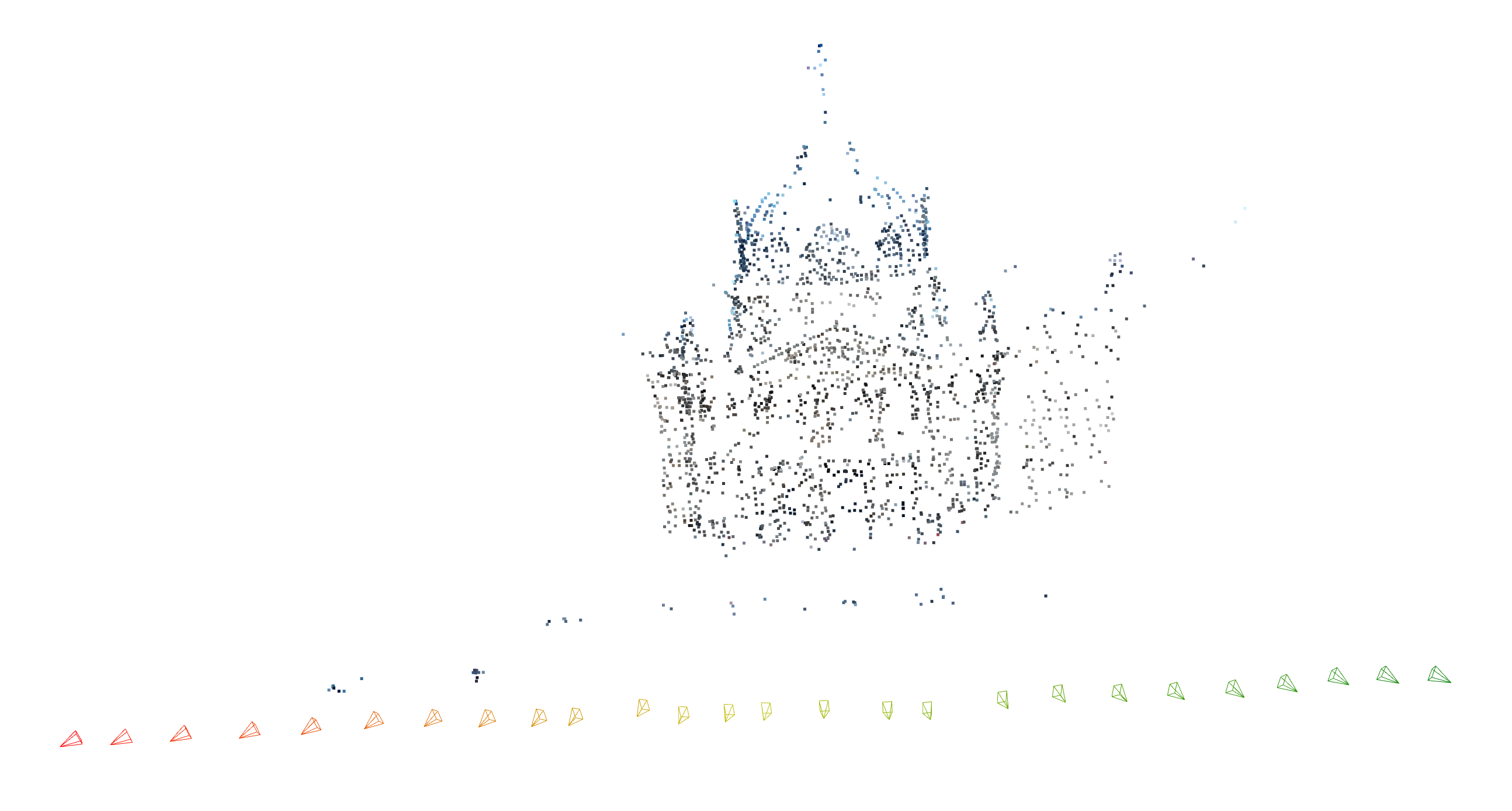}&
		\includegraphics[width=0.16\linewidth]{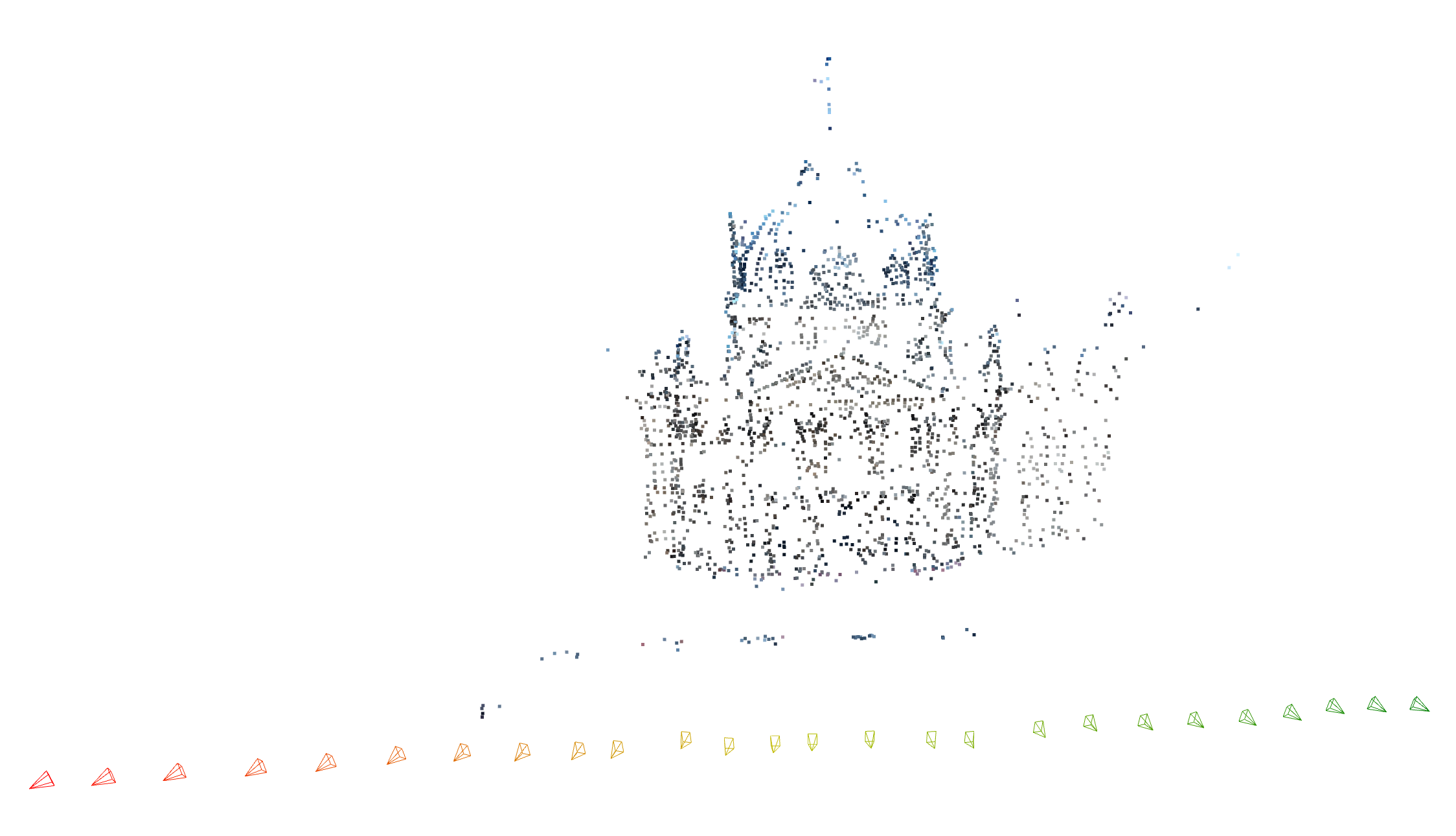}&
		\includegraphics[width=0.16\linewidth]{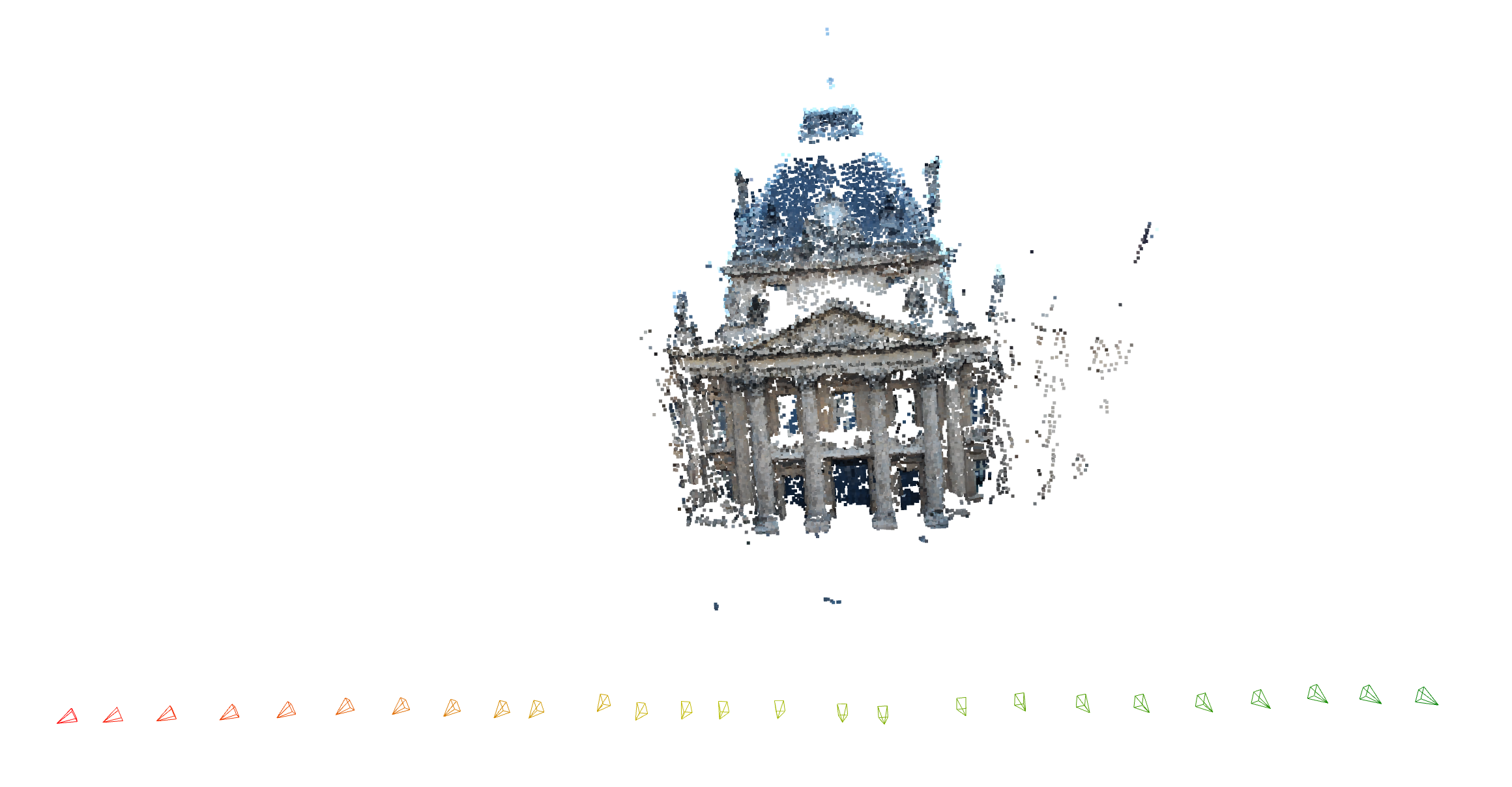}\\
		\includegraphics[width=0.16\linewidth]{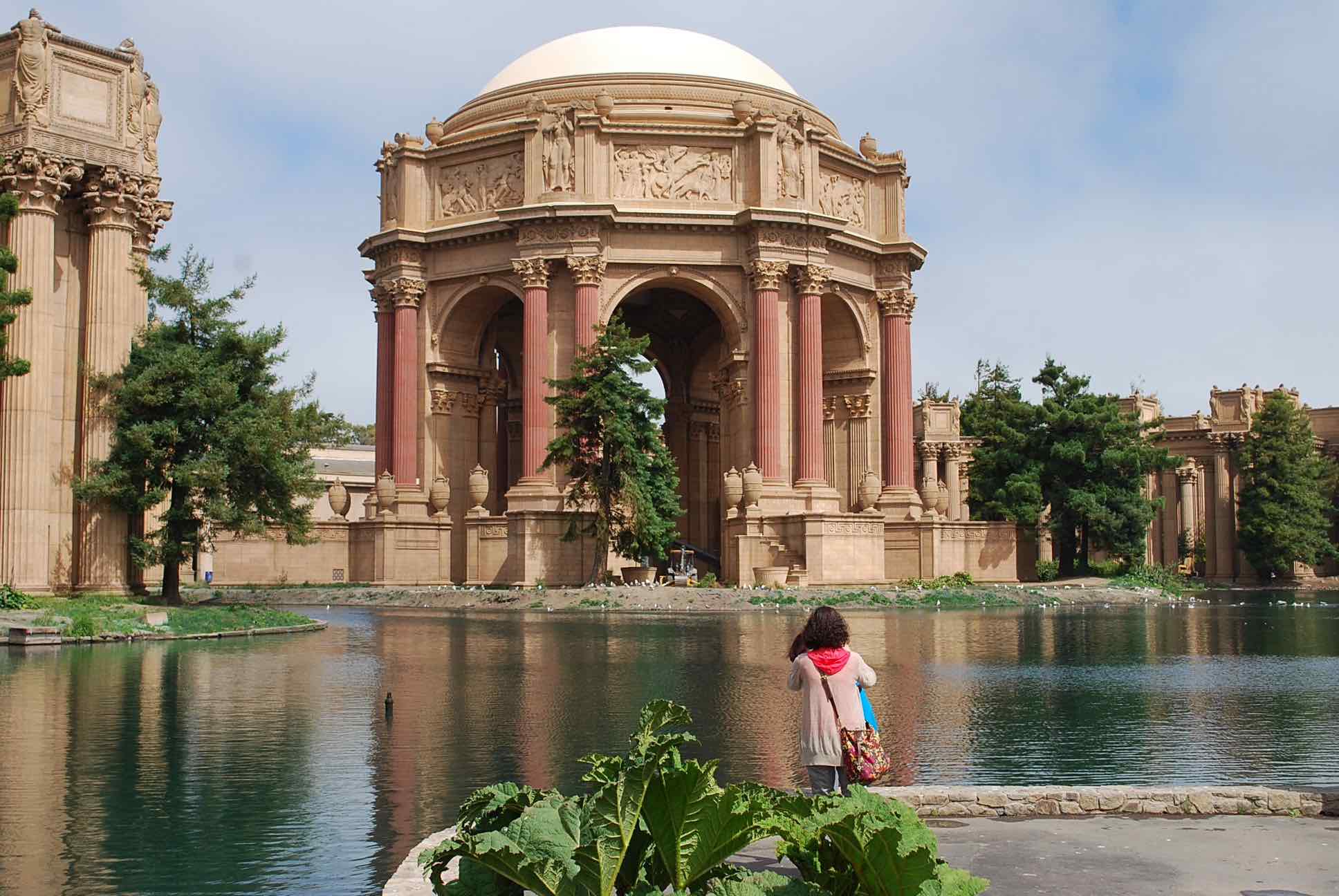}&
		\includegraphics[width=0.16\linewidth]{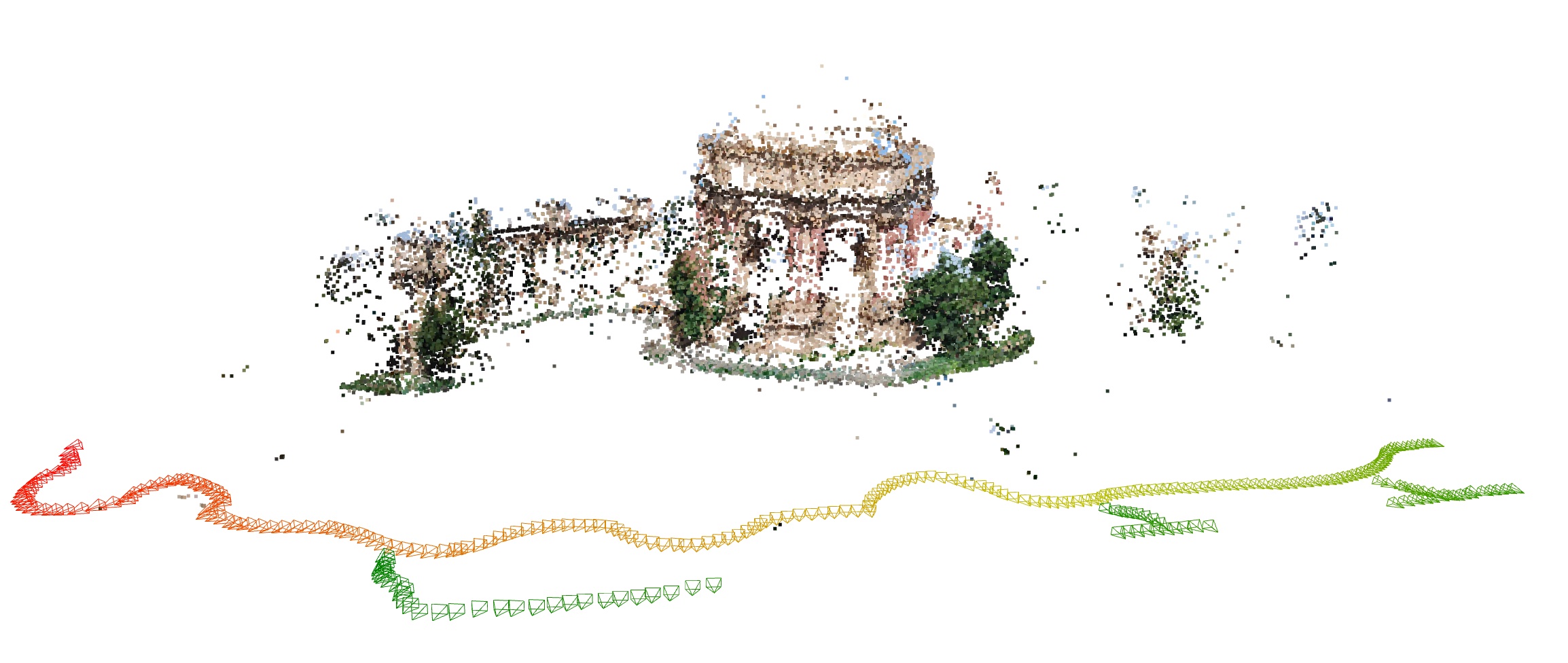}&
		\includegraphics[width=0.16\linewidth]{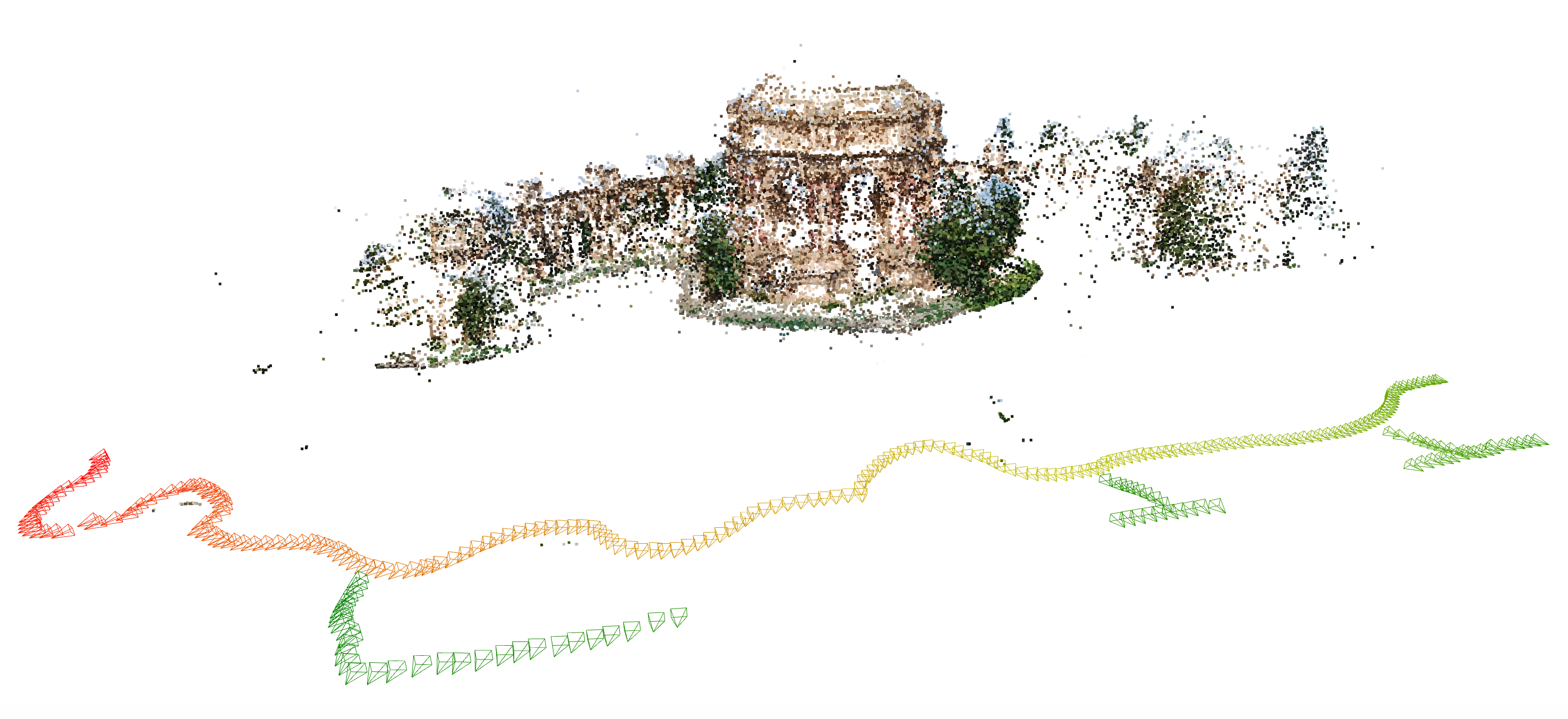}&
		\includegraphics[width=0.16\linewidth]{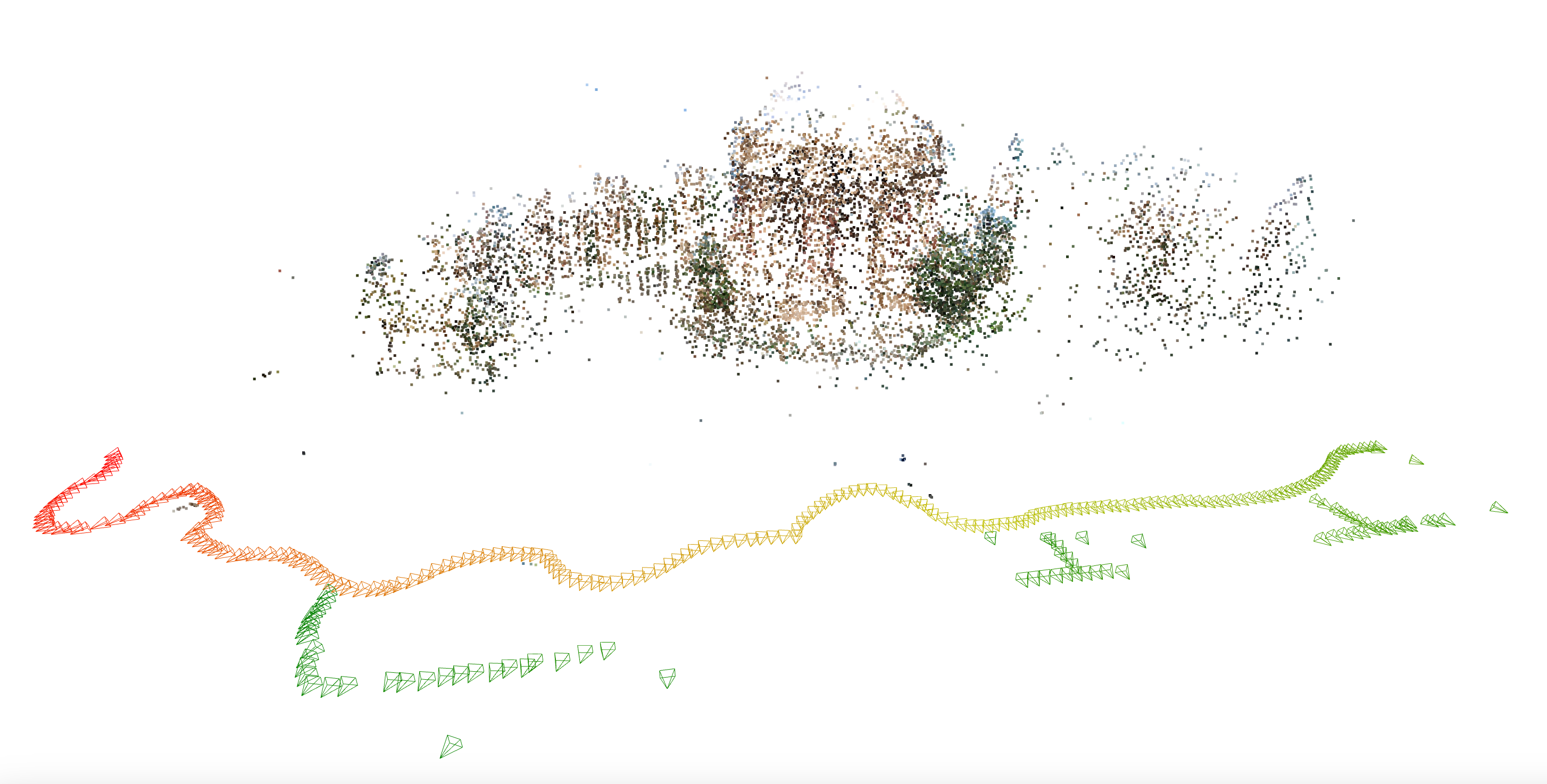}&
		\includegraphics[width=0.16\linewidth]{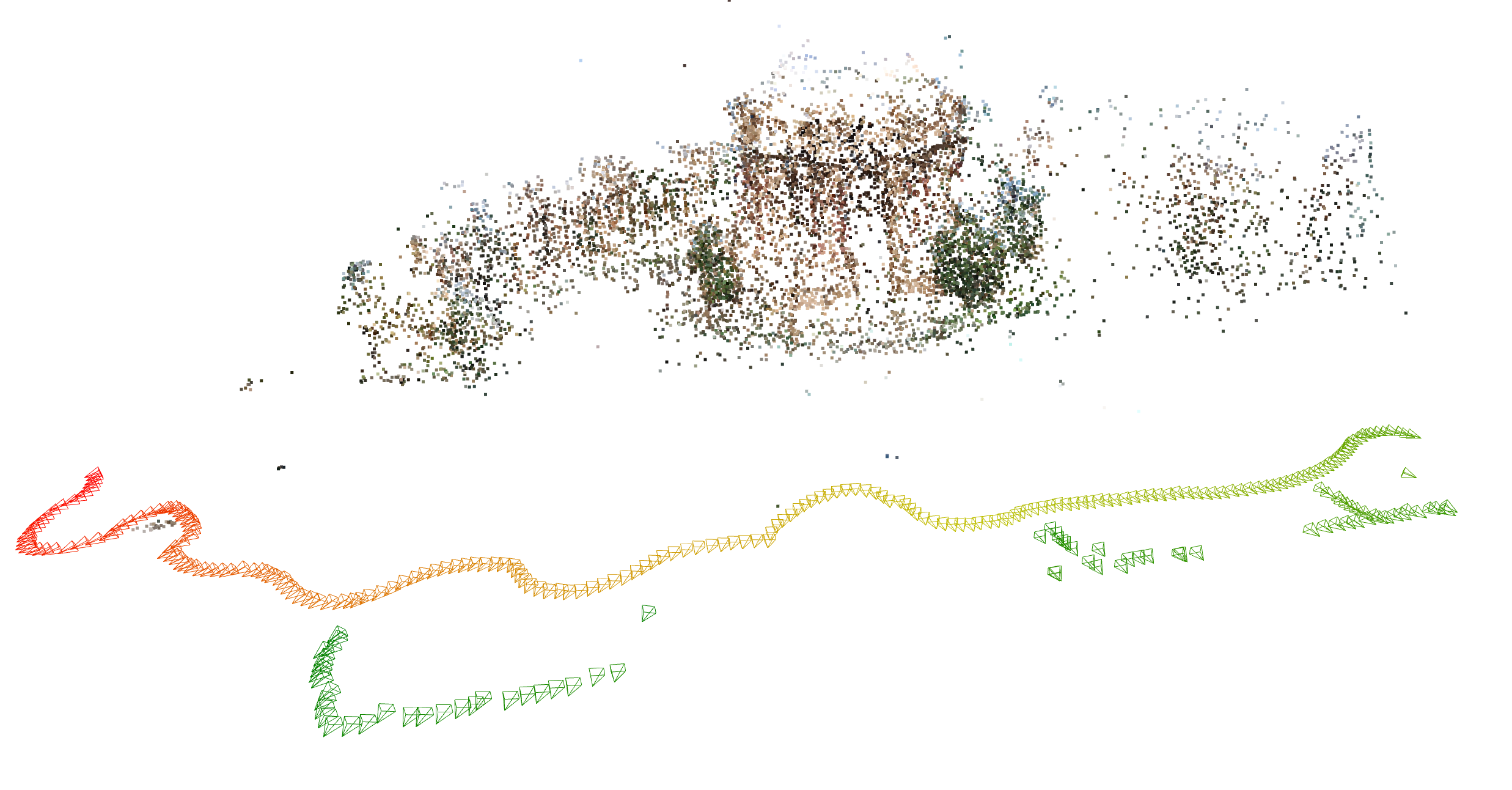}&
		\includegraphics[width=0.16\linewidth]{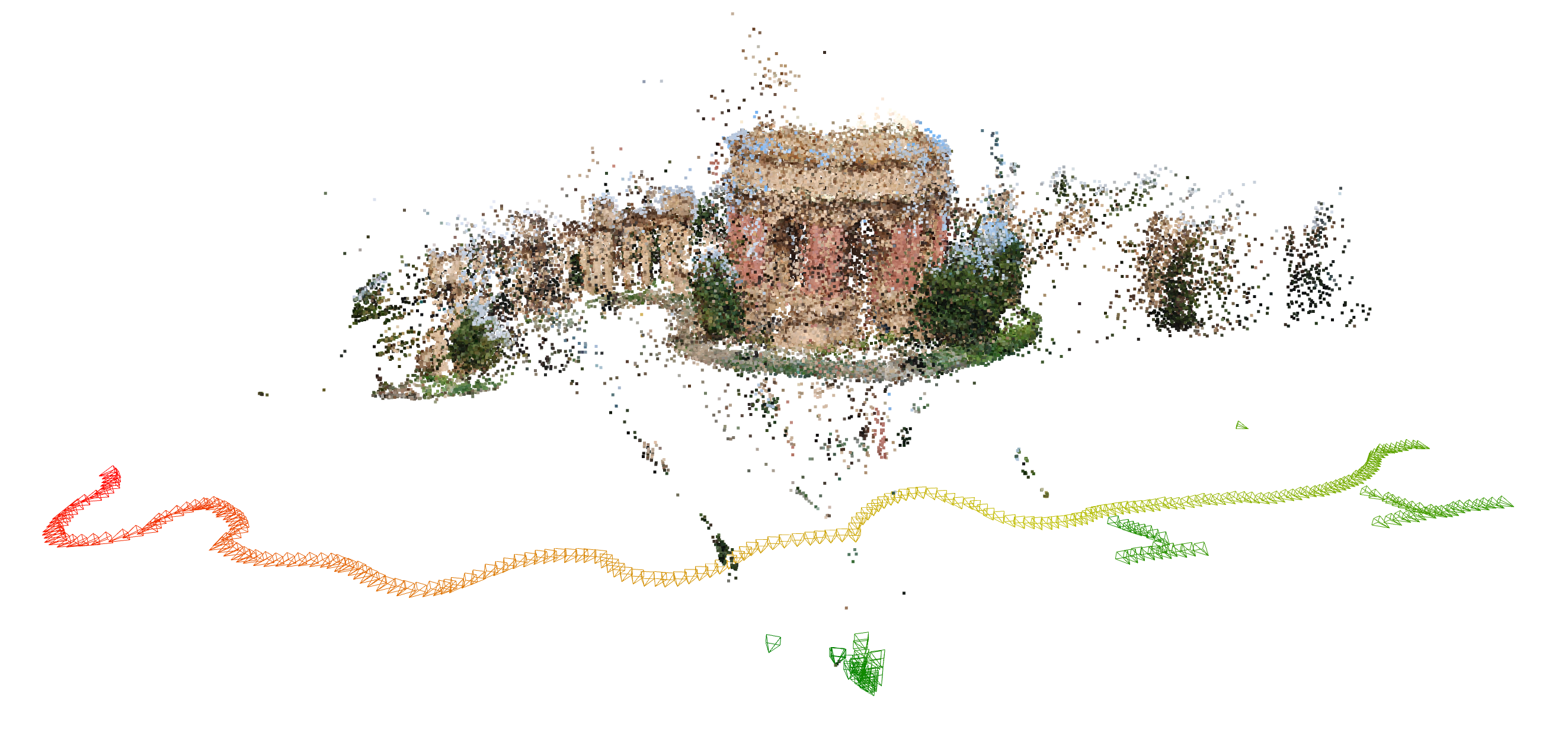}\\
	\includegraphics[width=0.16\linewidth]{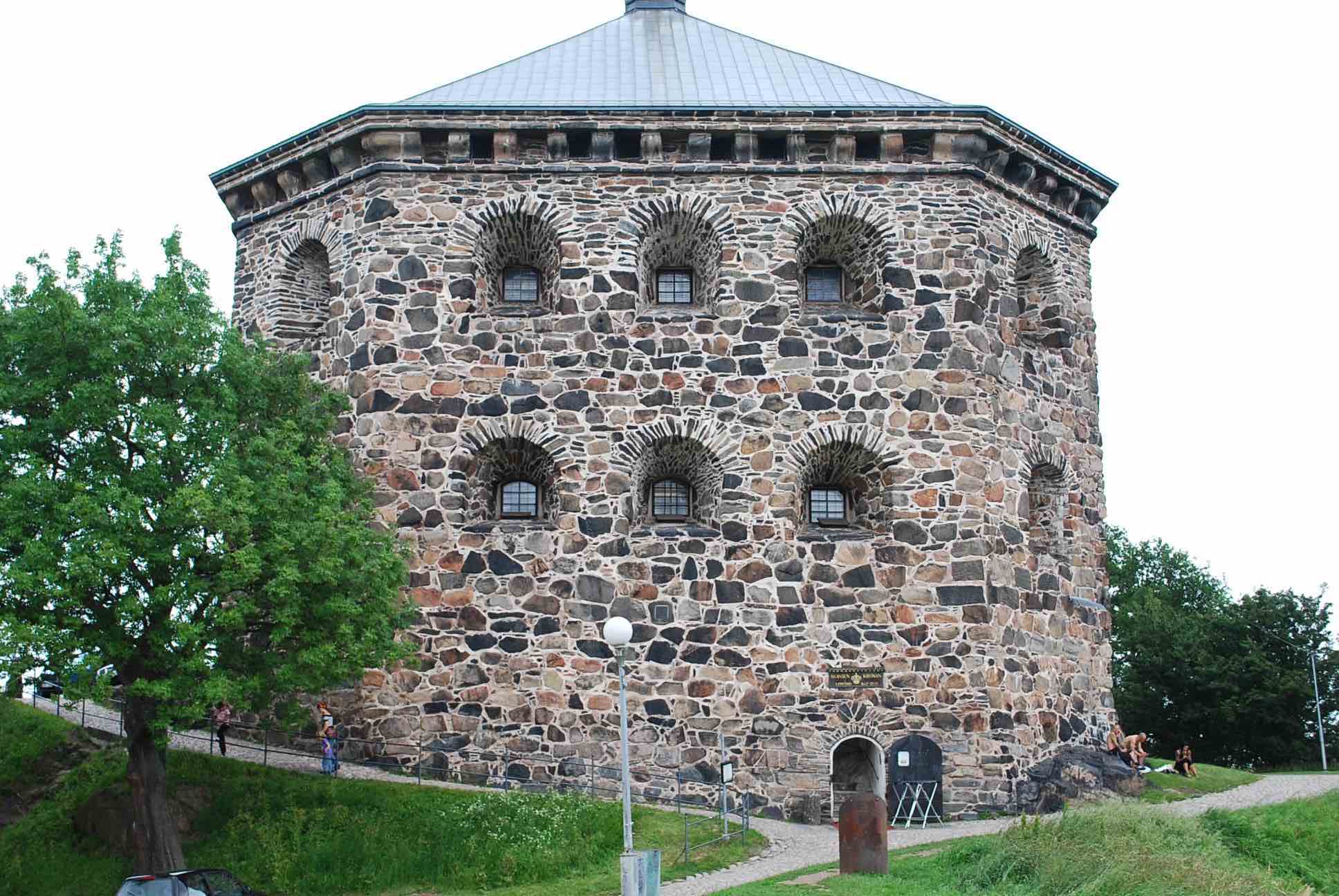}&
		\includegraphics[width=0.16\linewidth]{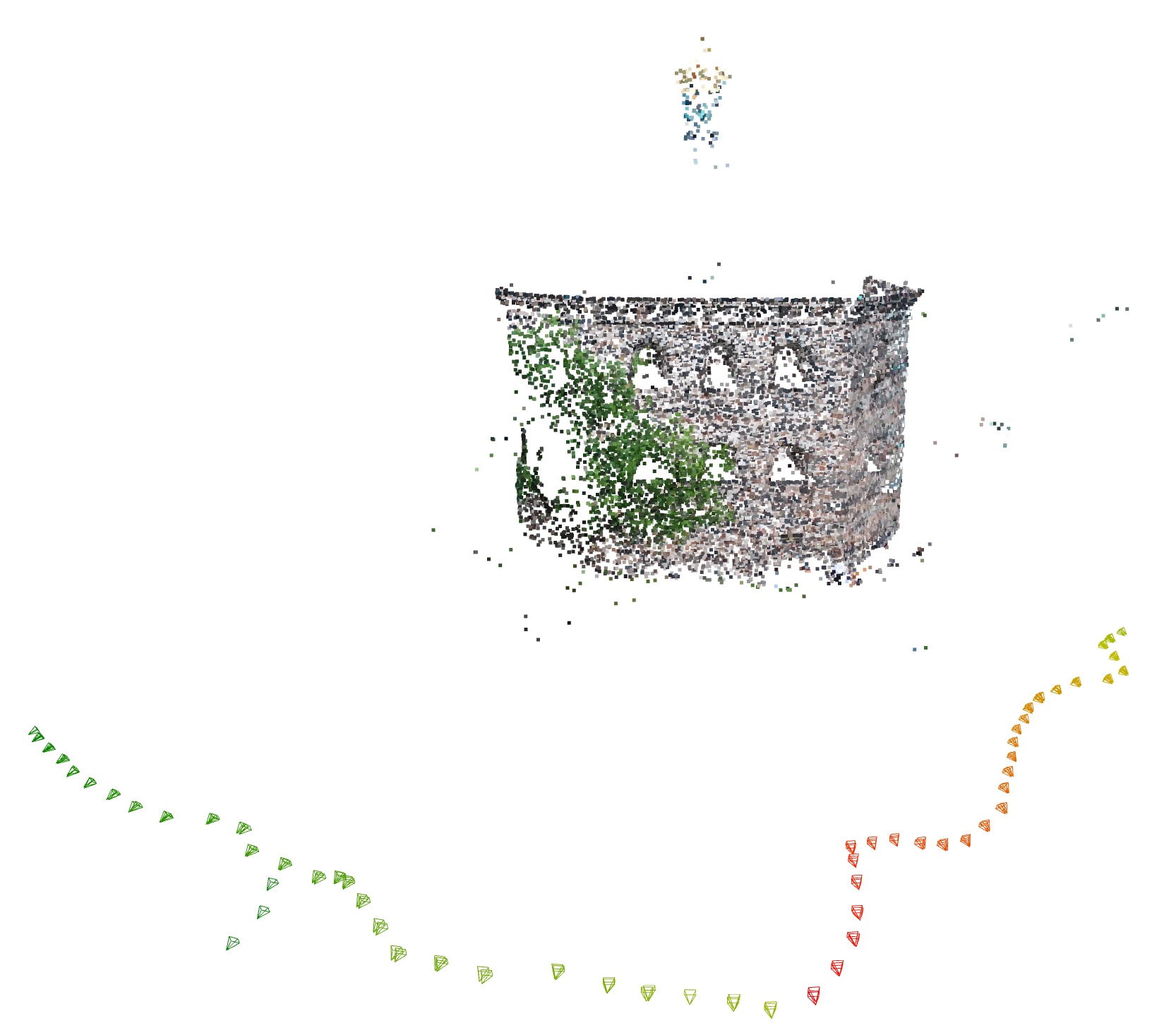}&
		\includegraphics[width=0.16\linewidth]{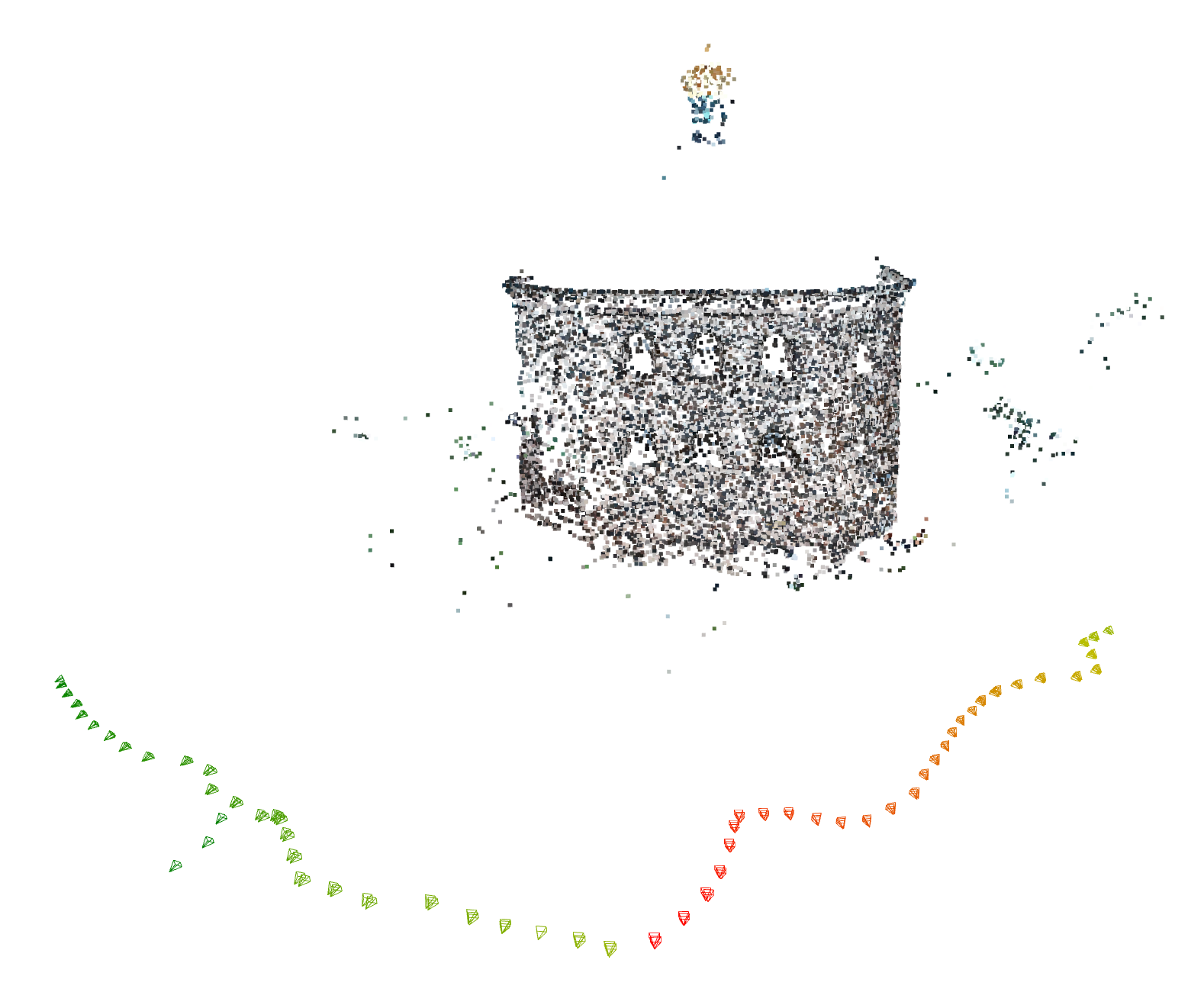}&
		\includegraphics[width=0.16\linewidth]{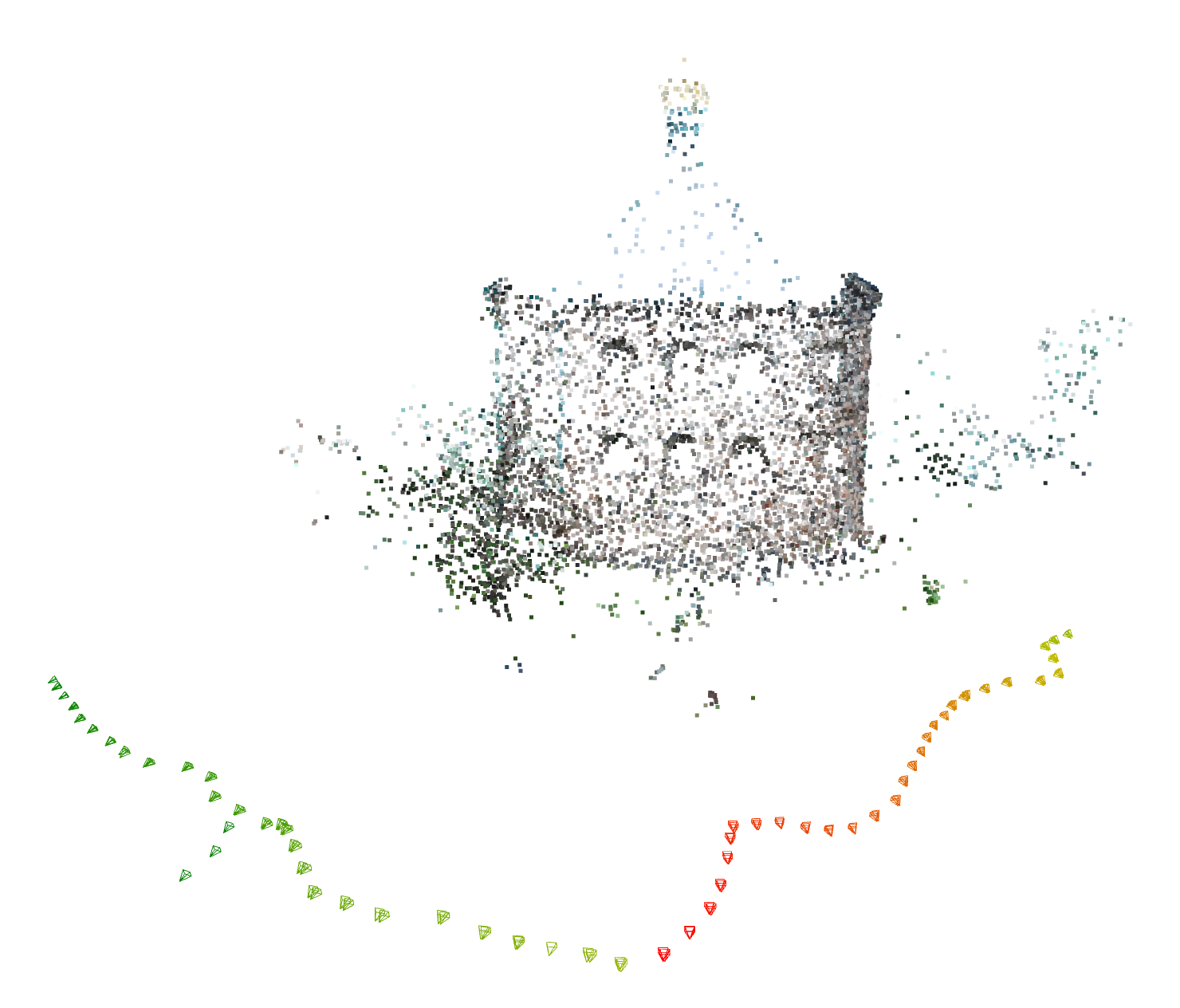}&
		\includegraphics[width=0.16\linewidth]{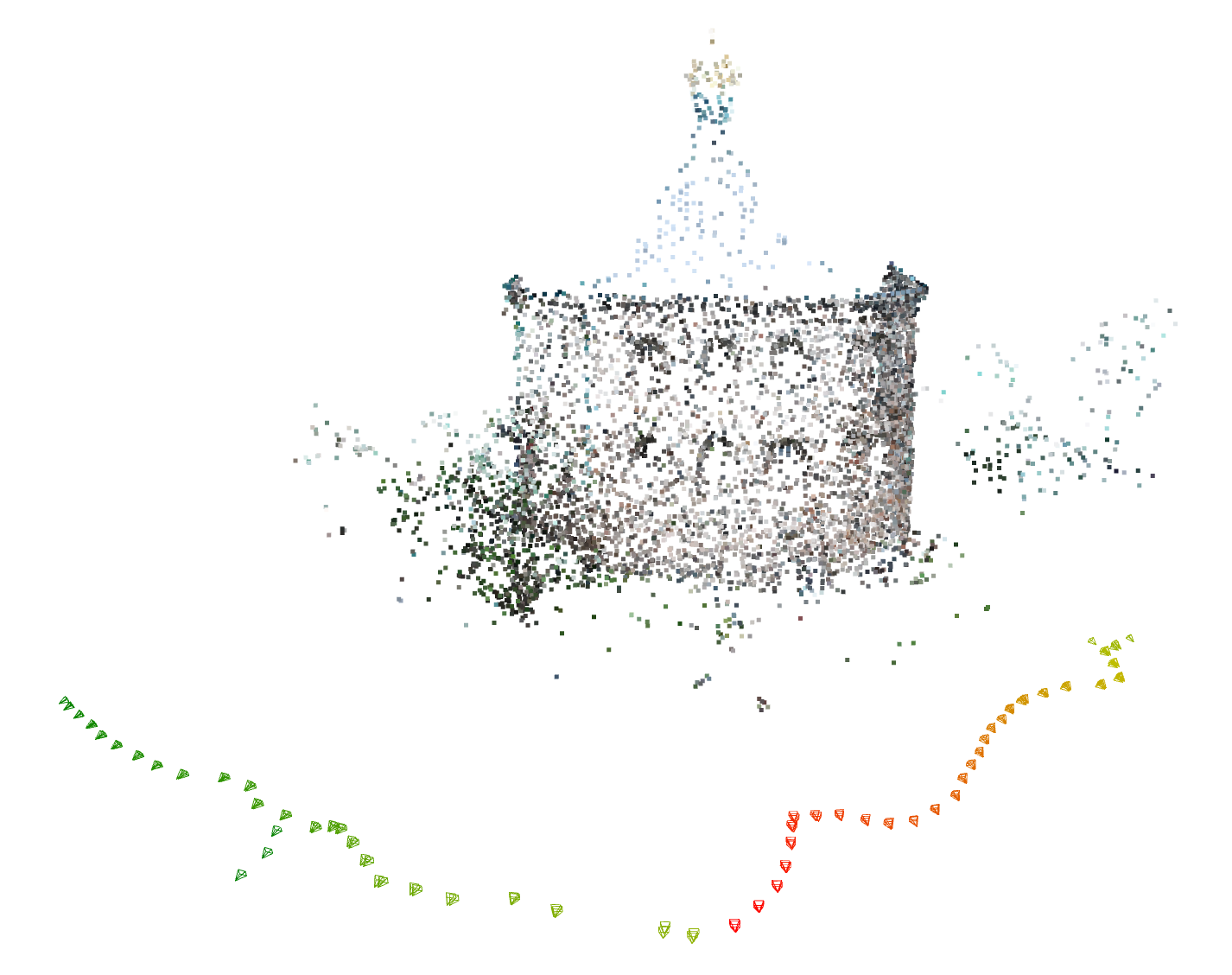}&
		\includegraphics[width=0.16\linewidth]{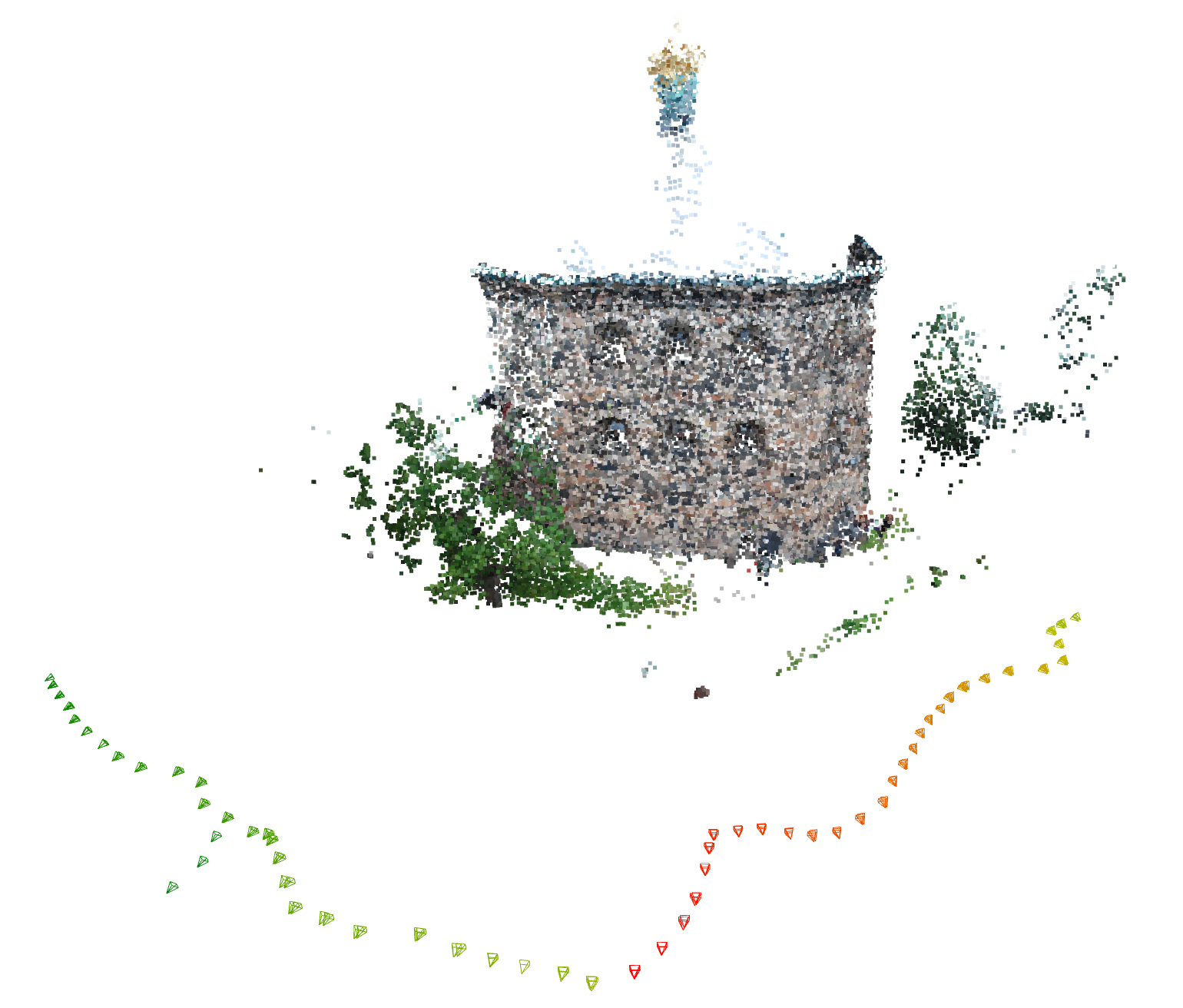}\\
		\includegraphics[width=0.16\linewidth]{figures/south_building_128/P1180221.jpeg}&
		\includegraphics[width=0.16\linewidth]{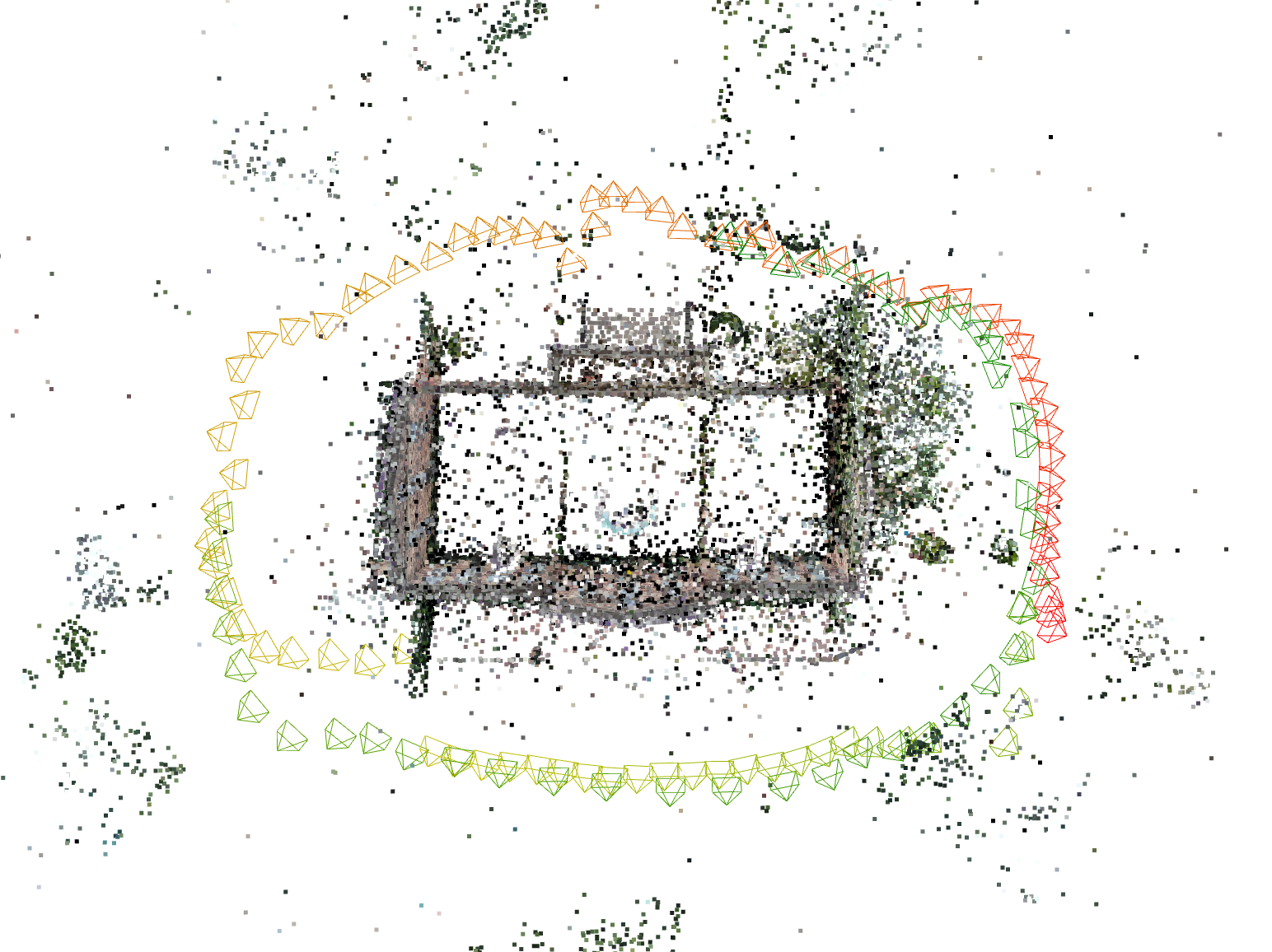}&
		\includegraphics[width=0.16\linewidth]{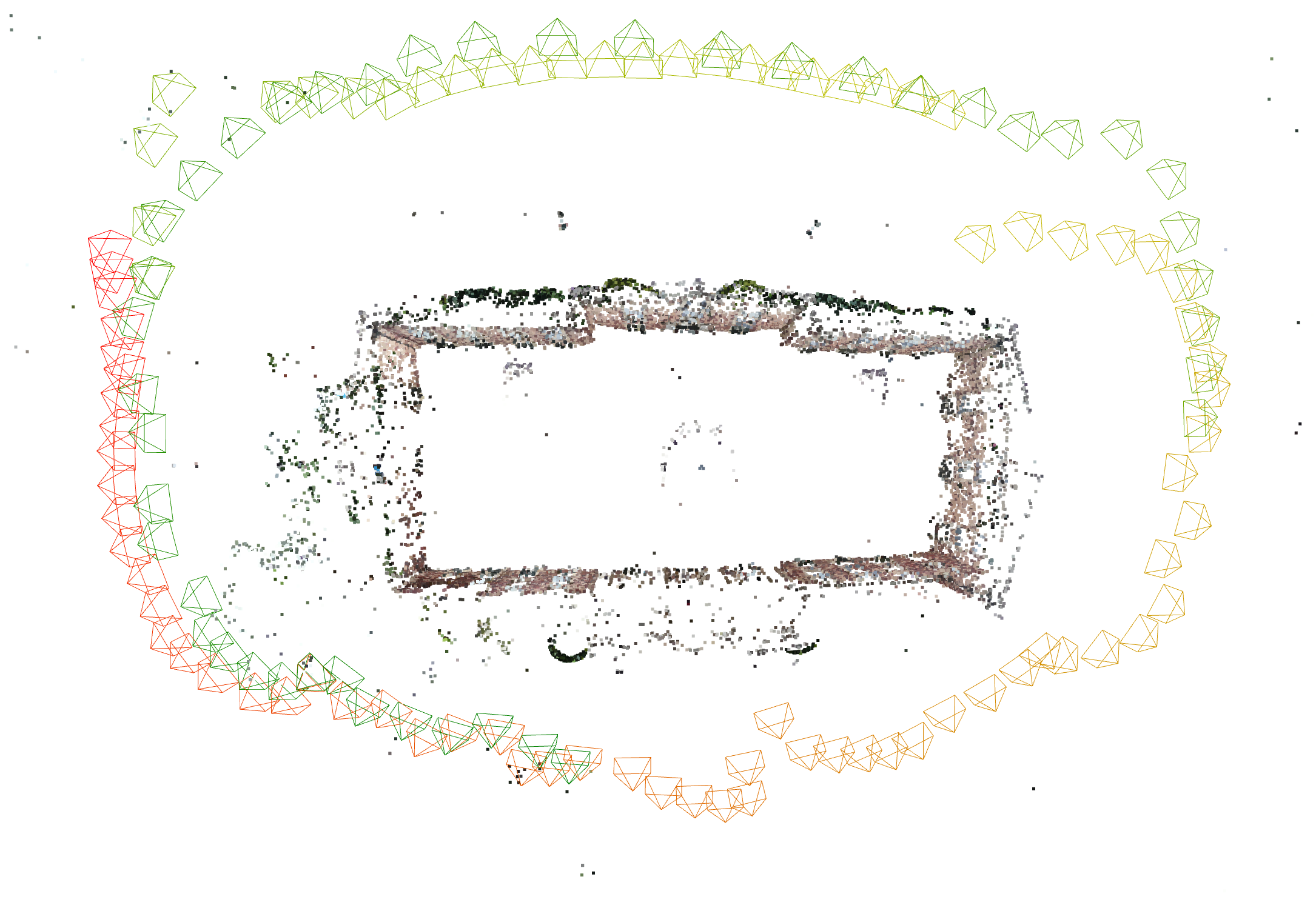}&
		\includegraphics[width=0.16\linewidth]{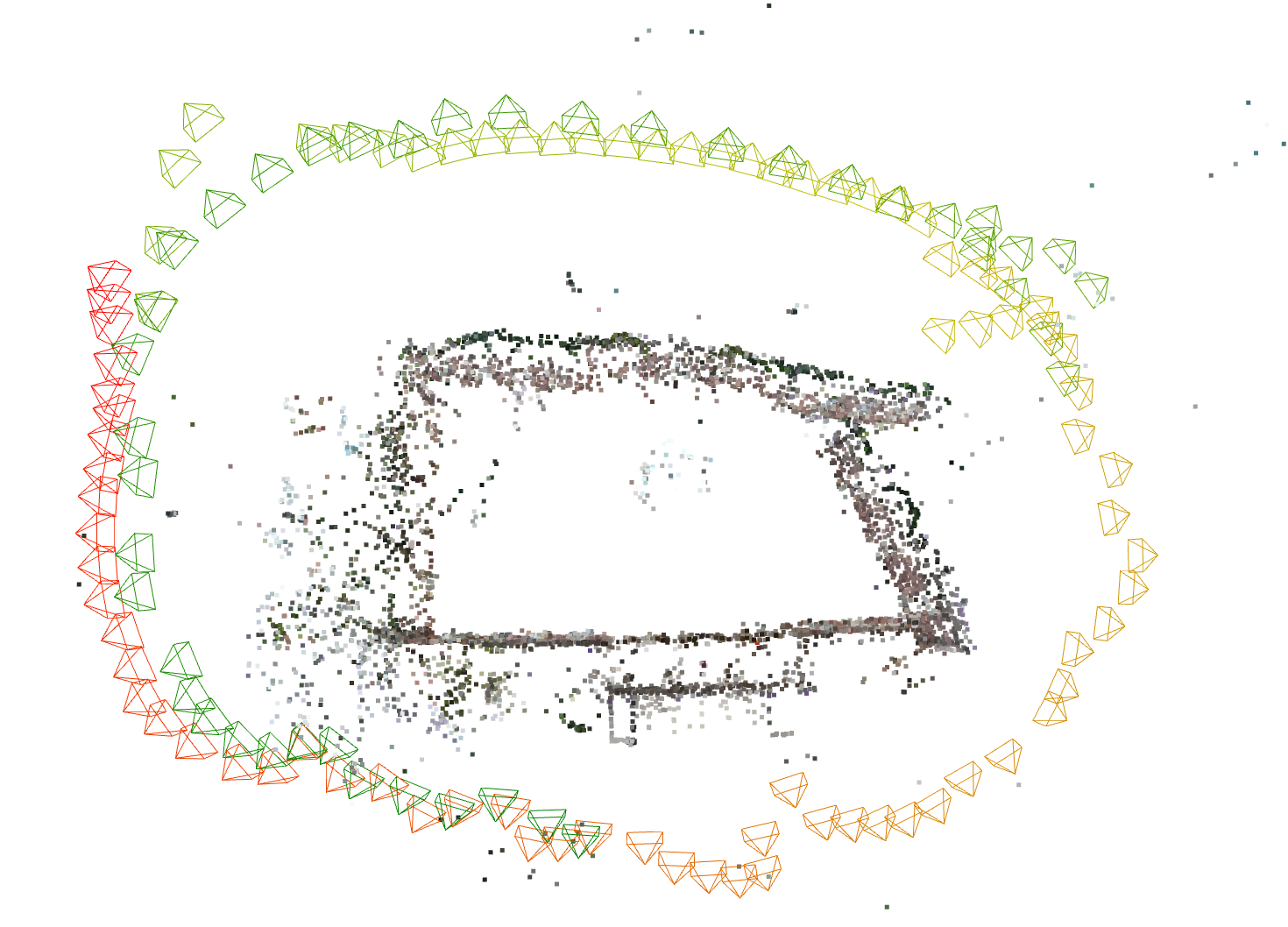}&
		\includegraphics[width=0.16\linewidth]{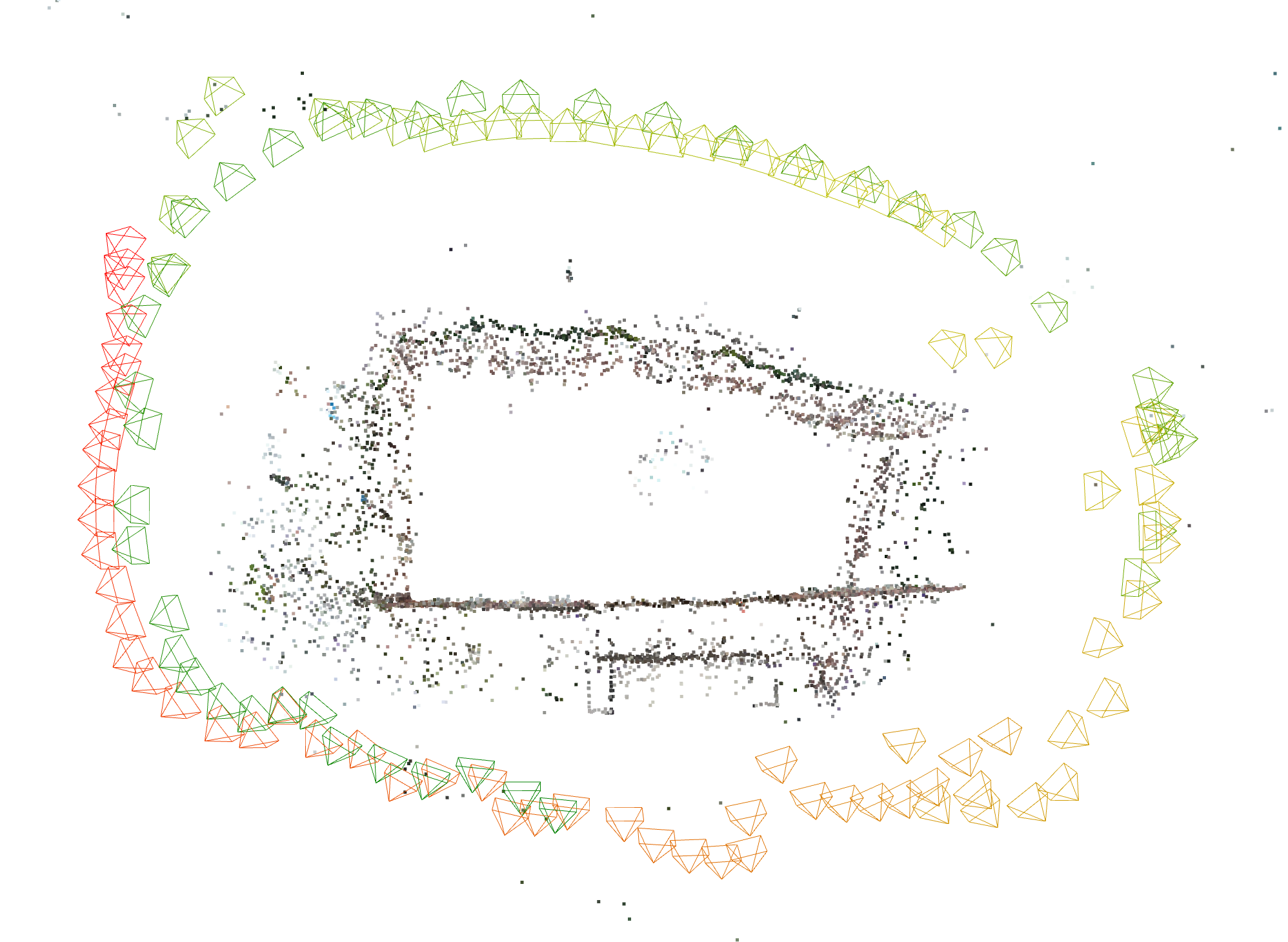}&
		\includegraphics[width=0.16\linewidth]{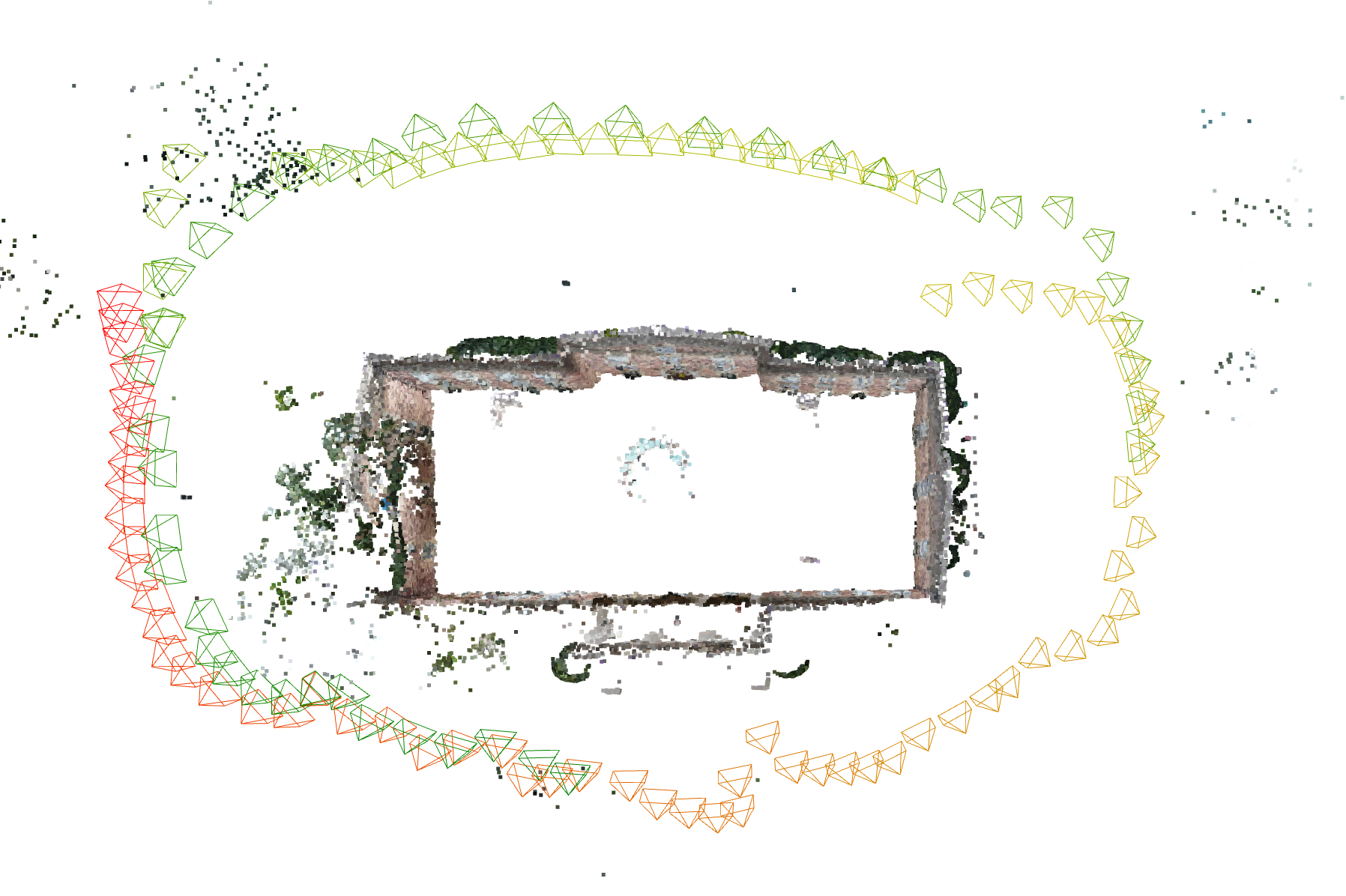}\\
		\includegraphics[width=0.16\linewidth, height=18.5mm]{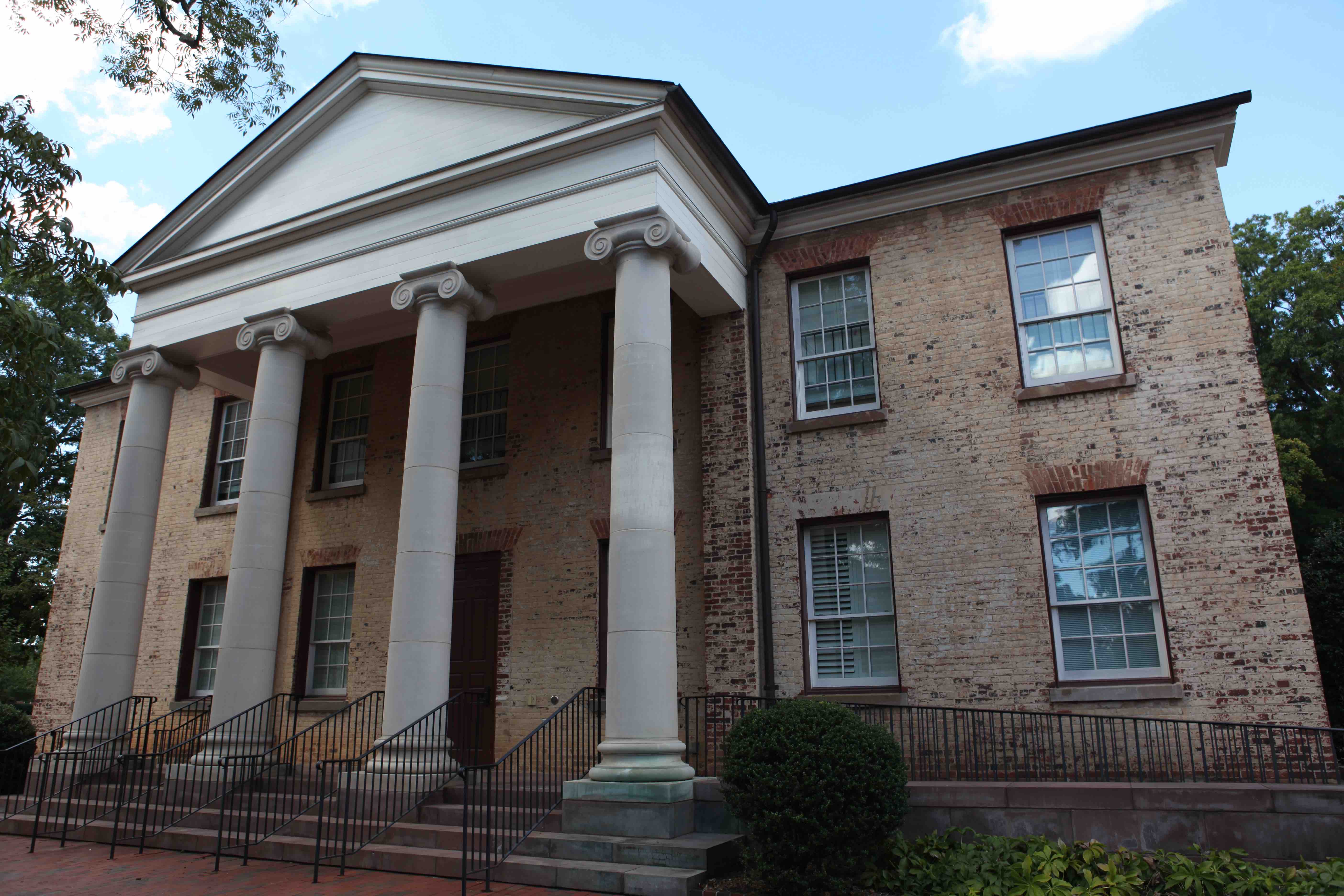}&
		\includegraphics[width=0.16\linewidth, height=18.5mm]{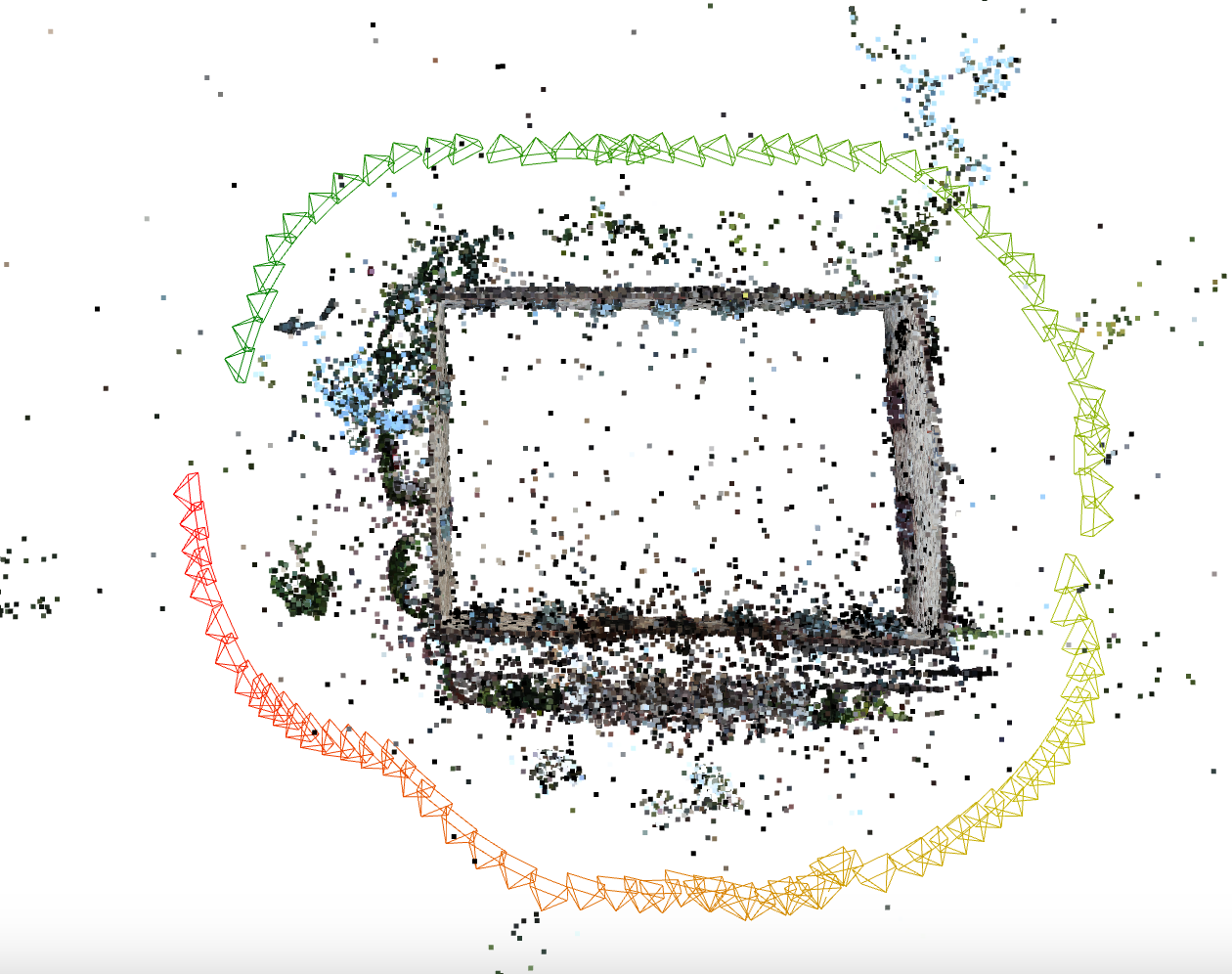}&
		\includegraphics[width=0.16\linewidth, height=18.5mm]{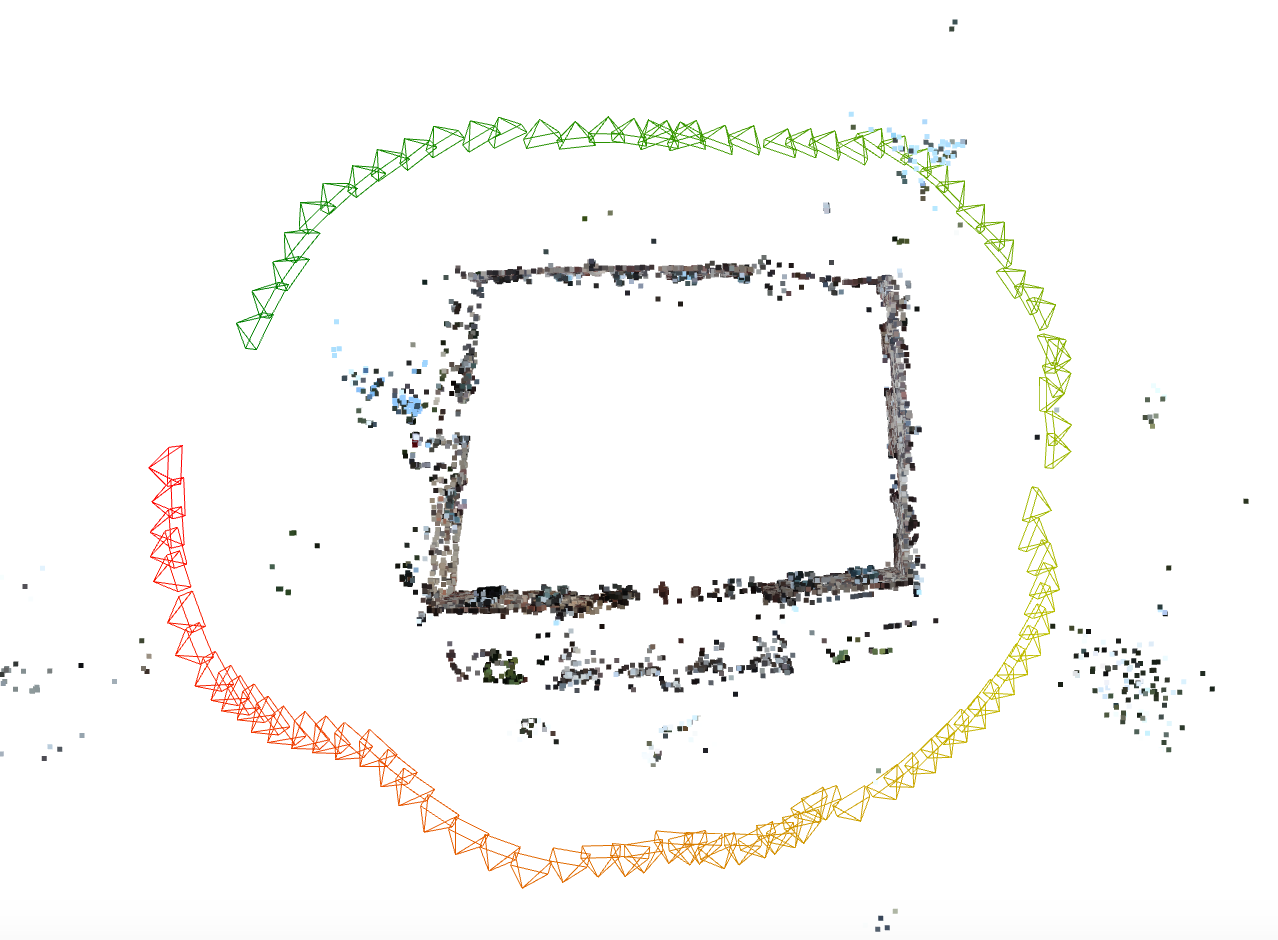}&
		\includegraphics[width=0.16\linewidth, height=18.5mm]{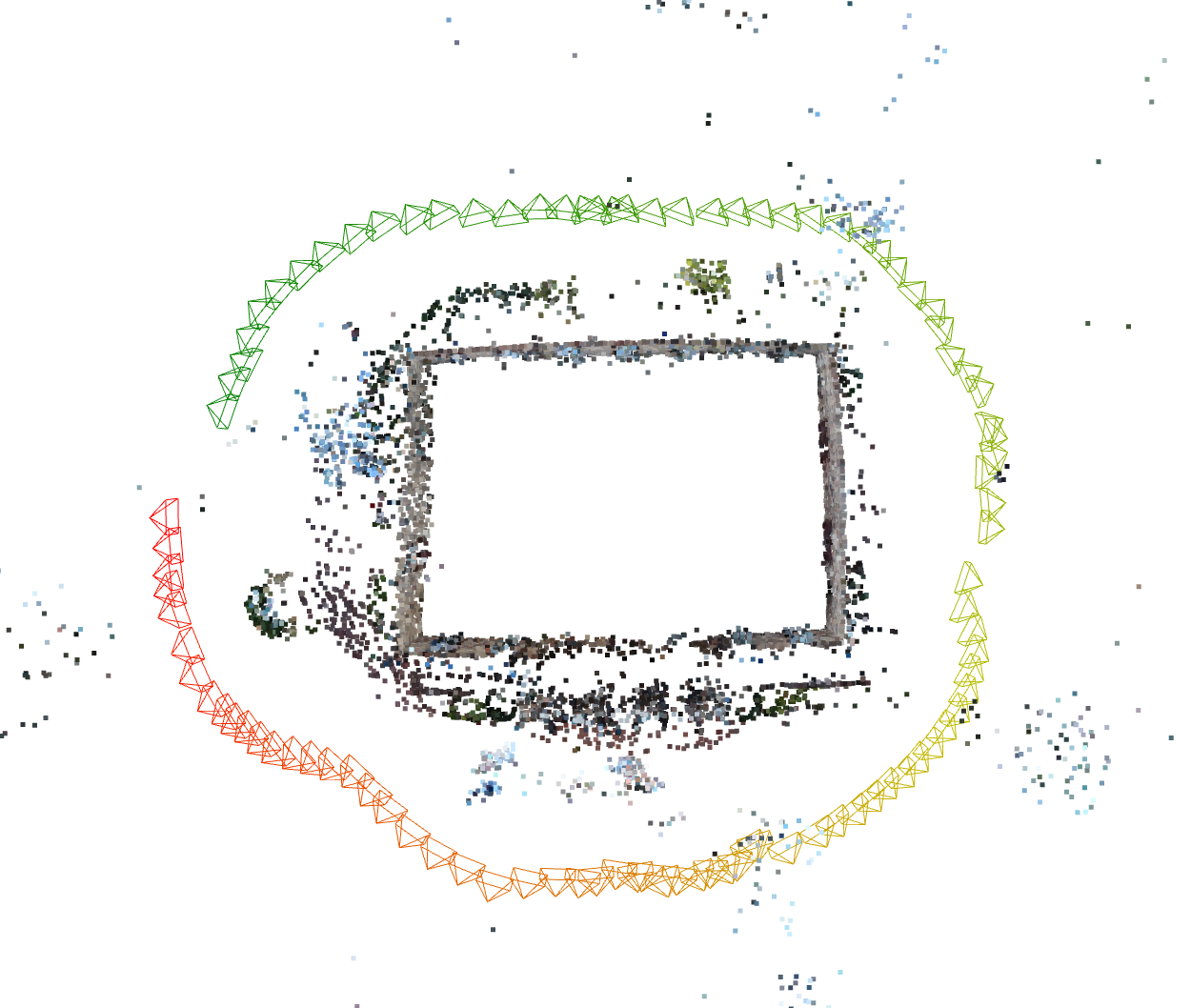}&
		\includegraphics[width=0.16\linewidth, height=18.5mm]{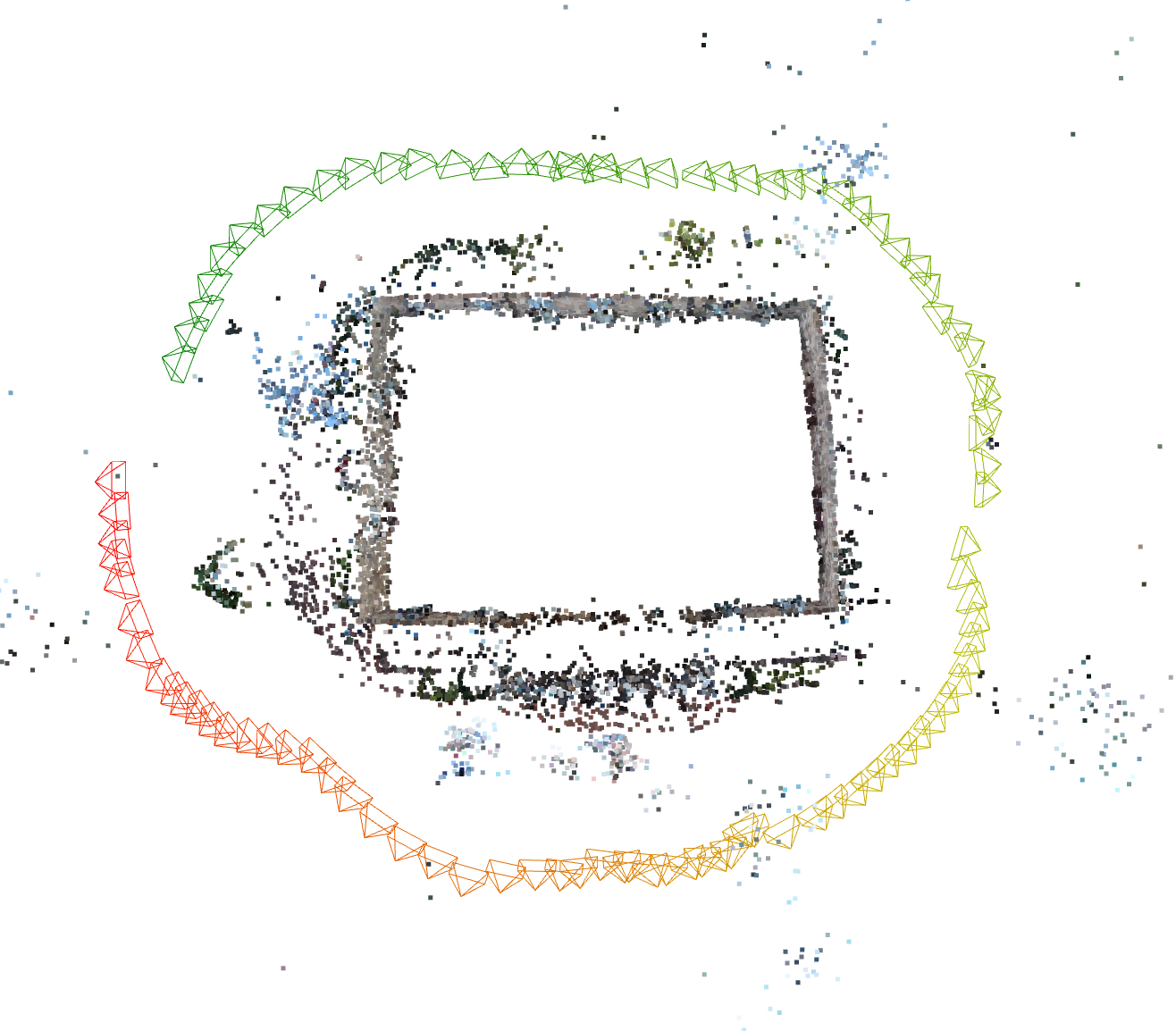}&
		\includegraphics[width=0.16\linewidth, height=18.5mm]{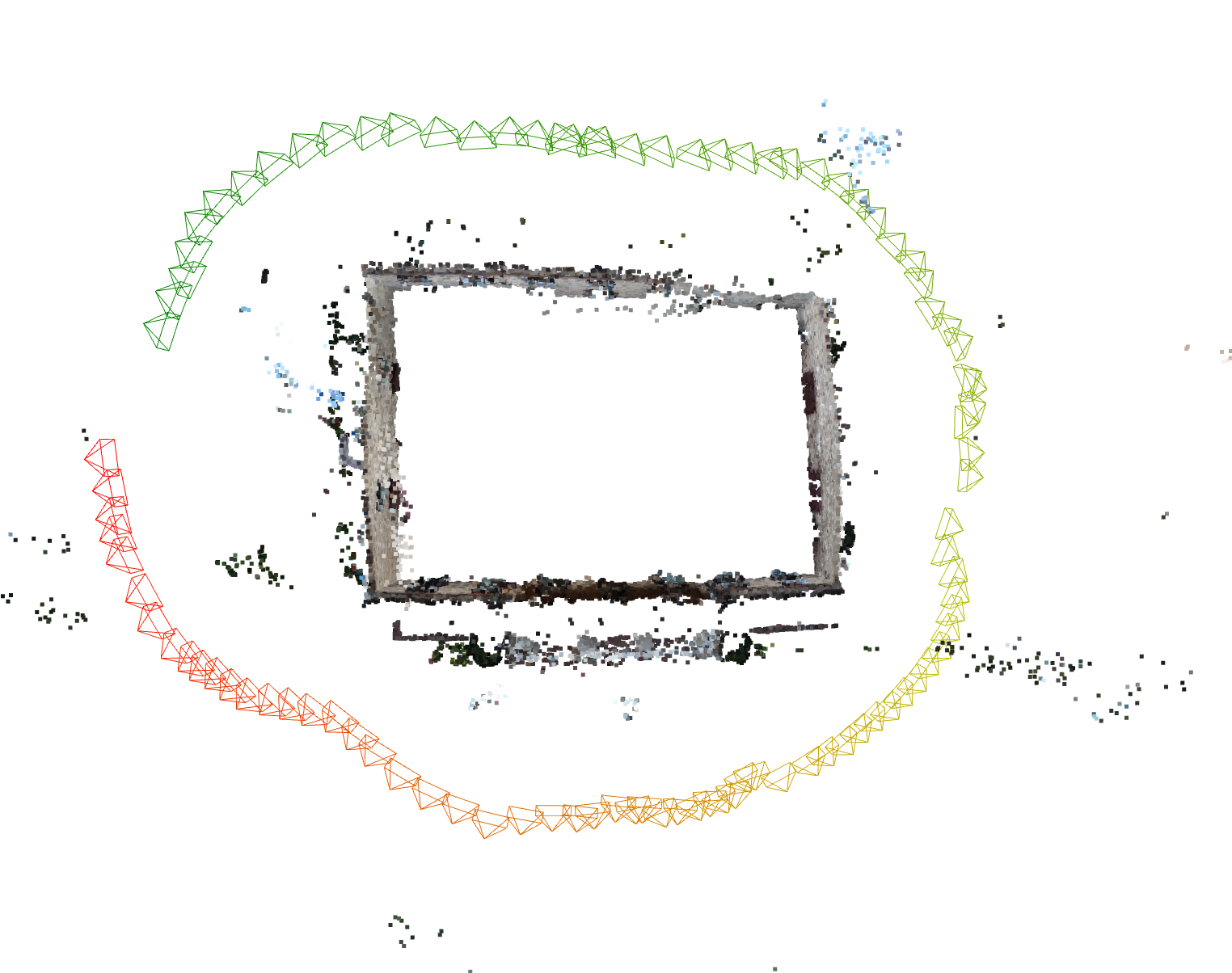}\\
    \includegraphics[width=0.16\linewidth, height=18.5mm]{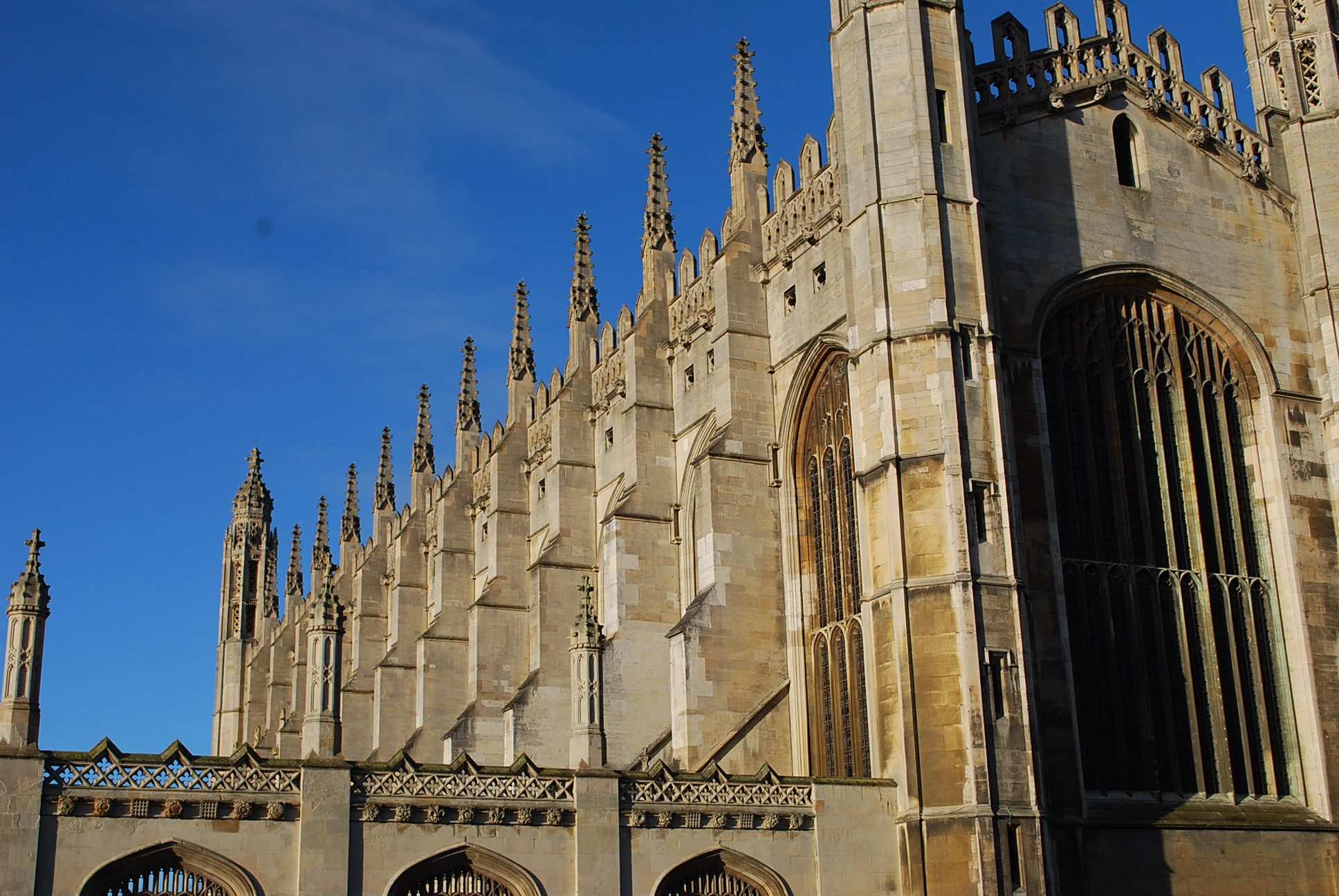}&
    \includegraphics[width=0.16\linewidth, height=18.5mm]{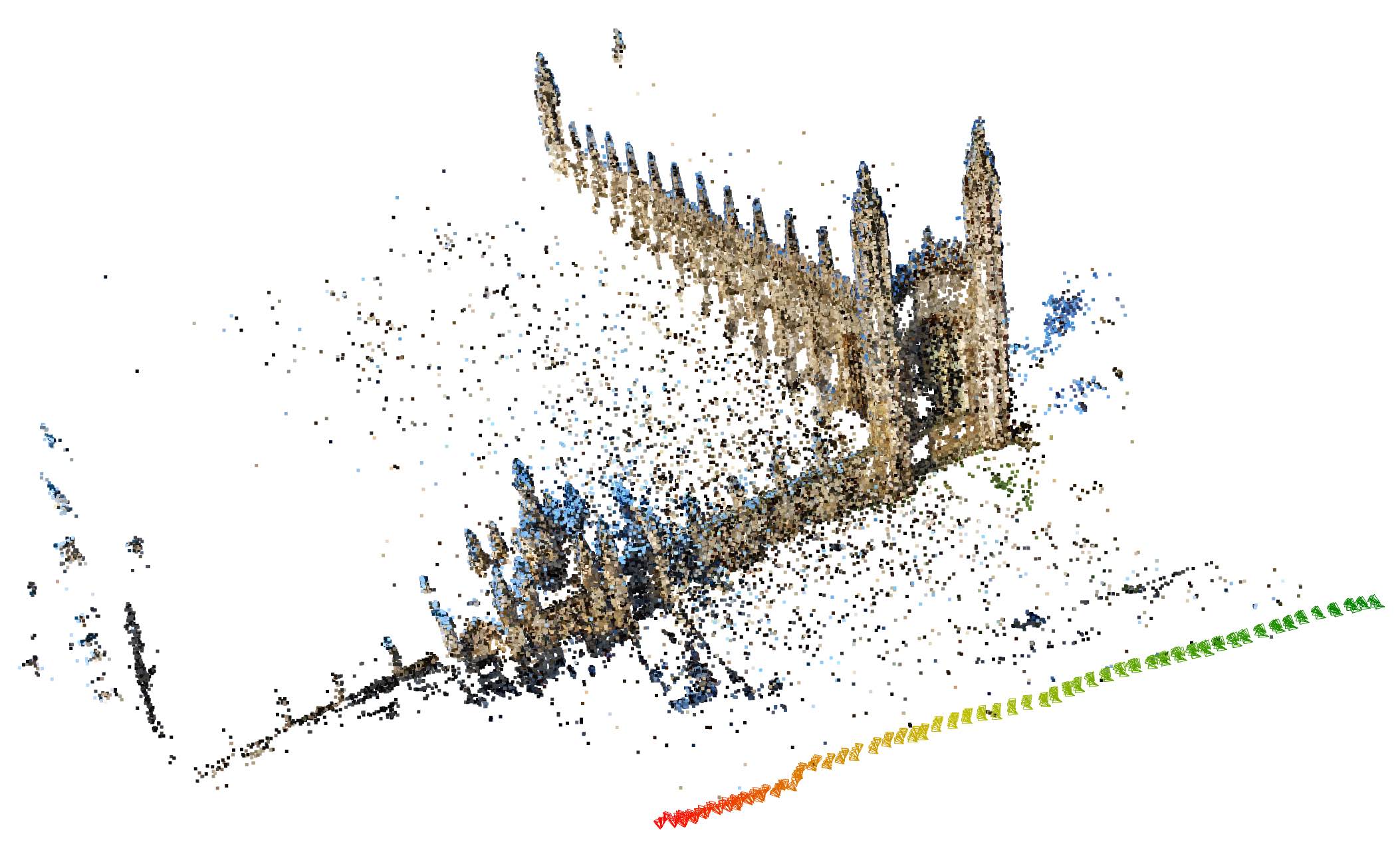}&
    \includegraphics[width=0.16\linewidth, height=18.5mm]{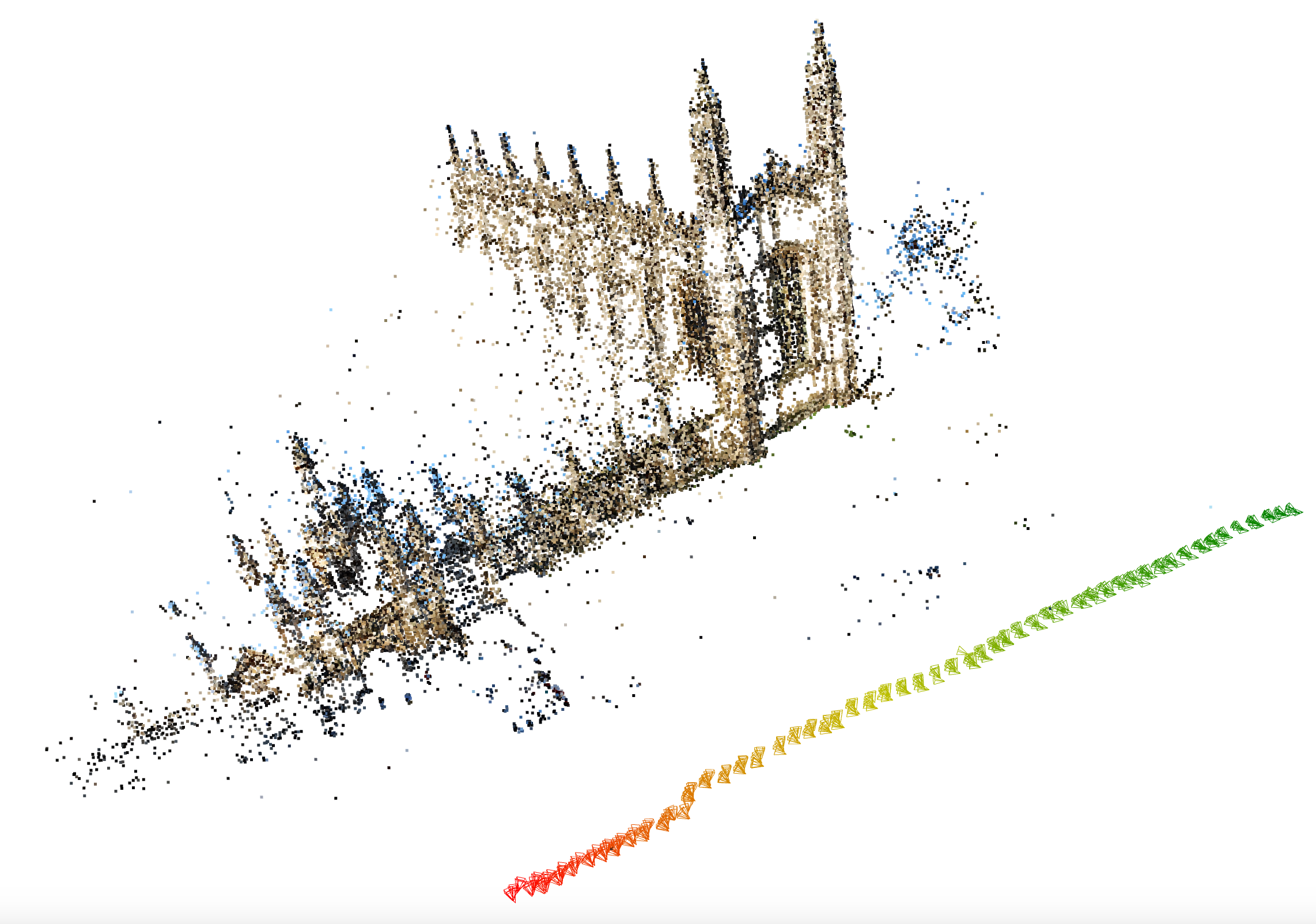}&
    \includegraphics[width=0.16\linewidth, height=18.5mm]{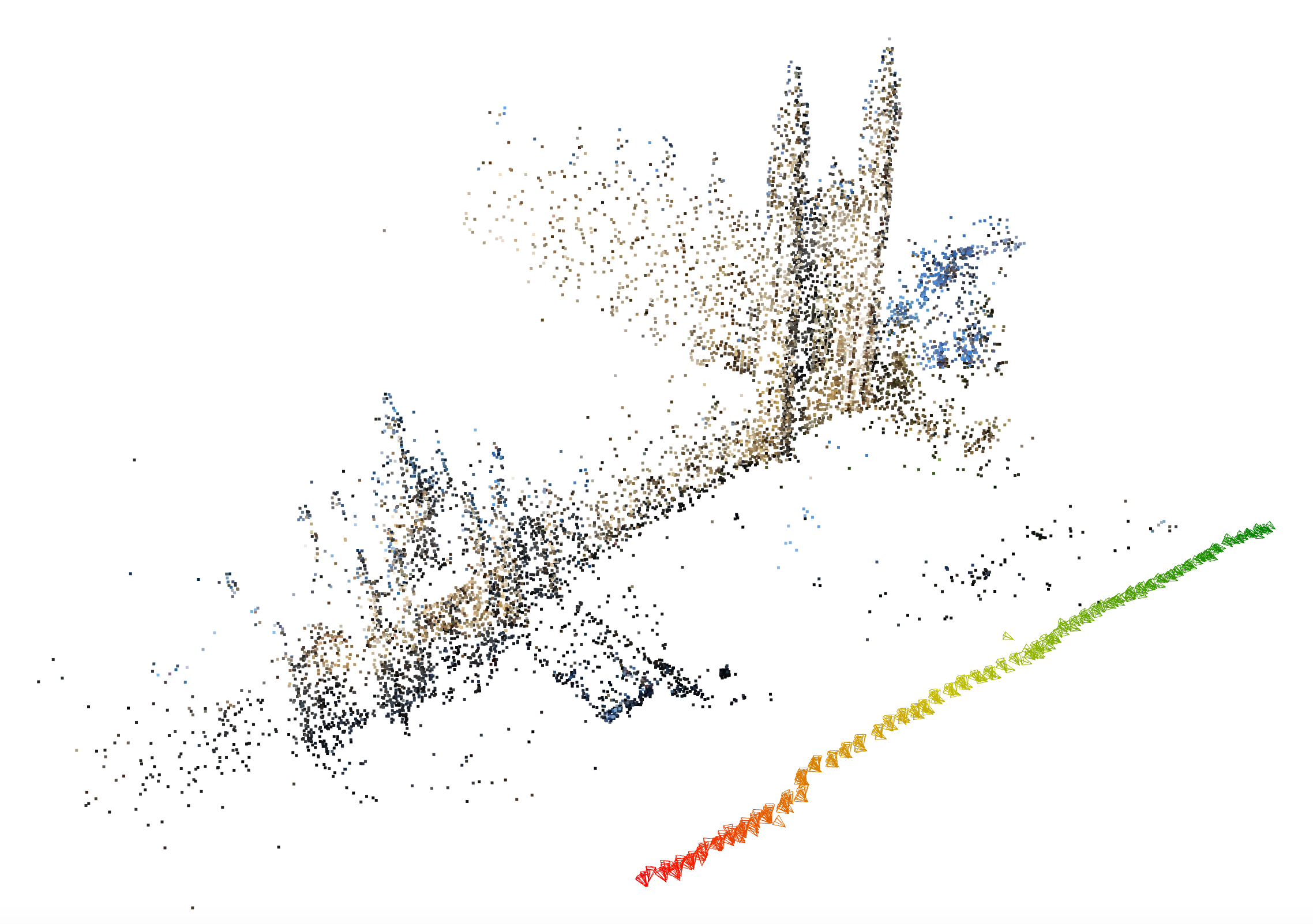}&
    \includegraphics[width=0.16\linewidth, height=18.5mm]{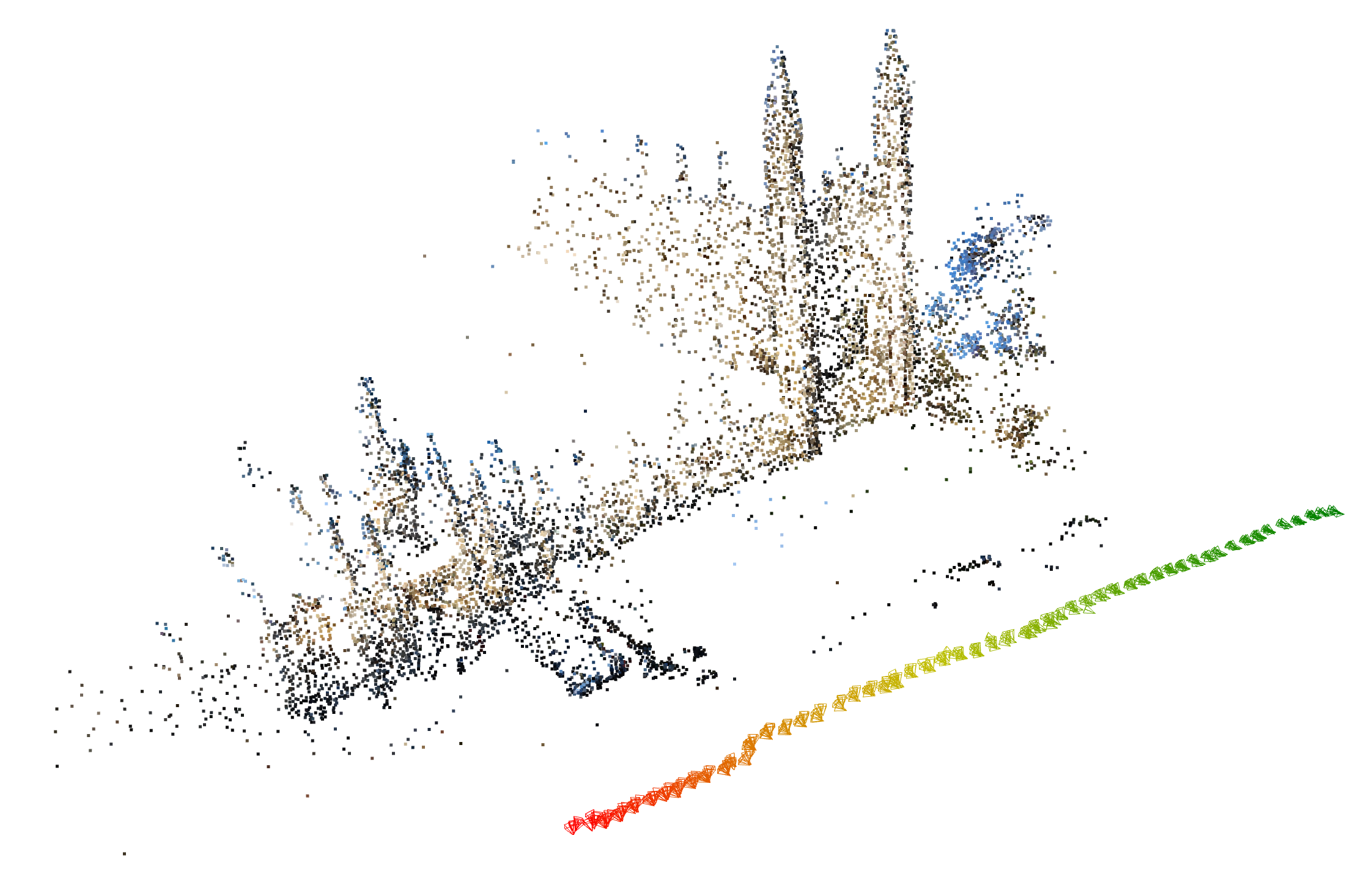}&
    \includegraphics[width=0.16\linewidth, height=18.5mm]{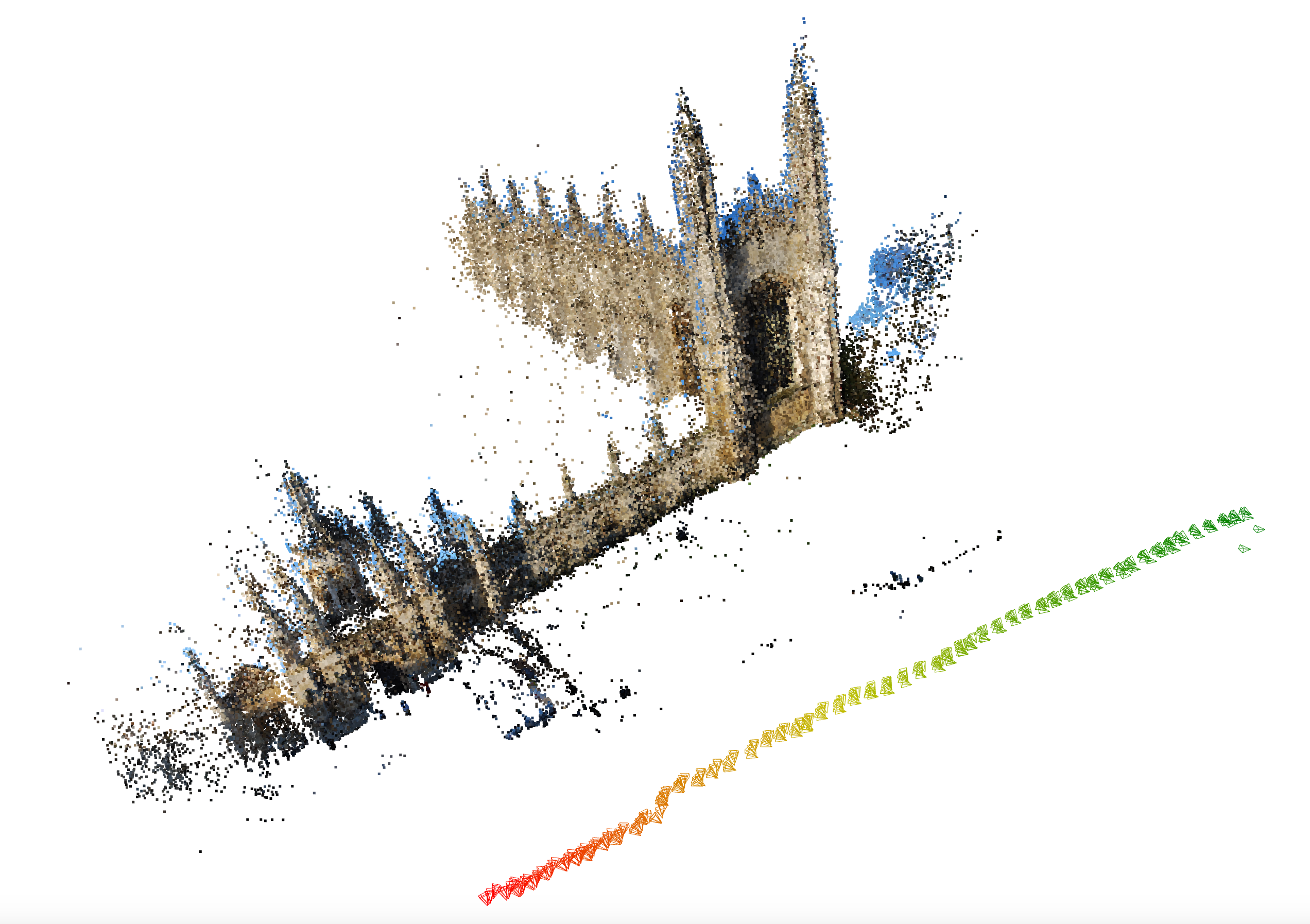}\\
    \includegraphics[angle=90,width=0.16\linewidth]{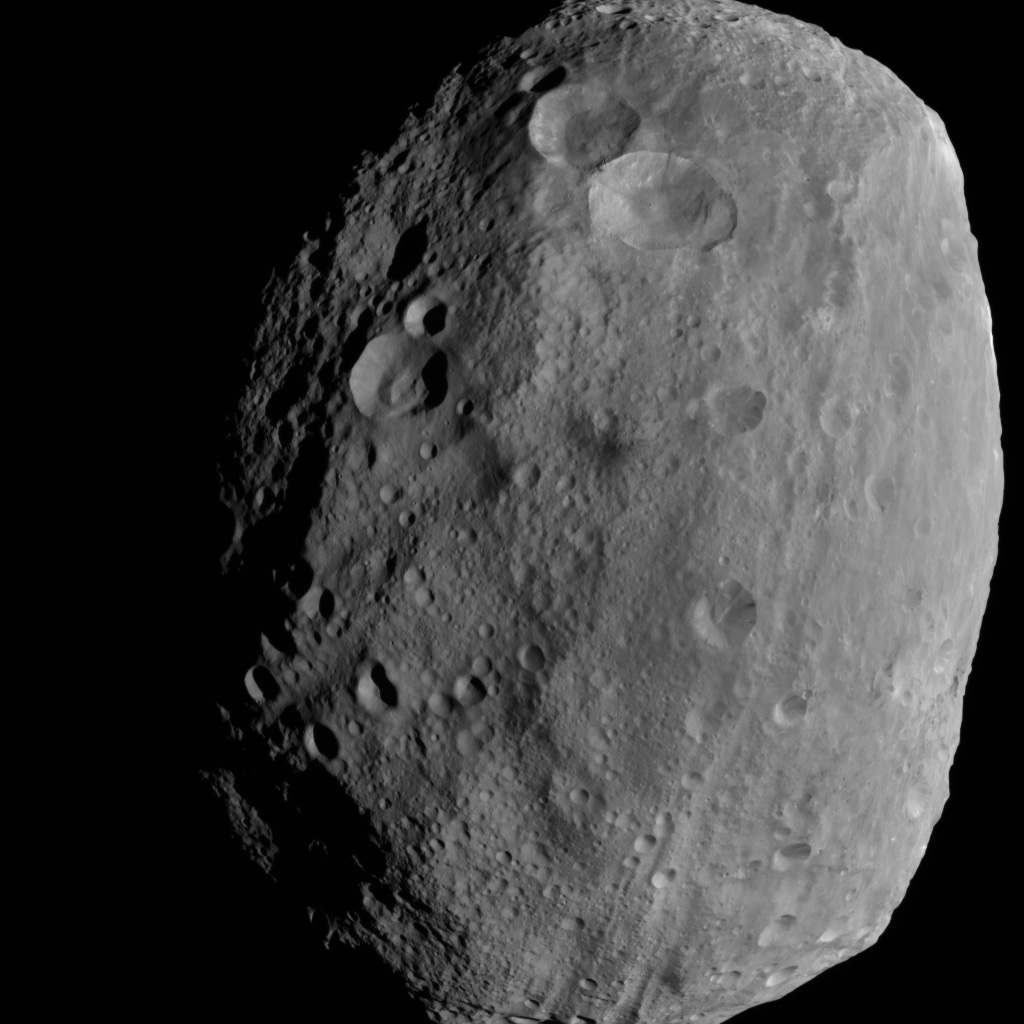}&
    \includegraphics[width=0.16\linewidth]{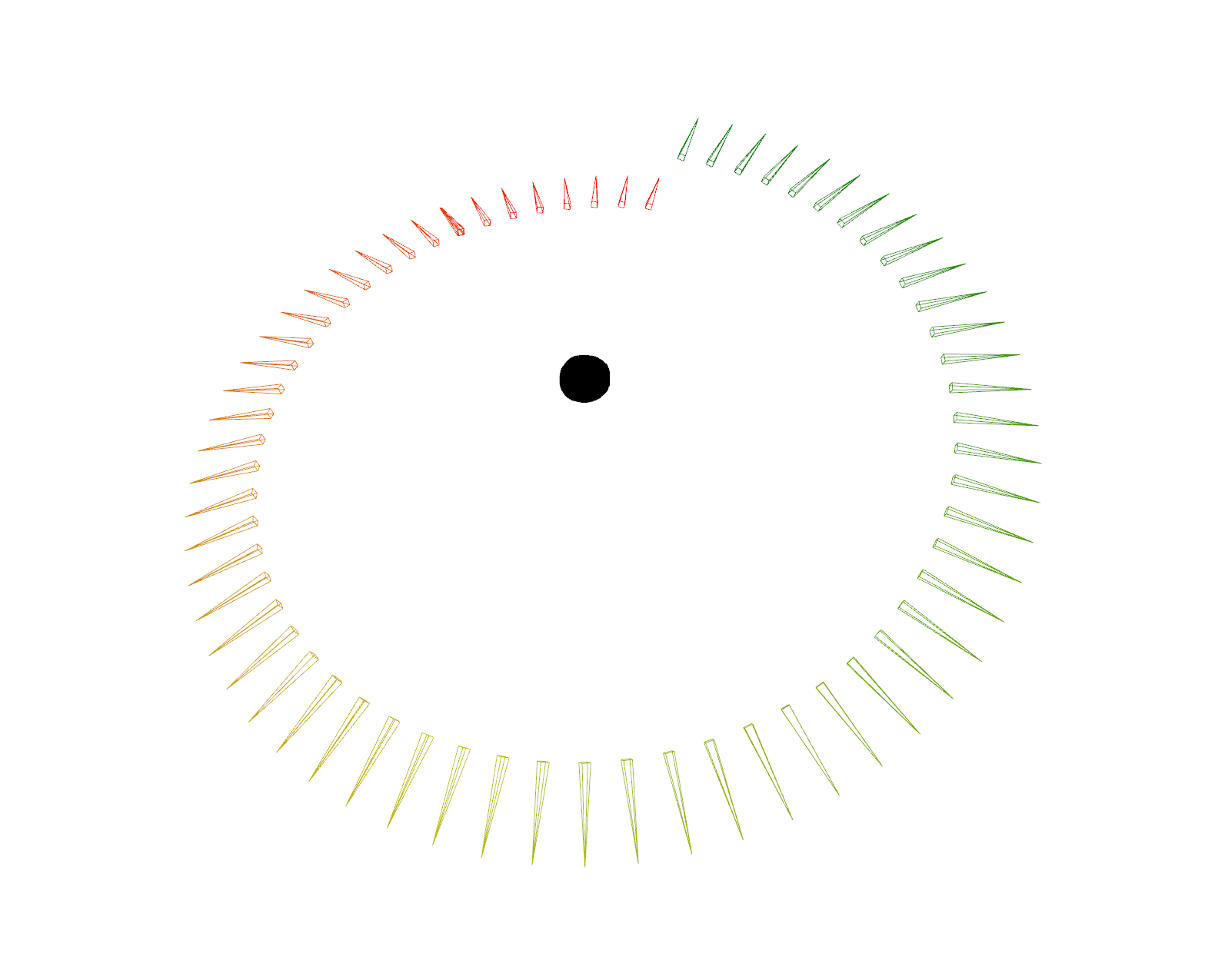}&
    \includegraphics[width=0.16\linewidth]{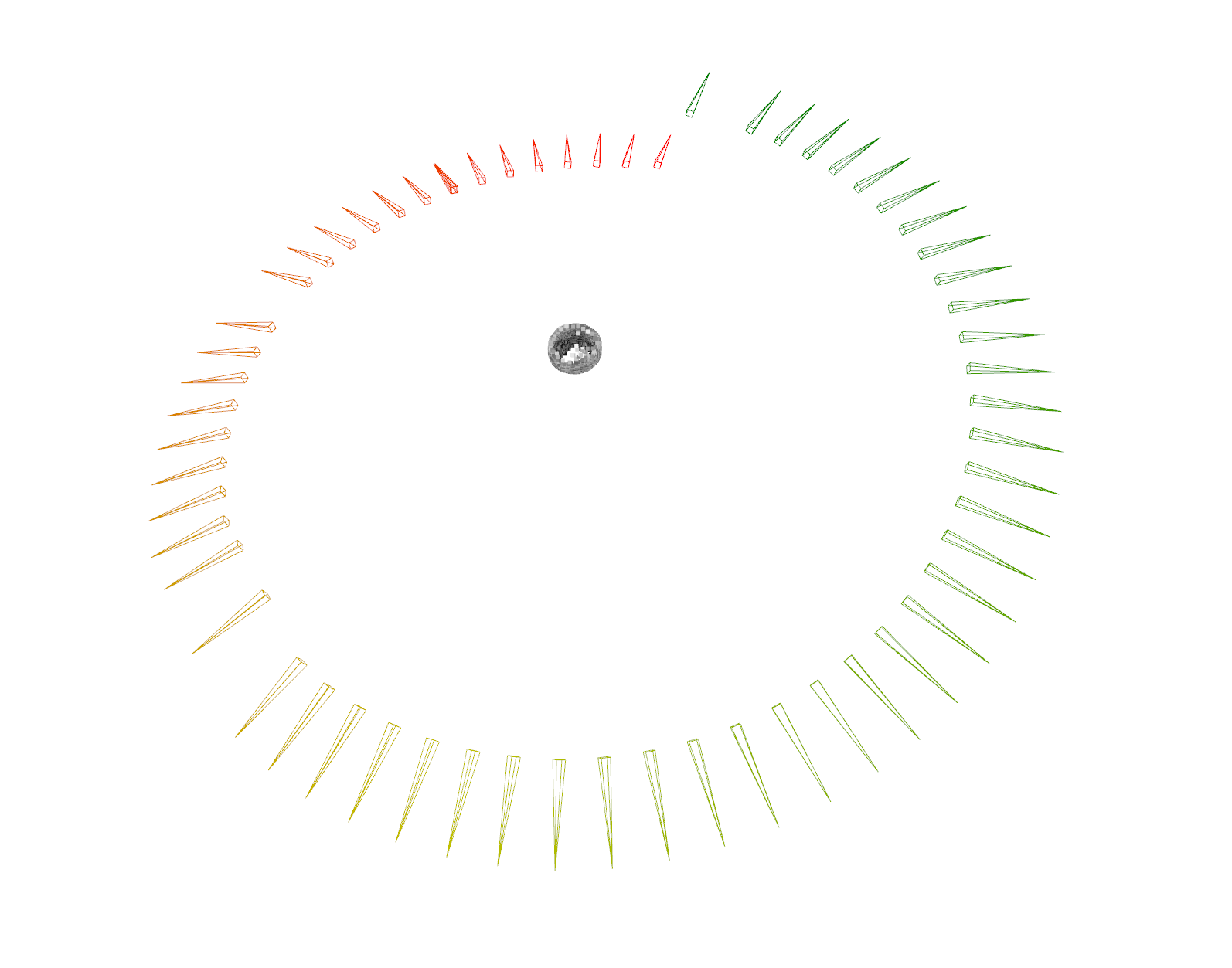}&
    \includegraphics[width=0.16\linewidth]{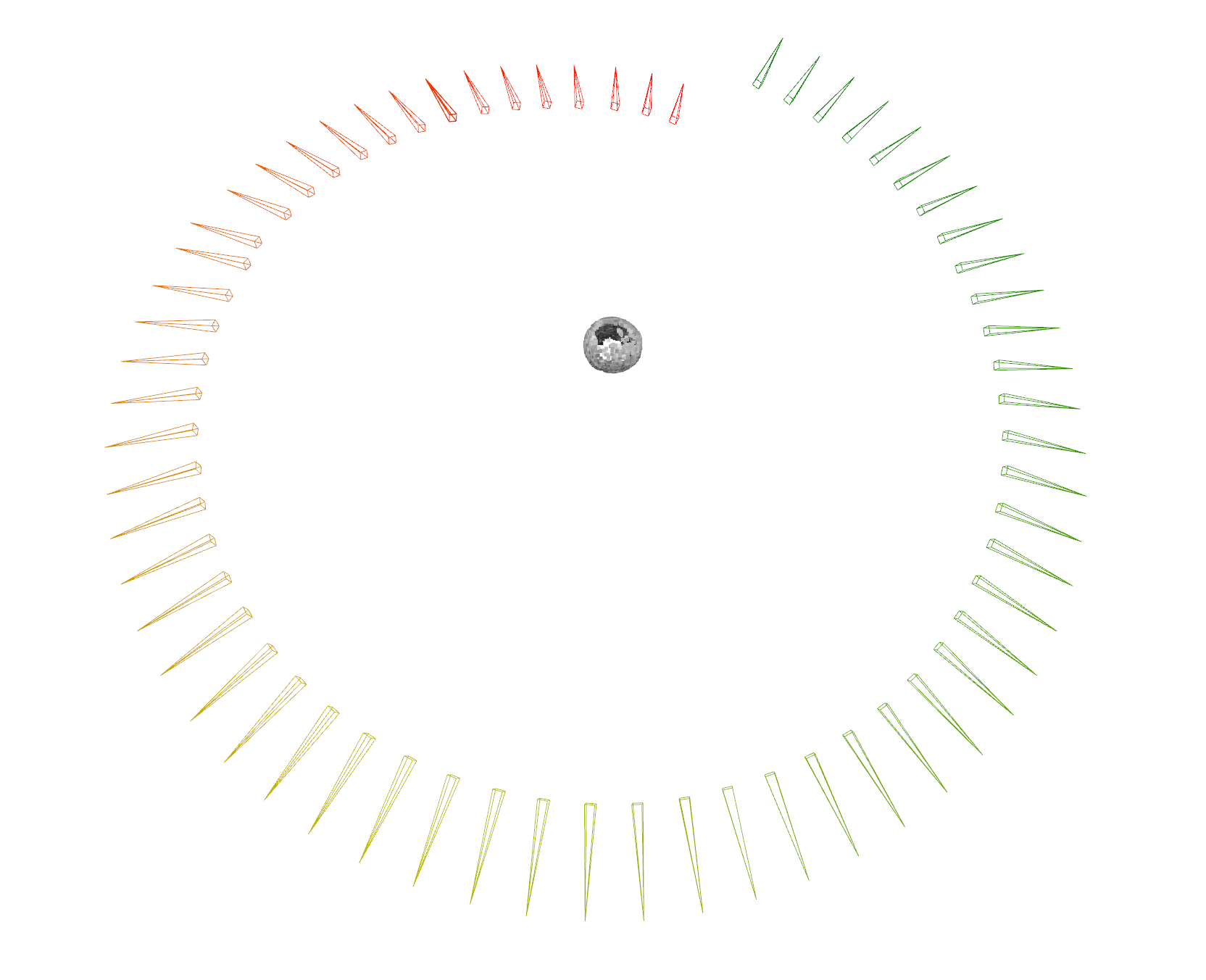}&
    \includegraphics[width=0.16\linewidth]{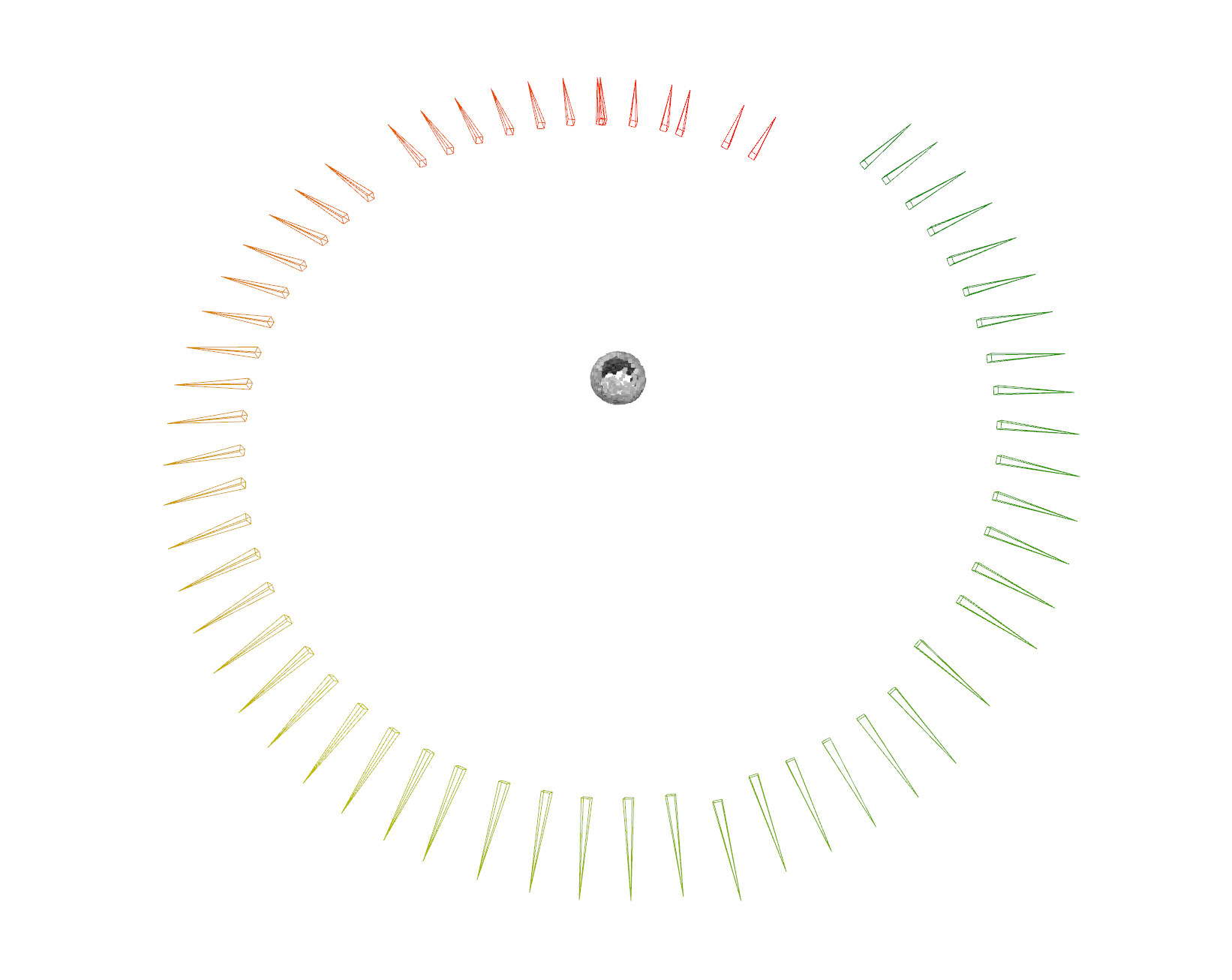}&
    \includegraphics[width=0.16\linewidth]{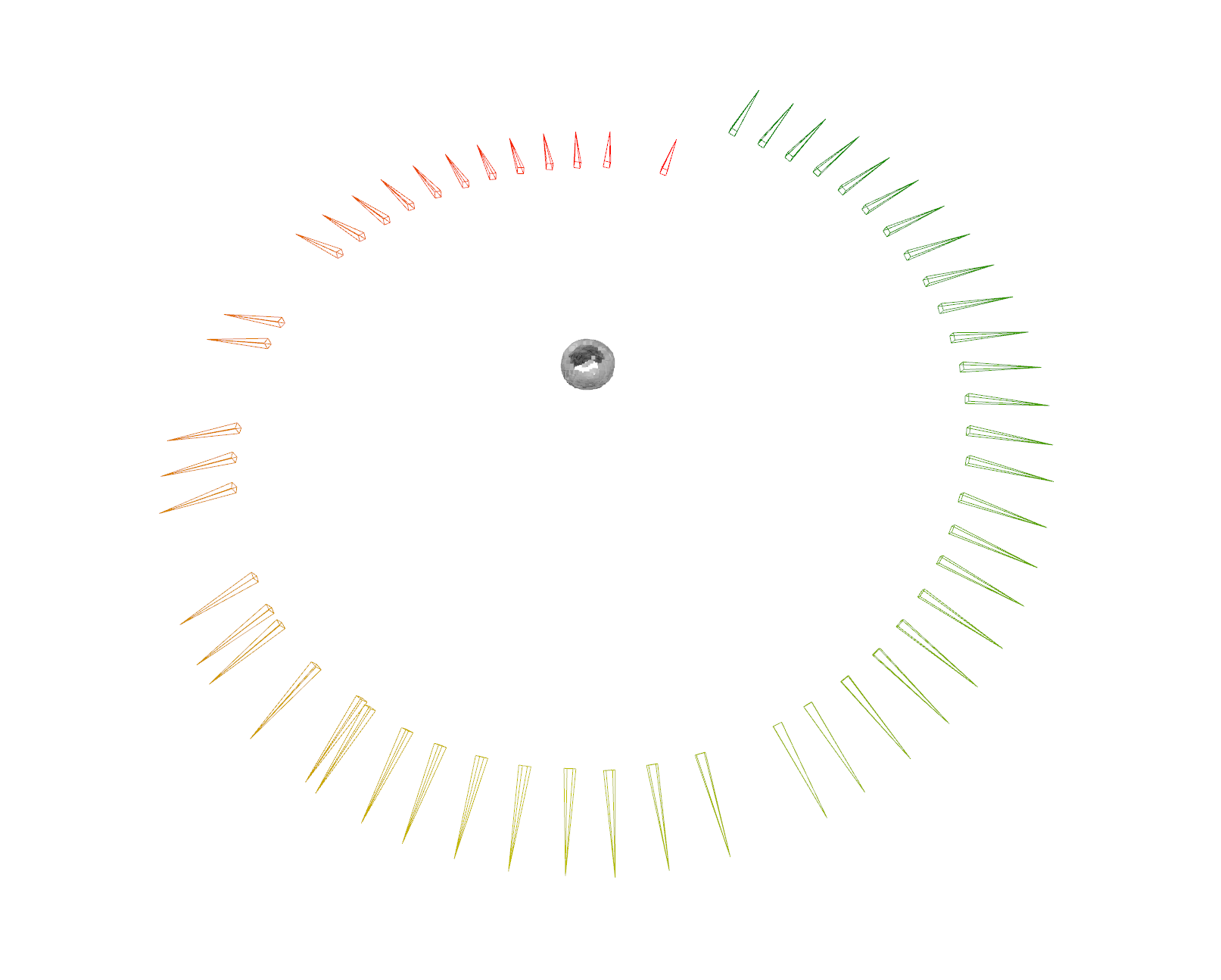} \\
    \includegraphics[width=0.16\linewidth]{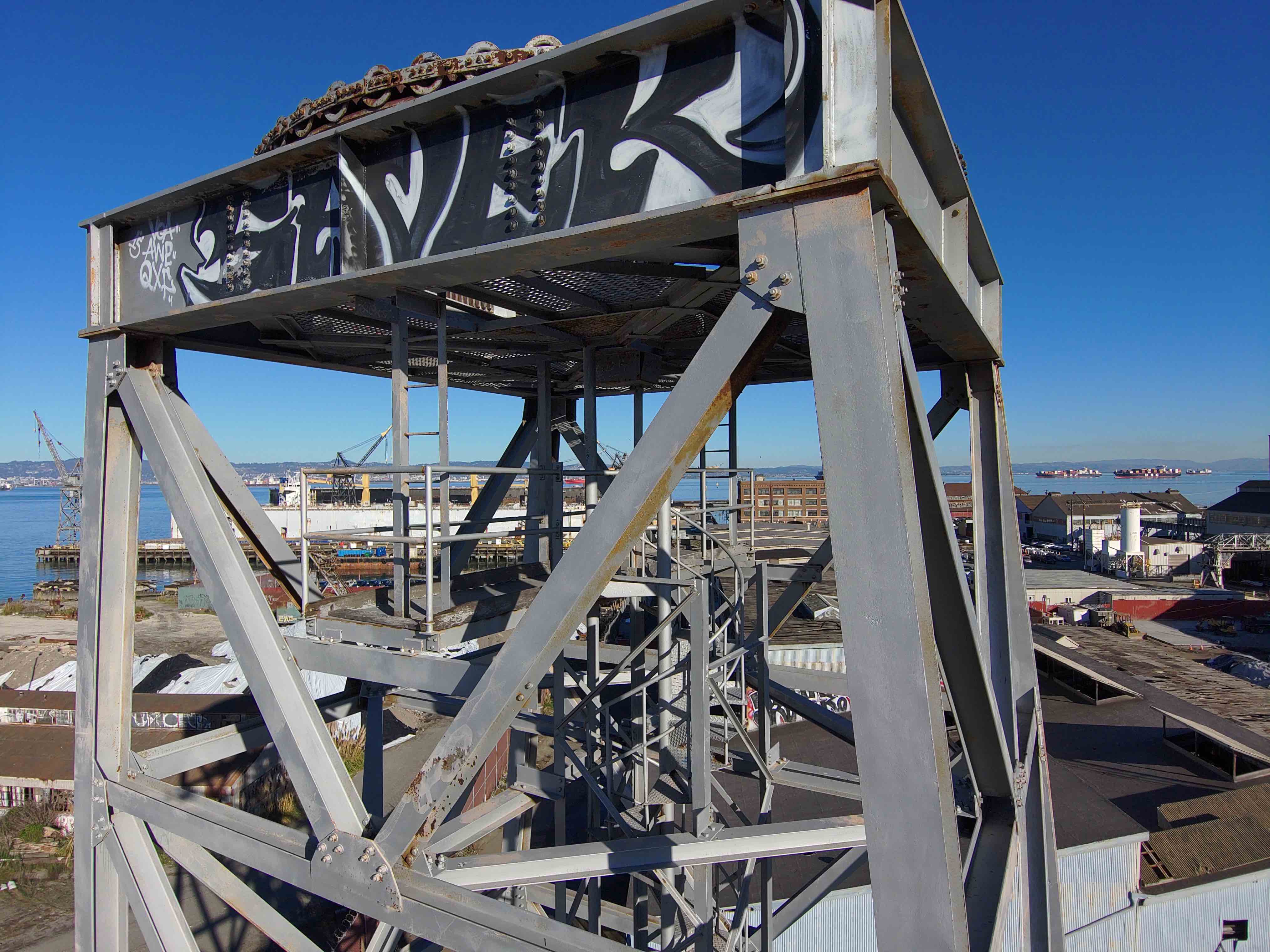}&
    \includegraphics[width=0.16\linewidth]{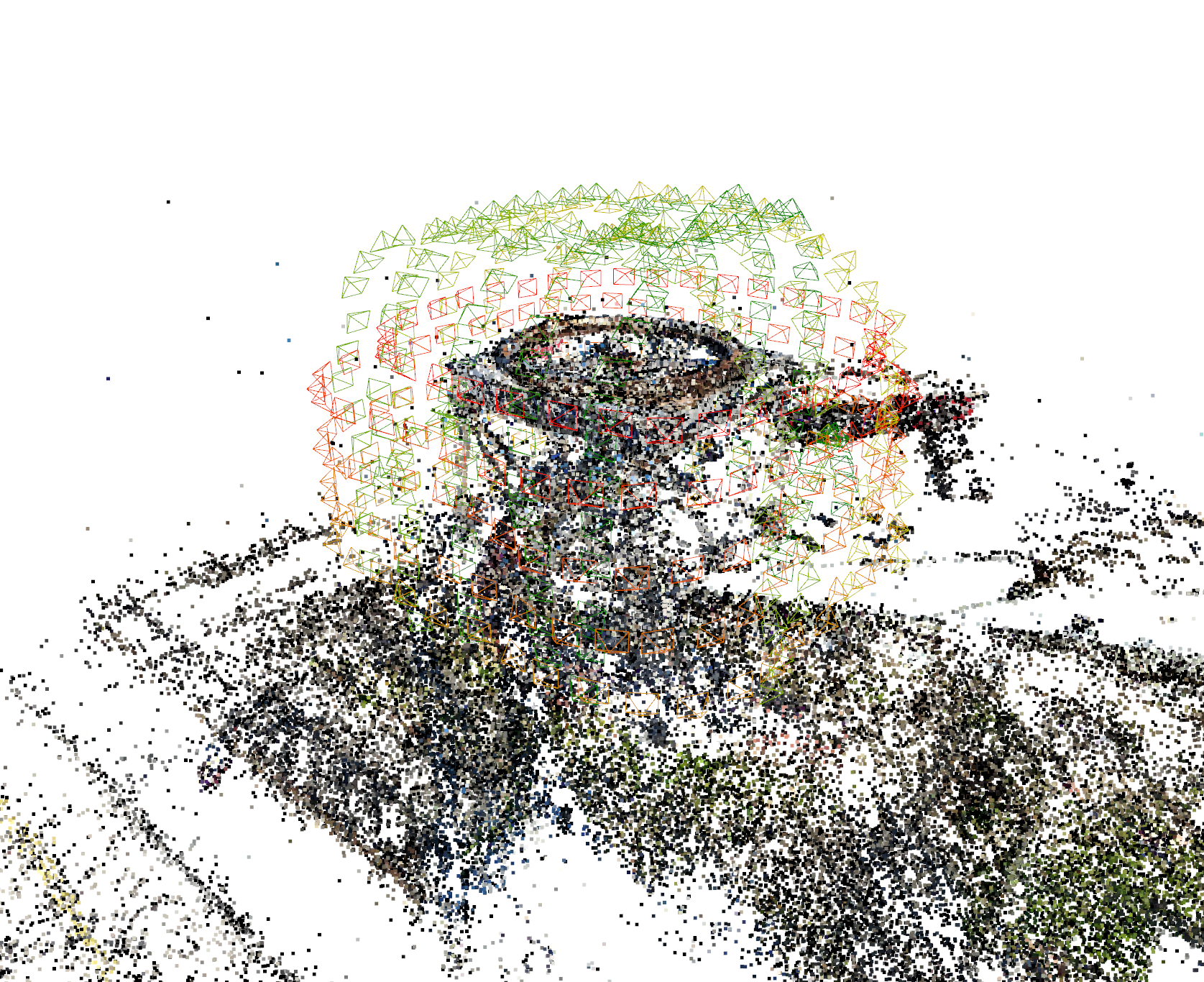}&
    \includegraphics[width=0.16\linewidth]{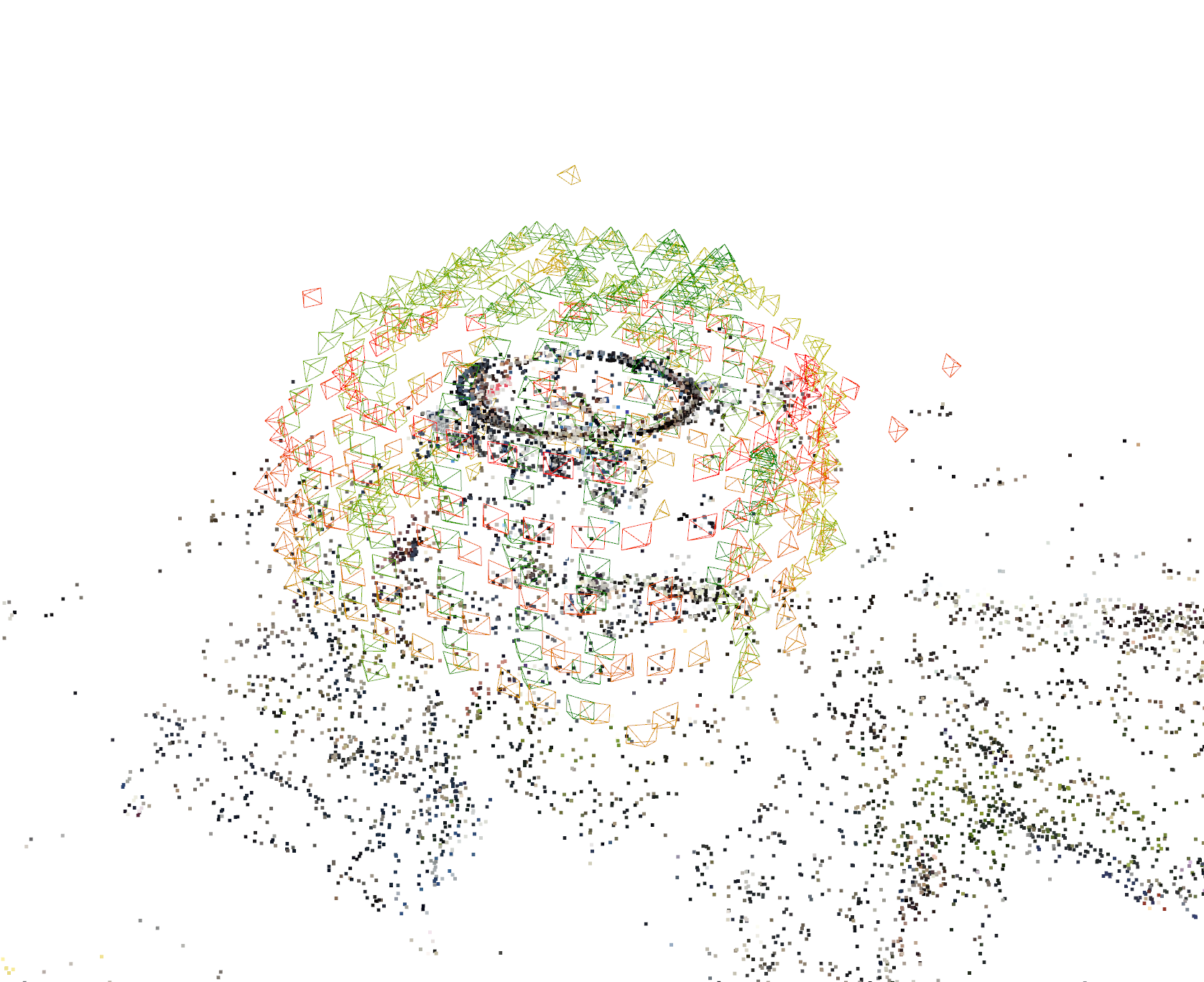}&
    \includegraphics[width=0.16\linewidth]{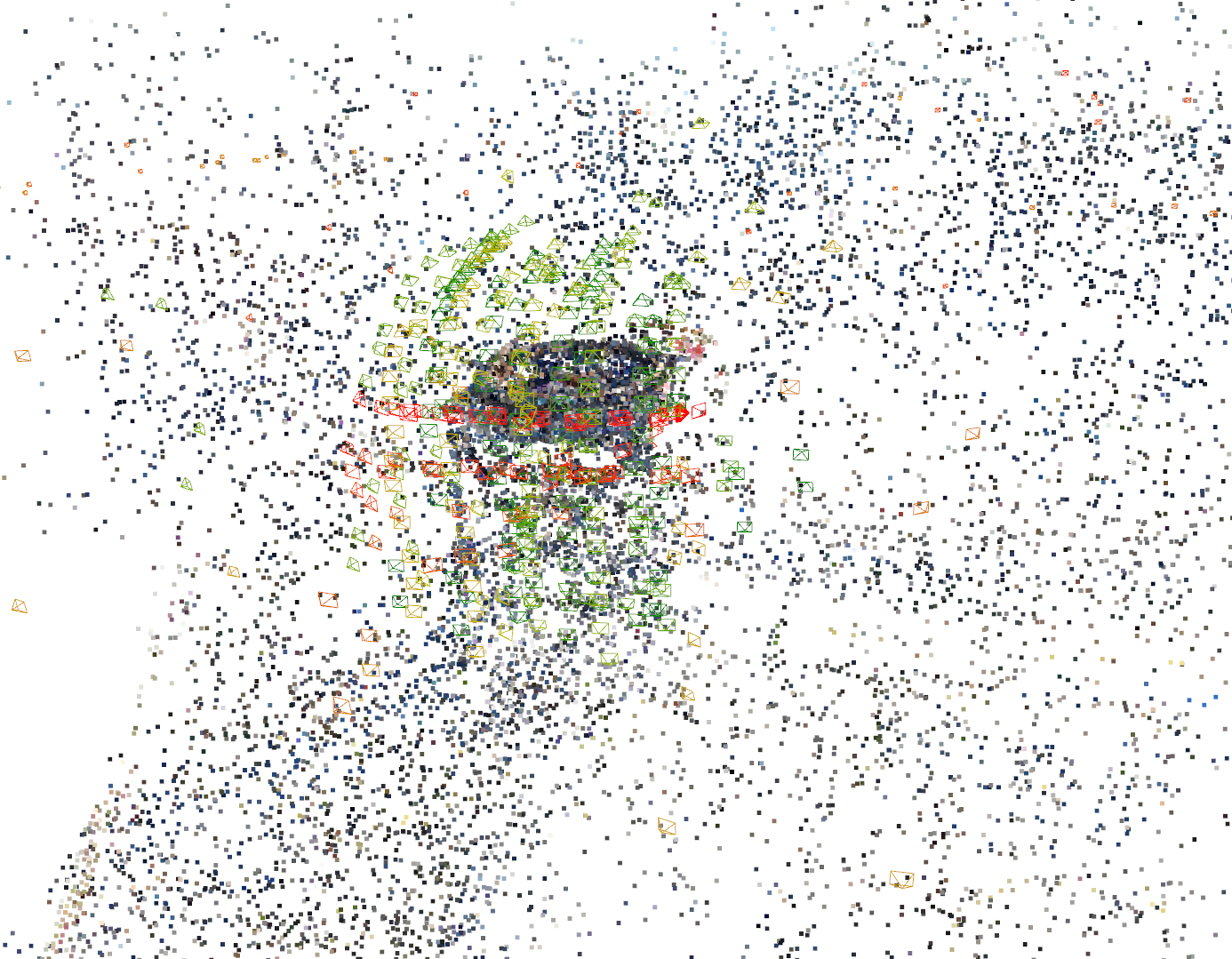}&
    \includegraphics[width=0.16\linewidth]{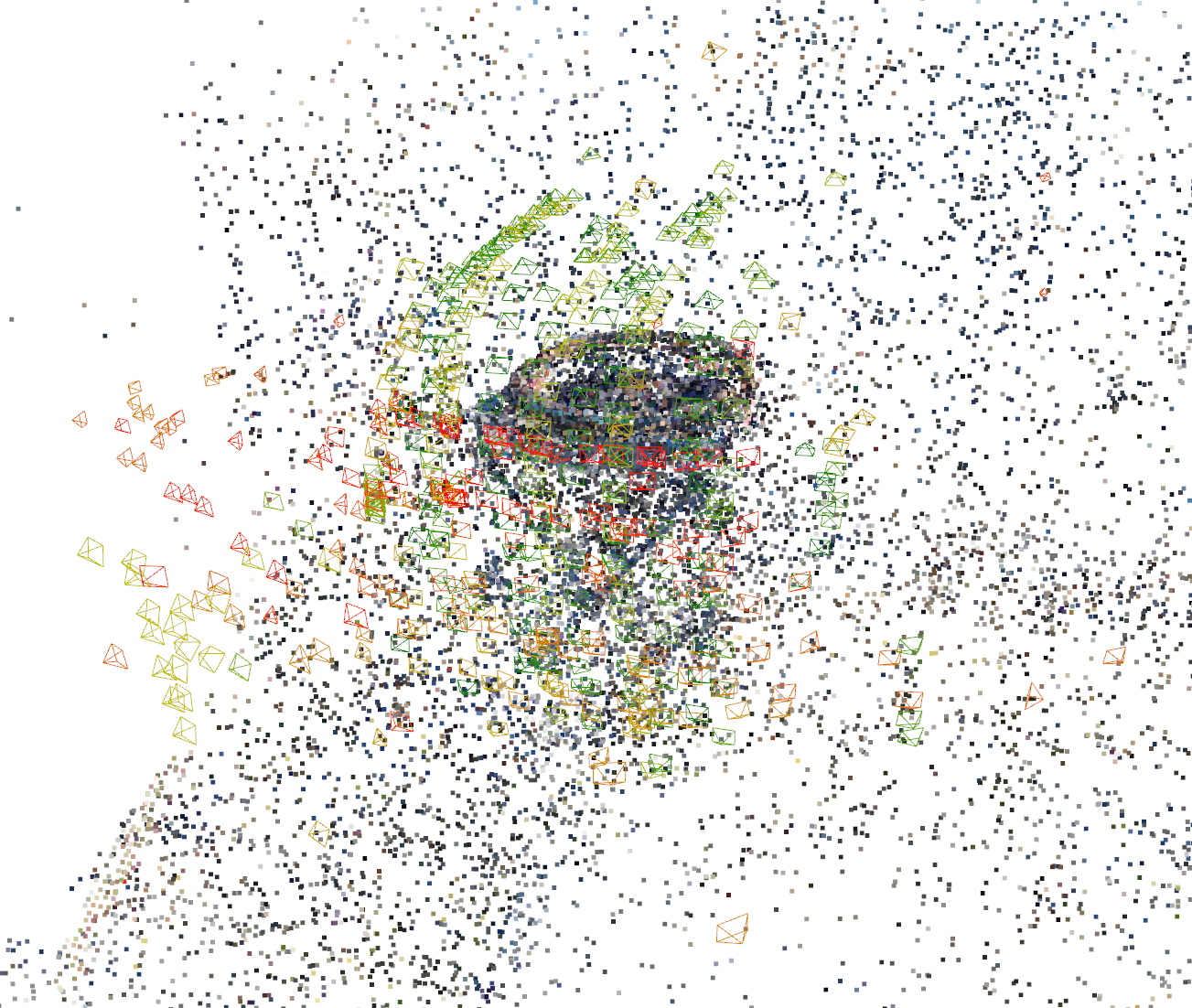}&
    \includegraphics[width=0.16\linewidth]{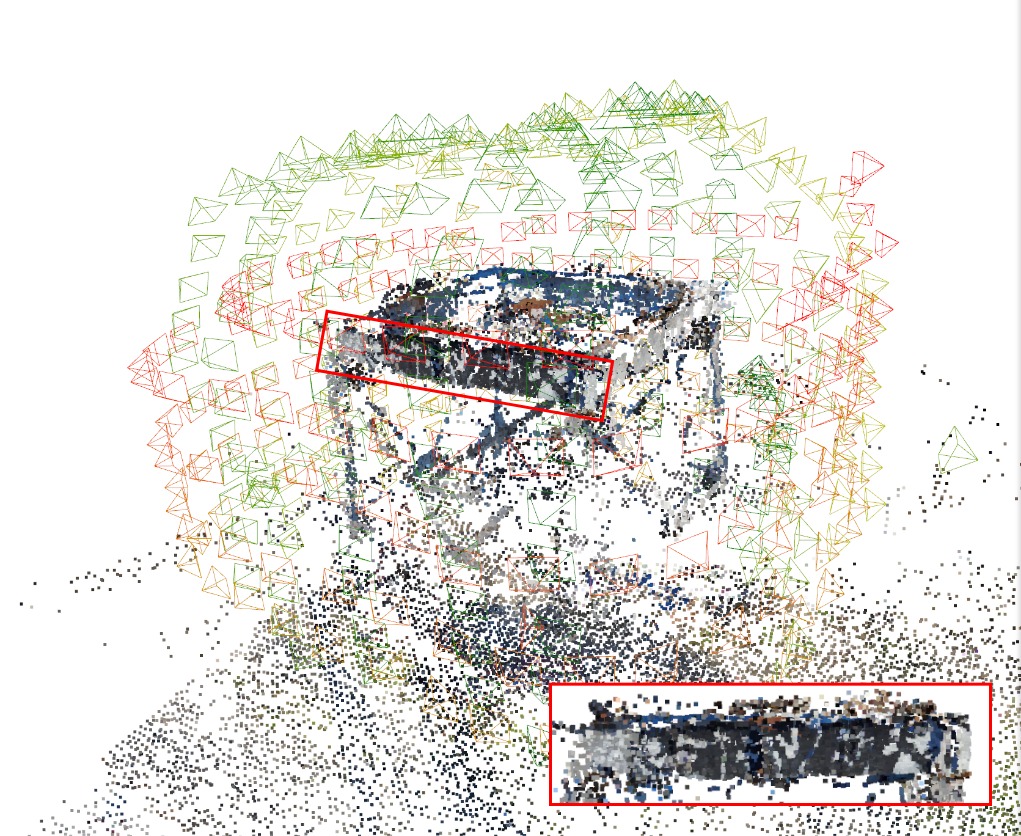}\\
    \includegraphics[width=0.16\linewidth]{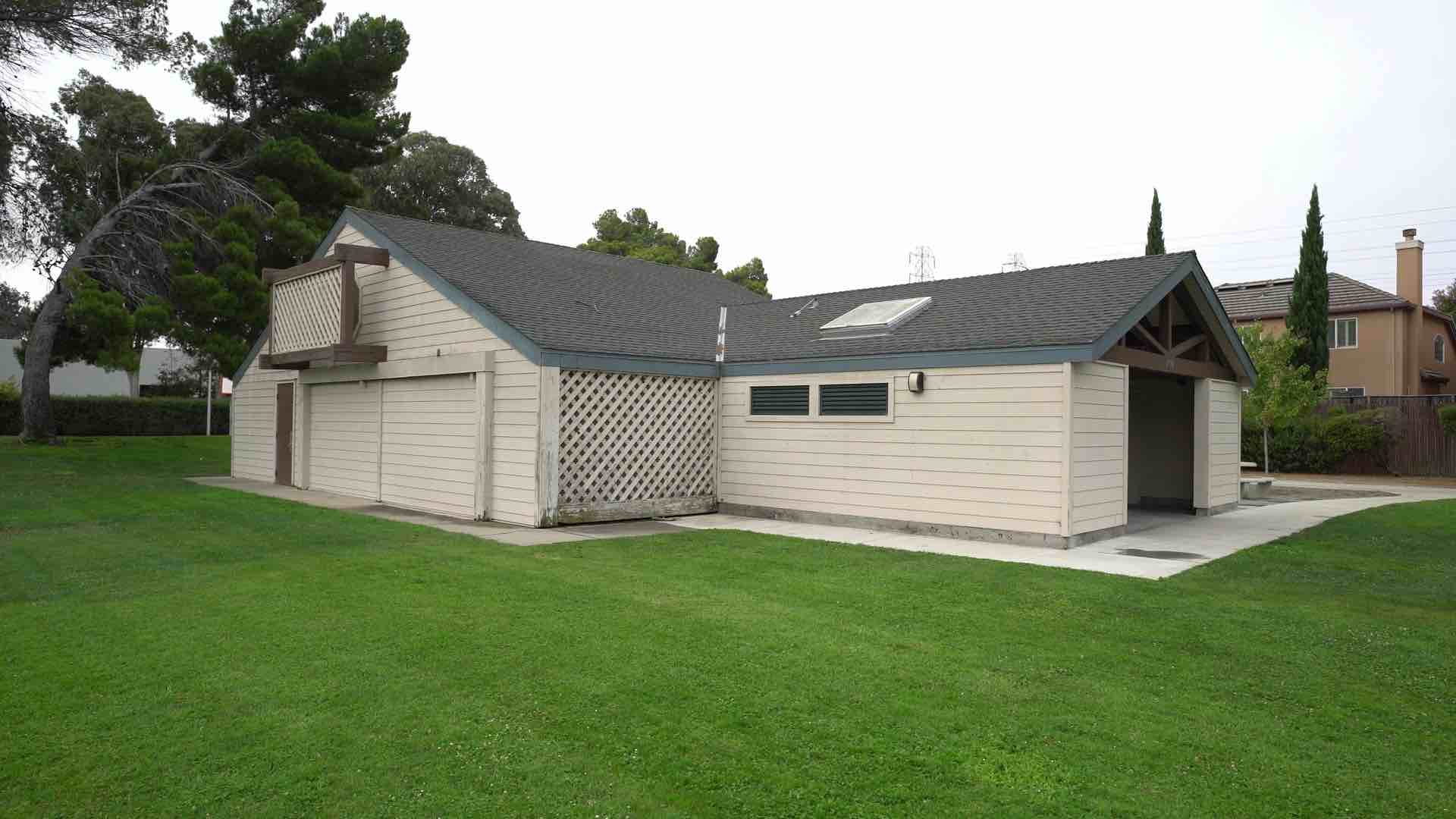}&
    \includegraphics[width=0.16\linewidth]{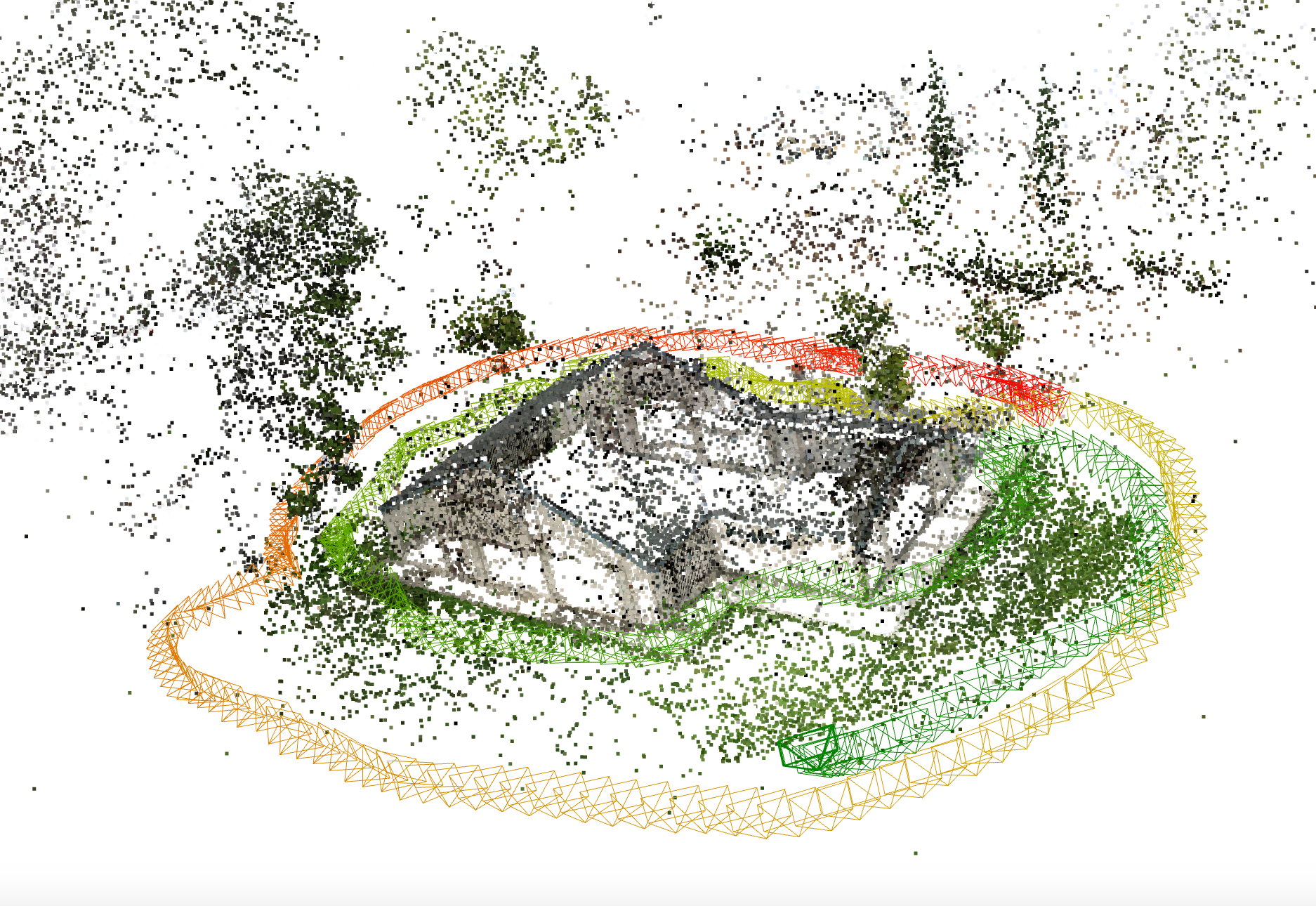}&
    \includegraphics[width=0.16\linewidth]{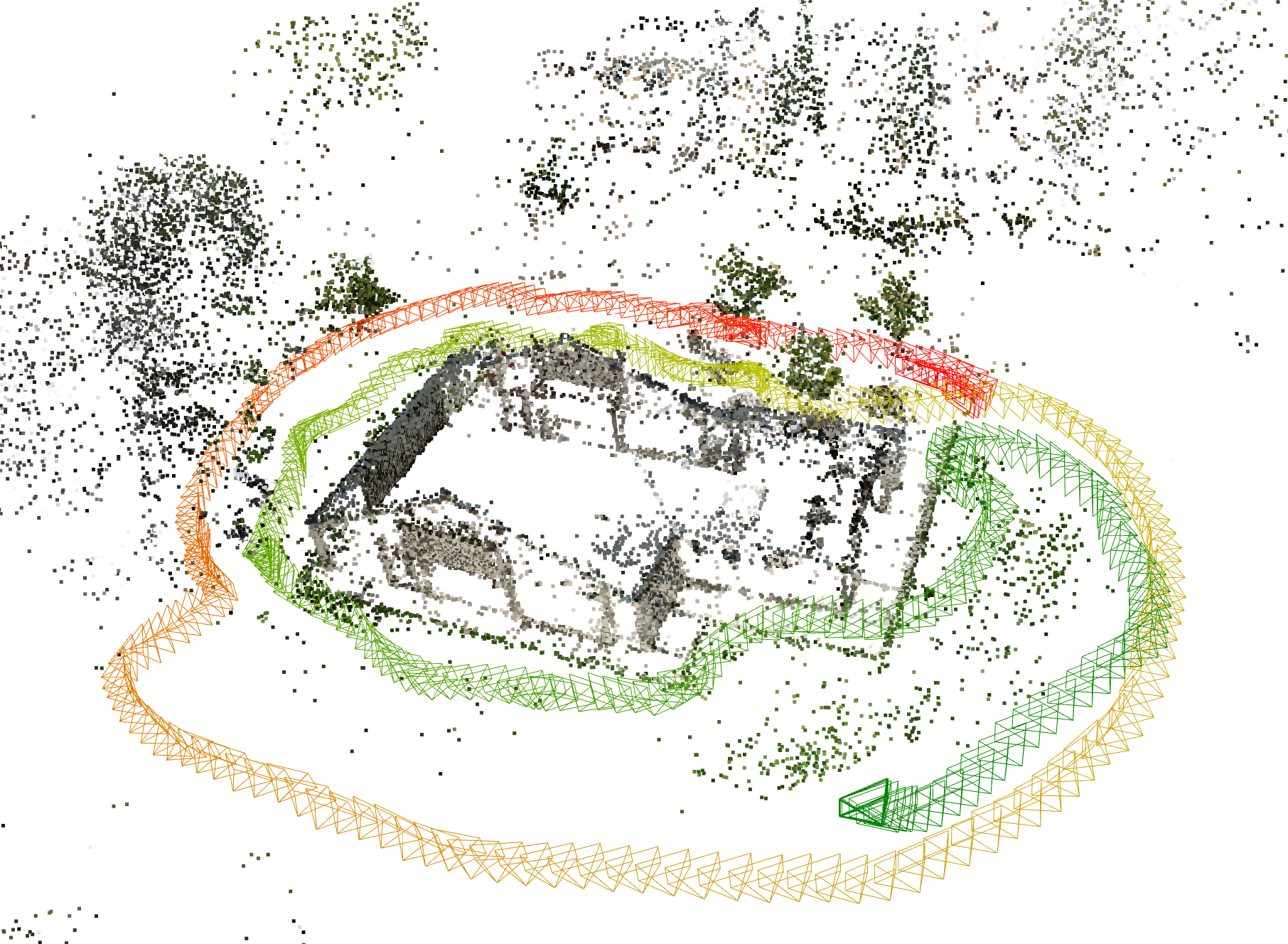}&
    \includegraphics[width=0.16\linewidth]{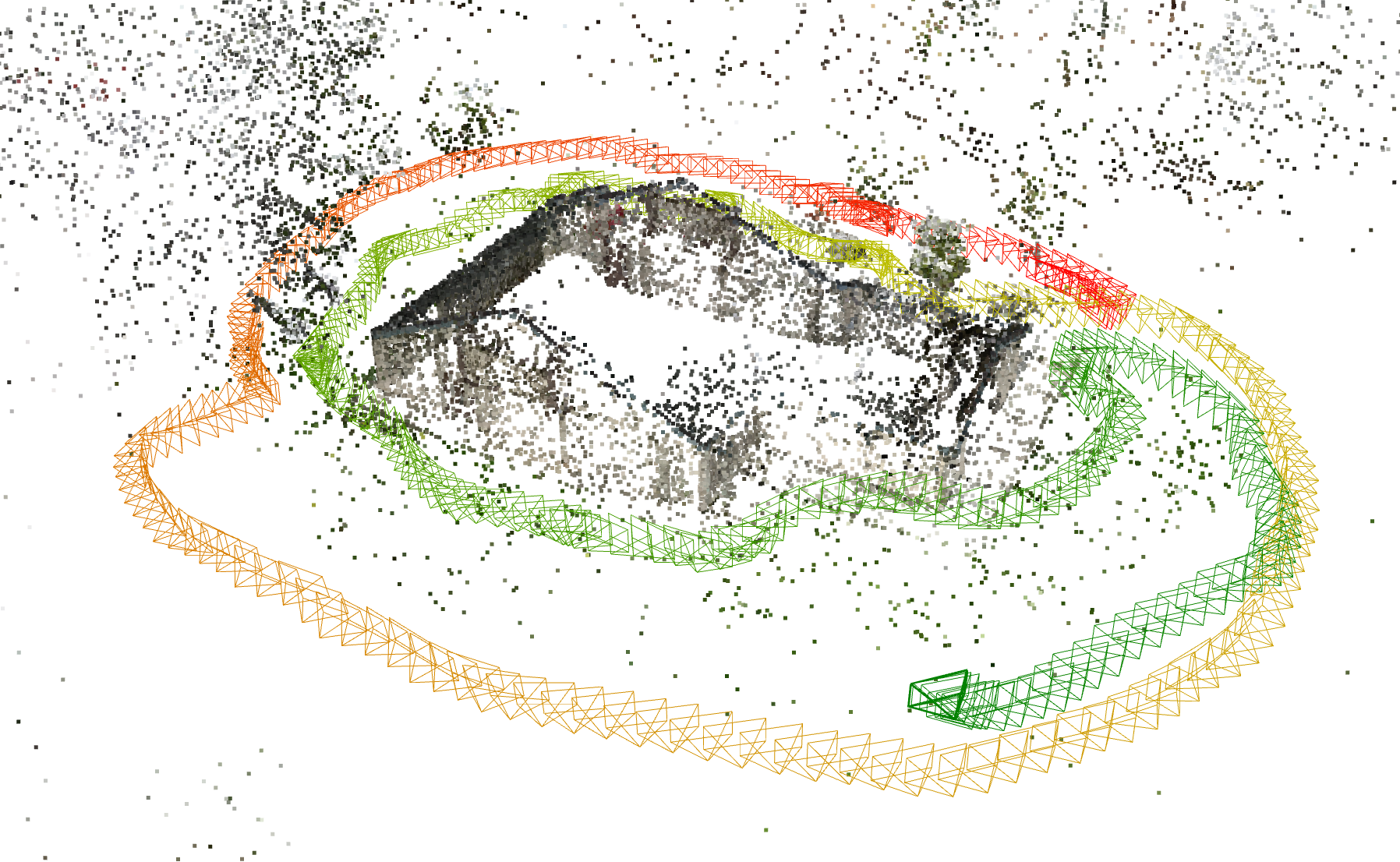}&
    \includegraphics[width=0.16\linewidth]{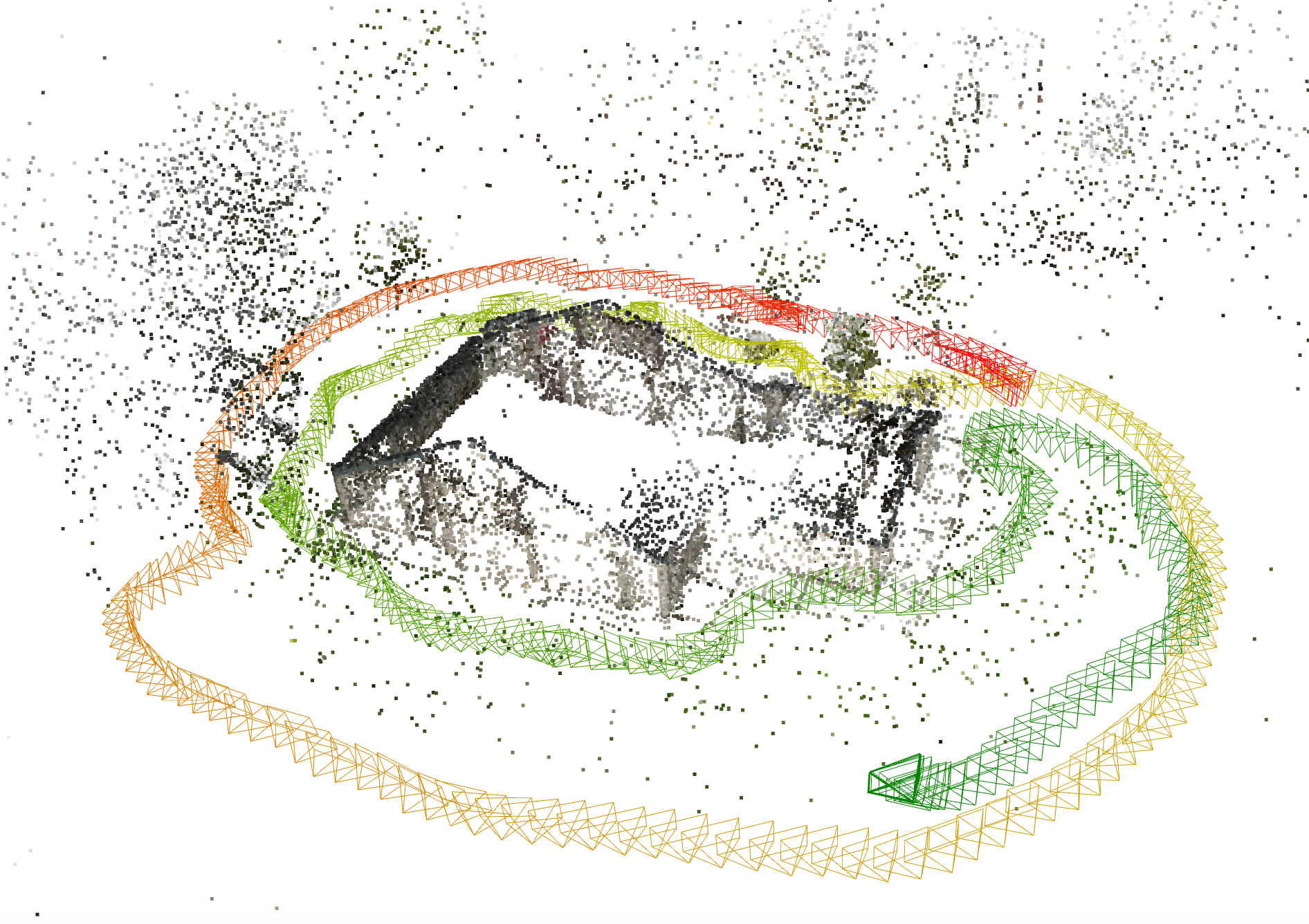}&
    \includegraphics[width=0.16\linewidth]{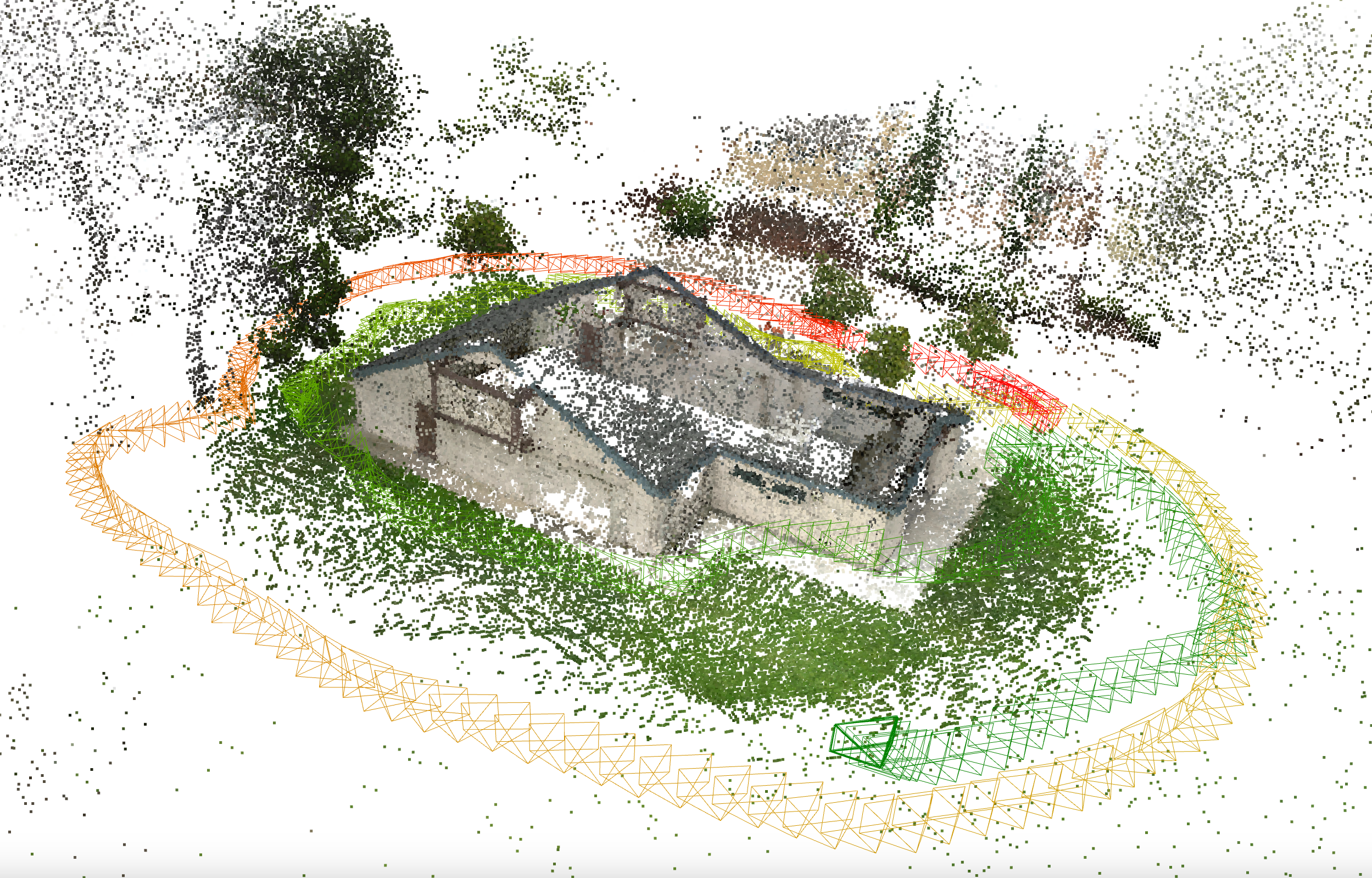}\\
	\end{tabular}
	\caption{Visualization of sparse results on each dataset \textbf{Row 1}: \'Ecole Superieure De Guerre. \textbf{Row 2}: Palace of Fine Arts. \textbf{Row 3}: Skansen Kronan, Gothenburg. \textbf{Row 4}: South Building, UNC. \textbf{Row 5}: Gerrard Hall, UNC. \textbf{Row 6}: King's College, Cambridge. \textbf{Row 7}: RC3 (AstroVision). \textbf{Row 8}: Skydio Crane Mast. \textbf{Row 9}: Tanks and Temples, Barn. Image order is indicated by a red to green colormap.}
	\label{fig:qual-visual}
\end{figure*}

\begin{table}[]
\caption{Average performance of each front-end over 9 datasets, as measured by Pose AUC @N deg. after bundle adjustment (higher is better).}
\vspace{-3mm}
    \centering
    \begin{adjustbox}{width=\linewidth}
\begingroup
\begin{tabular}{lccccc}
\toprule
\rowcolorize Front-End         & @1 deg.                        & @2.5 deg.                      & @5 deg.                        & @10 deg.                       & @20 deg.                       \\

\midrule
LightGlue & 39.2                           & 53.7                           & 63.8                           & 72.1                           & 77.9                           \\
\rowcolorize SuperGlue & 43.3                           & 57.8                           & 67.0                           & 74.2                           & 79.0                     \\

 LoFTR     & 40.0                           & 58.0                           & 70.8                           & 80.3                           & 86.2                           \\
\rowcolorize SIFT      & \textbf{53.1} & \textbf{67.7} & \textbf{76.5} & \textbf{84.3} & \textbf{90.3} \\
\bottomrule
\end{tabular}
\endgroup
\end{adjustbox}
\label{tab:avg-front-end-perf}
\end{table}

\begin{figure}[t!]
    \centering
    \includegraphics[width=0.75\linewidth]{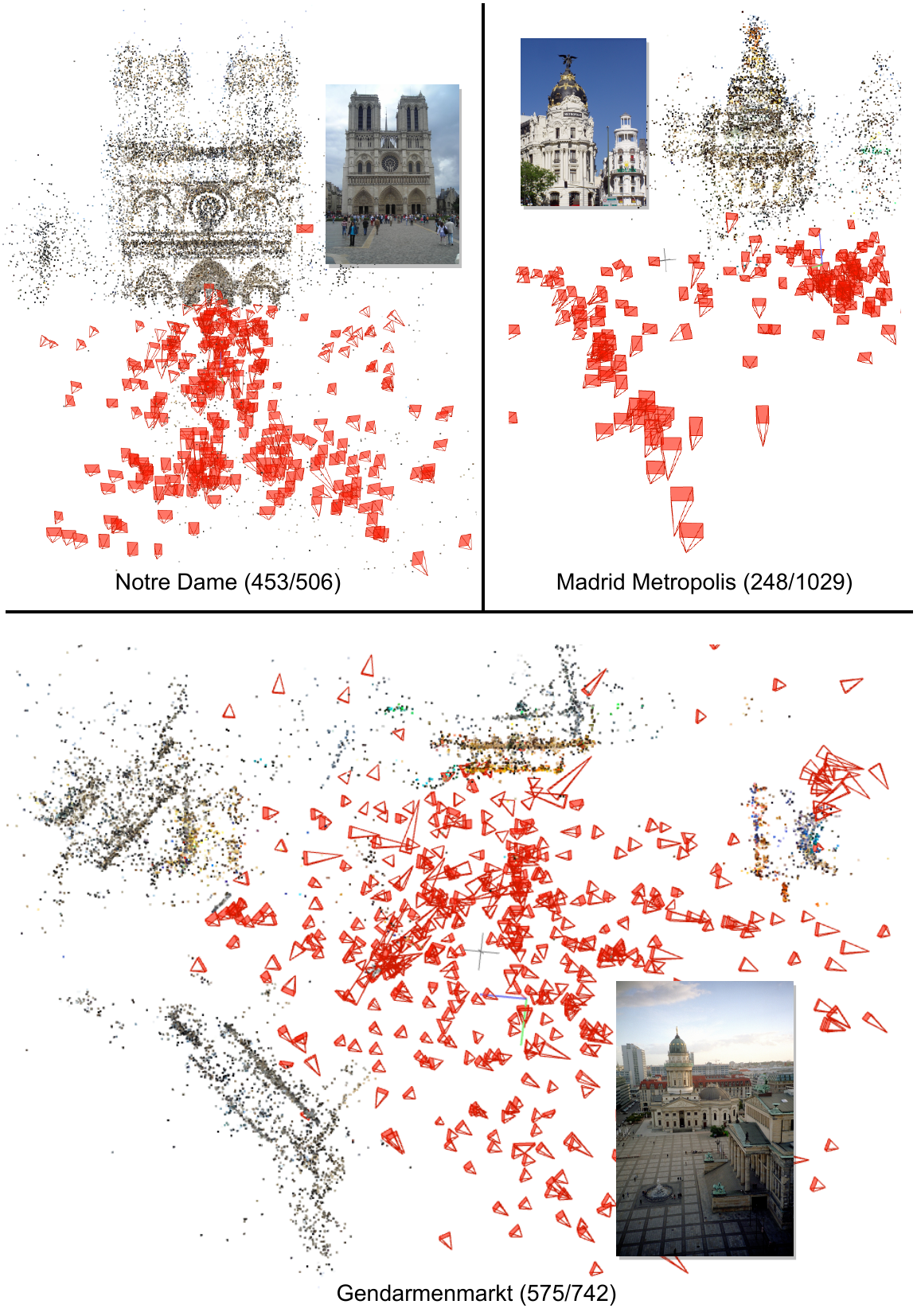}
    \caption{Qualitative results on the 1DSfM~\cite{Wilson14eccv_1DSfM} internet datasets (\# registered images / \# total images).}
    \label{fig:1dsfm-qual-results}
\end{figure}

\noindent \textbf{Putative Correspondence Verification}: We estimate an Essential matrix using GC-RANSAC \cite{Barath18cvpr_GCRANSAC}, with a 4 pixel estimation threshold.

\noindent \textbf{Two-View Bundle Adjustment} We triangulate points from 2-views, pruning based on a reprojection threshold of 0.5 px. We then perform up to 100 bundle adjustment iterations. 

\noindent \textbf{Image Pair Rejection} To reduce the effect of spurious image pair measurements, we discard any image pairs with an inlier ratio (with respect to epipolar geometry estimate) below 10\% or with less than 15 absolute inlier matches, per \cite{Schonberger16cvpr_COLMAP}. 


\noindent \textbf{Rotation Averaging} We start optimization over $\mathbb{SO}(p)$ at $p=3$ and proceed up to $p=30$ or until convergence.

\noindent \textbf{Data Association} We sample up to 100 hypotheses using RANSAC, with a minimum required track length of 3 views.

\noindent \textbf{Global Bundle Adjustment} We perform 3 rounds of global bundle adjustment, with multi-stage point filtering at reprojection thresholds of 10, 5, and 3 pixels, respectively.

We use implementations of Shonan, 1DSfM, and bundle adjustment from the GTSAM \cite{Dellaert12_GTSAM, Dellaert22software_GTSAM} library.


\subsection{Datasets}

We test GTSfM on novel and previously studied datasets:\\
\noindent \textbf{COLMAP Datasets} \cite{colmap_dataset} Two datasets, Gerrard Hall (100 images) and South Building (128 images), captured on the UNC Chapel Hill campus, as used in \cite{Jiang13iccv_GlobalLinearMethodCamPoseReg,Cui15bmvc_LinearGlobTransEstFeatTracks,Zhou20eccv_StochasticBundleAdjustment, Chen20pr_GraphBasedParallelSfM}.

\noindent \textbf{Skydio-Crane-Mast-501} A dataset of extreme difficulty, consisting of 501 images captured in Crane Cove Park, San Francisco\footnote{Available on Sketchfab at \href{https://sketchfab.com/3d-models/crane-mast-3b943b2211284d0cb0bbad32399be58c}{[download link]} with high-resolution images provided by Skydio \href{https://drive.google.com/drive/folders/1SUlpJUwAgEsGaJwrq_WcDzvhKBTa1uY2}{[download link]}.},
featuring object symmetry, repetitive features, thin structures, and extreme depth ranges (from centimeters up to several miles). It is captured by a drone that makes multiple horizontal circular passes around a crane tower, and then makes multiple sweeps over the top of the tower.

\begin{table*}[t]
\vspace{-5mm}
    \centering
    \begin{adjustbox}{width=\linewidth}
\begingroup
\begin{tabular}{lllllllllllllllllllllllll}
\toprule
\rowcolorize Dataset    & Front-End & \# Input & \# Retrieved & Front-End      & Front-End      & Front-End      & VG           & VG           & VG             & VG inlier      & VG inlier      & RA             & TA             & \# Reg.      & BA           & BA           & BA             & BA             & BA             & BA           & BA           & BA           & BA           & BA           \\
\rowcolorize           &           & Images   & Image        & Rot. angular   & Trans.         & Pose           & \# input     & \# inlier    & \# outlier     & Rot            & trans.         & rot.           & trans.         & Cameras      & \# Tracks    & track        & reproj         & rot.           & trans.         & Pose         & Pose         & Pose         & Pose         & Pose         \\
\rowcolorize           &           &          & Pairs        & errors         & Angular        & Errors         & edges        & edges        & edges          & angular        & angular        & angular        & angle          &              & (filtered)   & lengths      & errors         & angle          & angle          & AUC          & AUC          & AUC          & AUC          & AUC          \\
\rowcolorize           &           &          &              & (deg.)         & errors (deg).  & (deg.)         &              &              &                & errors (deg.)  & errors (deg.)  & error (deg.)   & error (deg.)   &              &              & (filtered)   & filtered (px)  & error (deg)    & error (deg.)   & @1 deg.      & @2.5 deg.    & @5 deg.      & @10 deg.     & @20 deg.     \\
\rowcolorize           &           &          &              & ($\downarrow$) & ($\downarrow$) & ($\downarrow$) & ($\uparrow$) & ($\uparrow$) & ($\downarrow$) & ($\downarrow$) & ($\downarrow$) & ($\downarrow$) & ($\downarrow$) & ($\uparrow$) & ($\uparrow$) & ($\uparrow$) & ($\downarrow$) & ($\downarrow$) & ($\downarrow$) & ($\uparrow$) & ($\uparrow$) & ($\uparrow$) & ($\uparrow$) & ($\uparrow$) \\
\midrule
RC3                                    & LightGlue & 65  & 626  & 2.3 / \textbf{21.0} & 1.2 / \textbf{13.7} & 2.3 / \textbf{25.9} & 441                            & 409  & 32  & 1.6 / 2.3                           & 0.9 / 2.4                           & 1.4 / 1.4                           & 0.7 / 1.6                           & 61                           & 1904                            & 4.0 / 4.9                           & 0.5 / 0.5                           & 0.1 / 0.3                           & \textbf{0.1 / 0.2} & 74.2                           & 84.1                           & 89.0                           & 91.4                           & 92.6                             \\
                                       & LoFTR     & 65  & 626  & 8.1 / 48.2                           & 4.3 / 35.4                           & 10.5 / 59.5                          & 204                            & 153  & 51  & 1.5 / 5.2                           & 1.0 / 5.4                           & 1.7 / 2.7                           & 1.1 / 3.6                           & 56                           & \textbf{12537} & 3.0 / 3.5                           & 0.4 / 0.5                           & 0.3 / 0.4                           & 0.2 / 0.4                           & 56.3                           & 72.3                           & 79.2                           & 82.7                           & 84.4                             \\
                                       & SIFT      & 65  & 626  & \textbf{1.4} / 22.1 & \textbf{0.7} / 16.6 & \textbf{1.4} / 27.7 & 288                            & 272  & 16  & \textbf{1.1 / 1.6} & \textbf{0.6} / 2.8 & \textbf{1.0 / 1.2} & 0.5 / 1.4                           & 61                           & 8615                            & 4.0 / 4.4                           & \textbf{0.1 / 0.1} & \textbf{0.1 / 0.2} & \textbf{0.1 / 0.2} & \textbf{80.1} & 87.3                           & 89.8                           & 92.1                           & 93.0                             \\
                                       & SuperGlue & 65  & 626  & 2.0 / 22.5                           & 1.0 / 14.0                           & 2.0 / 27.2                           & \textbf{449}  & 395  & 54  & 1.4 / 2.0                           & 0.7 / \textbf{1.9} & 1.6 / 1.6                           & \textbf{0.5 / 0.7} & \textbf{64} & 2125                            & \textbf{5.0 / 5.2} & 0.5 / 0.5                           & 0.2 / 0.2                           & \textbf{0.1 / 0.2} & 79.9                           & \textbf{91.0} & \textbf{94.8} & \textbf{96.6} & \textbf{97.5}   \\
\rowcolorize \'Ecole      & LightGlue & 35  & 309  & 0.1 / 0.2                            & 0.2 / 0.7                            & 0.2 / 0.7                            & 232                            & 232  & 0   & 0.1 / 0.1                           & 0.2 / 0.3                           & 0.1 / 0.1                           & 0.1 / 0.2                           & 26                           & 2276                            & 5.0 / 7.1                           & 0.5 / 0.6                           & 0.0 / 0.0                           & 0.0 / 0.0                           & 69.8                           & 72.5                           & 73.4                           & 73.8                           & 74.1                             \\
\rowcolorize Superieure & LoFTR     & 35  & 309  & 0.2 / 0.3                            & 0.3 / 1.3                            & 0.4 / 1.4                            & 233                            & 233  & 0   & 0.2 / 0.3                           & 0.3 / 1.3                           & 0.2 / 0.2                           & 0.3 / 0.4                           & 26                           & 24165                           & 4.0 / 4.2                           & 0.4 / 0.6                           & 0.0 / 0.0                           & 0.0 / 0.0                           & 70.3                           & 72.7                           & 73.5                           & 73.9                           & 74.1                             \\
\rowcolorize de Guerre  & SIFT      & 35  & 309  & 0.2 / 0.3                            & 0.5 / 0.6                            & 0.5 / 0.6                            & 295                            & 295  & 0   & 0.2 / 0.3                           & 0.5 / 0.6                           & 0.1 / 0.2                           & 0.4 / 0.5                           & 35                           & 13284                           & 4.0 / 5.5                           & 0.2 / 0.2                           & 0.0 / 0.0                           & 0.1 / 2.8                           & \textbf{84.4} & \textbf{90.2} & \textbf{93.7} & \textbf{95.4} & \textbf{96.3}   \\
\rowcolorize            & SuperGlue & 35  & 309  & 0.1 / 0.2                            & 0.2 / 0.7                            & 0.2 / 0.7                            & 234                            & 234  & 0   & 0.1 / 0.2                           & 0.2 / 0.7                           & 0.1 / 0.1                           & 0.2 / 0.2                           & 26                           & 2227                            & 5.0 / 7.2                           & 0.5 / 0.6                           & 0.0 / 0.1                           & 0.0 / 0.0                           & 69.9                           & 72.5                           & 73.4                           & 73.8                           & 74.1                             \\
Gerrard                                & LightGlue & 98  & 948  & 1.9 / 3.3                            & 1.3 / 3.4                            & 2.1 / 4.3                            & 834                            & 827  & 7   & 1.8 / 2.0                           & 1.2 / 1.7                           & 2.8 / 3.1                           & 1.2 / 1.5                           & 98                           & 10700                           & 4.0 / 4.8                           & 0.8 / 0.9                           & 2.1 / 2.1                           & 0.6 / 0.9                           & \textbf{14.8} & 34.7                           & 58.1                           & 79.0                           & 89.5                             \\
Hall                                   & LoFTR     & 98  & 948  & 2.0 / 7.0                            & 1.5 / 10.4                           & 2.2 / 11.4                           & 648                            & 618  & 30  & 1.7 / 1.9                           & 1.3 / 1.6                           & 3.1 / 3.5                           & 1.0 / 1.5                           & 98                           & 50487                           & 3.0 / 4.0                           & 0.6 / 0.7                           & 1.6 / 2.0                           & 0.6 / 0.8                           & 10.0                           & \textbf{36.4} & 61.5                           & 80.7                           & \textbf{90.4}   \\
                                       & SIFT      & 98  & 948  & 1.6 / 3.5                            & 1.5 / 5.0                            & 2.0 / 5.7                            & 615                            & 608  & 7   & 1.5 / 1.8                           & 1.4 / 2.2                           & 3.9 / 4.0                           & 1.3 / 1.8                           & 98                           & 12281                           & 4.0 / 4.9                           & 0.4 / 0.6                           & 2.3 / 2.5                           & 0.8 / 1.2                           & 4.5                            & 26.0                           & 50.4                           & 74.9                           & 87.4                             \\
                                       & SuperGlue & 98  & 948  & 1.8 / 2.5                            & 1.3 / 2.4                            & 2.1 / 3.2                            & 834                            & 828  & 6   & 1.7 / 1.9                           & 1.2 / 1.6                           & 2.5 / 2.9                           & 1.1 / 1.4                           & 98                           & 10741                           & 4.0 / 4.9                           & 0.8 / 0.9                           & 1.6 / 1.9                           & 0.8 / 0.9                           & 6.0                            & 33.2                           & \textbf{61.7} & \textbf{80.8} & \textbf{90.4}   \\
\rowcolorize Kings      & LightGlue & 328 & 4339 & 0.6 / 5.7                            & 4.7 / 31.1                           & 4.7 / 32.8                           & 2867                           & 2799 & 68  & 0.5 / 1.2                           & 3.6 / 26.8                          & 1.0 / 1.2                           & 4.2 / 24.6                          & 327                          & 9412                            & 4.0 / 7.3                           & 0.6 / 0.7                           & 0.1 / 0.2                           & 0.2 / 0.8                           & 60.3                           & 79.1                           & 88.2                           & 93.5                           & 96.5                             \\
\rowcolorize College    & LoFTR     & 328 & 4339 & 0.6 / 5.6                            & 4.6 / 29.7                           & 4.6 / 31.8                           & 2204                           & 2171 & 33  & 0.6 / 1.2                           & 3.8 / 25.2                          & 1.3 / 1.4                           & 3.7 / 20.8                          & 327                          & 130935                          & 3.0 / 4.1                           & 0.6 / 0.7                           & 0.1 / 0.1                           & 0.1 / 0.8                           & 68.8                           & 83.9                           & 90.9                           & 94.7                           & 96.9                             \\
\rowcolorize Cambridge  & SIFT      & 328 & 4339 & 0.6 / 2.2                            & 5.1 / 27.7                           & 5.1 / 28.1                           & 2845                           & 2792 & 53  & 0.6 / 1.1                           & 4.7 / 25.7                          & 1.0 / 1.1                           & 4.9 / 25.4                          & 328                          & 39440                           & 4.0 / 6.7                           & 0.4 / 0.5                           & 0.1 / 0.1                           & 0.1 / 0.7                           & \textbf{73.1} & \textbf{87.4} & \textbf{92.9} & \textbf{96.0} & \textbf{97.7}   \\
\rowcolorize            & SuperGlue & 328 & 4339 & 0.6 / 5.9                            & 4.6 / 32.2                           & 4.7 / 34.1                           & 2837                           & 2756 & 81  & 0.5 / 1.2                           & 3.5 / 27.4                          & 1.0 / 1.1                           & 4.8 / 27.0                          & 327                          & 9632                            & 4.0 / 7.8                           & 0.7 / 0.8                           & 0.2 / 0.3                           & 0.2 / 1.0                           & 57.6                           & 77.3                           & 86.5                           & 92.3                           & 95.6                             \\
Palace                                 & LightGlue & 281 & 2811 & 0.1 / 1.0                            & 0.8 / 3.9                            & 0.8 / 4.3                            & 2333                           & 2332 & 1   & 0.1 / 0.5                           & 0.7 / 3.3                           & 2.3 / 3.7                           & 0.9 / 3.0                           & 281                          & 9724                            & 5.0 / 8.2                           & 0.6 / 0.7                           & 0.7 / 1.2                           & 0.2 / 1.0                           & 20.9                           & 53.5                           & 72.5                           & 85.8                           & 92.7                            \\
of Fine                                & LoFTR     & 281 & 2811 & 0.4 / 3.9                            & 2.3 / 10.9                           & 2.3 / 12.6                           & 2216                           & 2211 & 5   & 0.4 / 1.1                           & 2.2 / 9.4                           & 2.5 / 4.2                           & 2.0 / 8.4                           & 279                          & 74138                           & 4.0 / 4.4                           & 0.6 / 0.7                           & 2.0 / 3.3                           & 0.7 / 3.1                           & 2.3                            & 20.0                           & 50.2                           & 66.9                           & 78.8                             \\
Arts                                   & SIFT      & 281 & 2811 & 0.1 / 0.4                            & 0.8 / 2.2                            & 0.8 / 2.3                            & 2500                           & 2498 & 2   & 0.1 / 0.3                           & 0.8 / 2.2                           & 0.6 / 0.6                           & 0.7 / 2.1                           & 281                          & 65921                           & 5.0 / 7.9                           & 0.2 / 0.3                           & 0.1 / 0.2                           & 0.0 / 0.4                           & \textbf{80.7} & \textbf{91.8} & \textbf{95.7} & \textbf{97.7} & \textbf{98.7}   \\
                                       & SuperGlue & 281 & 2811 & 0.1 / 1.2                            & 0.7 / 3.9                            & 0.8 / 4.3                            & 2364                           & 2363 & 1   & 0.1 / 0.4                           & 0.7 / 3.3                           & 2.2 / 3.6                           & 1.0 / 3.2                           & 281                          & 10032                           & 5.0 / 7.4                           & 0.6 / 0.7                           & 0.2 / 0.4                           & 0.1 / 0.7                           & 61.7                           & 82.0                           & 90.2                           & 94.7                           & 97.1                             \\
\rowcolorize Skansen    & LightGlue & 131 & 1359 & 0.3 / 3.6                            & 1.2 / 5.1                            & 1.3 / 6.6                            & 1146                           & 1143 & 3   & 0.3 / 0.8                           & 1.2 / 3.2                           & 5.9 / 5.2                           & 2.2 / 4.9                           & 127                          & 6779                            & 5.0 / 6.4                           & 0.6 / 0.7                           & 0.1 / 0.2                           & 0.1 / 1.3                           & 73.5                           & 85.2                           & 90.0                           & 92.7                           & 94.1                             \\
\rowcolorize Kronan     & LoFTR     & 131 & 1359 & 0.4 / 3.1                            & 1.6 / 6.7                            & 1.6 / 7.7                            & 1106                           & 1082 & 24  & 0.4 / 0.7                           & 1.5 / 3.9                           & 0.6 / 0.7                           & 1.5 / 3.2                           & 131                          & 65860                           & 4.0 / 4.6                           & 0.5 / 0.7                           & 0.1 / 0.1                           & 0.1 / 1.3                           & 77.0                           & 87.8                           & 92.6                           & 95.6                           & 97.0                             \\
\rowcolorize Gothenburg & SIFT      & 131 & 1359 & 0.3 / 1.9                            & 1.4 / 5.0                            & 1.4 / 6.0                            & 1166                           & 1165 & 1   & 0.3 / 0.5                           & 1.3 / 4.4                           & 0.4 / 0.5                           & 1.3 / 2.9                           & 131                          & 25137                           & 6.0 / 9.8                           & 0.3 / 0.4                           & 0.0 / 0.0                           & 0.0 / 0.8                           & \textbf{87.3} & \textbf{93.7} & \textbf{96.3} & \textbf{97.8} & \textbf{98.5}   \\
\rowcolorize            & SuperGlue & 131 & 1359 & 0.3 / 1.5                            & 1.2 / 3.9                            & 1.2 / 4.4                            & 1161                           & 1160 & 1   & 0.3 / 0.5                           & 1.2 / 3.2                           & 0.3 / 0.4                           & 1.2 / 2.4                           & 131                          & 7479                            & 5.0 / 8.7                           & 0.7 / 0.8                           & 0.1 / 0.1                           & 0.1 / 1.2                           & 82.4                           & 90.6                           & 94.4                           & 96.4                           & 97.4                             \\
South                                  & LightGlue & 128 & 1448 & 0.6 / 12.1                           & \textbf{0.5} / 14.2 & 0.8 / 16.2                           & 1268                           & 1177 & 91  & 0.5 / 6.0                           & 0.5 / 6.3                           & 27.1 / 26.2                         & 2.1 / 13.8                          & 127                          & 5687                            & 3.0 / 4.0                           & 0.6 / 0.7                           & 5.4 / 7.1                           & 2.8 / 4.6                           & 0.0                            & 0.6                            & 15.9                           & 39.0                           & 64.6                             \\
Building                               & LoFTR     & 128 & 1448 & 0.5 / 4.4                            & \textbf{0.5} / 5.8  & 0.7 / 6.8                            & 939                            & 927  & 12  & 0.4 / 0.7                           & 0.5 / 0.9                           & 0.7 / 0.8                           & 0.4 / 0.6                           & 128                          & 74455                           & 4.0 / 4.3                           & 0.5 / 0.6                           & 0.4 / 0.5                           & 0.2 / 0.3                           & \textbf{48.8} & \textbf{79.5} & \textbf{89.8} & \textbf{94.9} & \textbf{97.4}  \\
                                       & SIFT      & 128 & 1448 & \textbf{0.5 / 1.5}  & 0.7 / \textbf{1.7}  & 0.9 / 2.0                            & 799                            & 794  & 5   & 0.5 / 0.8                           & 0.7 / 1.1                           & 1.8 / 1.8                           & 0.7 / 1.0                           & 128                          & 20671                           & 4.0 / 5.1                           & 0.2 / 0.3                           & 0.9 / 1.1                           & 0.2 / 0.6                           & 28.4                           & 58.3                           & 78.7                           & 89.3                           & 94.7                             \\
                                       & SuperGlue & 128 & 1448 & 0.5 / 11.3                           & \textbf{0.5} / 13.2 & 0.7 / 15.2                           & 1270                           & 1166 & 104 & 0.4 / 4.2                           & 0.4 / 4.5                           & 19.5 / 18.3                         & 1.5 / 8.6                           & 128                          & 6224                            & 3.0 / 4.1                           & 0.6 / 0.7                           & 6.4 / 8.2                           & 4.4 / 4.4                           & 1.1                            & 8.0                            & 20.7                           & 42.6                           & 63.5                            \\
\rowcolorize Tanks      & LightGlue & 410 & 4309 & 0.5 / 1.6                            & \textbf{0.6 / 1.5}  & 0.8 / 2.1                            & 3727                           & 3727 & 0   & 0.5 / 1.0                           & 0.6 / 1.1                           & 1.2 / 1.5                           & 0.6 / 0.9                           & 410                          & 25590                           & 4.0 / 5.4                           & 0.7 / 0.8                           & 0.6 / 0.6                           & 0.2 / 0.3                           & 38.9                           & 73.8                           & 86.9                           & 93.4                           & 96.7                             \\
\rowcolorize and        & LoFTR     & 410 & 4309 & 0.5 / 1.6                            & 0.8 / 3.0                            & 0.9 / 3.3                            & 3467                           & 3447 & 20  & 0.5 / 0.9                           & 0.8 / 1.2                           & 1.3 / 1.5                           & 0.6 / 1.0                           & 410                          & 225682                          & 3.0 / 3.9                           & 0.6 / 0.7                           & 0.8 / 0.8                           & 0.4 / 0.4                           & 26.7                           & 67.3                           & 83.7                           & 91.8                           & 95.9                            \\
\rowcolorize Temples    & SIFT      & 410 & 4309 & \textbf{0.4 / 1.2}  & 0.7 / 2.2                            & 0.8 / 2.4                            & 3340                           & 3329 & 11  & 0.4 / 0.9                           & 0.7 / 1.5                           & 1.2 / 1.3                           & 0.6 / 1.0                           & 410                          & 63735                           & 4.0 / 5.6                           & 0.4 / 0.5                           & 0.5 / 0.6                           & 0.3 / 0.3                           & \textbf{39.5} & \textbf{74.3} & \textbf{87.1} & \textbf{93.6} & \textbf{96.8}   \\
\rowcolorize Barn       & SuperGlue & 410 & 4309 & 0.5 / 1.3                            & 0.7 / 1.5                            & 0.8 / 1.8                            & 3744                           & 3742 & 2   & 0.5 / 1.2                           & 0.7 / 1.2                           & 1.7 / 2.9                           & 0.7 / 1.0                           & 410                          & 25349                           & 4.0 / 5.2                           & 0.7 / 0.8                           & 0.7 / 0.8                           & 0.4 / 0.6                           & 30.9                           & 65.3                           & 81.7                           & 90.8                           & 95.4                             \\
Skydio                                 & LightGlue & 500 & 6263 & 1.6 / 16.7                           & 2.9 / 32.4                           & 3.6 / 36.4                           & \textbf{2345} & 2127 & 218 & 1.1 / 6.7                           & 1.6 / 15.0                          & 46.7 / 60.8                         & 3.2 / 16.2                          & 480                          & 17228                           & 3.0 / 3.8                           & 0.8 / 0.9                           & 50.9 / 64.6                         & 50.9 / 64.0                         & 0.0                            & 0.0                            & 0.1                            & 0.2                            & 0.7                              \\
Crane                                  & LoFTR     & 500 & 6263 & \textbf{1.3} / 4.4  & \textbf{2.5 / 27.4} & \textbf{2.8 / 27.9} & 1977                           & 1861 & 116 & \textbf{1.0 / 2.3} & 1.8 / 16.7                          & 2.3 / 3.7                           & 1.9 / 8.9                           & 467                          & 36462                           & 3.0 / 3.5                           & 0.6 / 0.7                           & 0.8 / 1.5                           & 4.5 / 8.1                           & \textbf{0.1}  & \textbf{1.8}  & \textbf{16.0} & \textbf{41.4} & \textbf{60.7}   \\
Mast                                   & SIFT      & 500 & 6263 & 1.5 / \textbf{3.8}  & 4.3 / 31.6                           & 4.7 / 31.9                           & 1991                           & 1874 & 117 & 1.3 / 2.4                           & 3.0 / 25.3                          & 7.7 / 7.5                           & 3.3 / 15.0                          & 472                          & 12579                           & 3.0 / 4.0                           & 0.6 / 0.8                           & 3.5 / 3.6                           & 8.1 / 10.3                          & \textbf{0.1}  & 0.2                            & 3.4                            & 22.0                           & 50.0                             \\
                                       & SuperGlue & 500 & 6263 & 1.5 / 16.5                           & 2.9 / 34.9                           & 3.3 / 38.9                           & 2186                           & 2019 & 167 & 1.0 / 7.8                           & 1.5 / 16.9                          & 51.2 / 73.7                         & 3.4 / 18.5                          & 488                          & 18342                           & 3.0 / 3.9                           & 0.8 / 0.9                           & 50.7 / 77.4                         & 55.8 / 59.6                         & 0.0                            & 0.0                            & 0.0                            & 0.0                            & 0.1                            \\
   
\bottomrule
\end{tabular}
\endgroup
\end{adjustbox}
    \caption{Results with 5 retrieval matches, 10 frame lookahead, at 760p resolution. `Median / Mean' statistics are given for 1d distributions.  Abbreviations: View Graph Estimation (VG), Rotation Averaging (RA), Translation Averaging (TA), Bundle Adjustment (BA)}
    \label{tab:acc-table-all-datasets}
\end{table*}


\noindent \textbf{Tanks and Temples \textit{Barn}} A dataset of 410 images from \cite{Knapitsch17acmtg_TanksTemples}, which provides both ground-truth LiDAR scans, and ground-truth cameras poses (via COLMAP). 


\noindent \textbf{RC3, Astrovision} \cite{driver2023astrovision} A dataset of 65 images of the dwarf planet Ceres captured by NASA's Dawn spacecraft during the Rotation Characterization 3 (RC3) phase at a distance of 8,500 miles (13,500 kilometers).

\noindent \textbf{Gendarmenmarkt} A phototourism dataset \cite{Snavely06siggraph_PhotoTourism} captured in a market square in Berlin and used in 1dSfM \cite{Wilson14eccv_1DSfM}.

\noindent \textbf{Olsson Datasets} Datasets captured by Carl Olsson \cite{Enqvist11tr_StableSfMRotConsistency}, including \textit{Palace of Fine Arts, S.F.}, \textit{Ecole Superieure de Guerre}, \textit{King's College, Cambridge}, and \textit{Skansen-Kronan, Gothenburg, Sweden}, with ground-truth provided.




\subsection{Evaluation Metrics}

We quantitatively measure the effect of different front-ends on GTSfM's performance, on each dataset, using as ground-truth either the output of COLMAP \cite{Schonberger16cvpr_COLMAP}, a high-quality incremental SfM system, or the ground truth provided by Olsson \emph{et al.} \cite{Enqvist11iccvw_NonsequentialSfM,Olsson11_StableSfmUnordered}.  

\noindent \textbf{Alignment to Ground Truth} To evaluate each SfM result, we estimate a $\mathrm{Sim}(3)$ transformation between the estimated and ground truth camera poses, by Karcher mean and
\cite{Zinsser05icprip_PointSetRegistration}.

\noindent We evaluate several metrics per each front-end combination. 

\noindent \textbf{Relative Rotation Angular Error}: Defined over all image pairs $(i,j)$ as $\theta_{\mathrm{rel. rot. error}} = \| \log \Big( {}^{j} \overline{\mathbf{R}}_{i}^{\top} \circ {}^{j} \mathbf{R}_{i} \Big)^\vee \|_2$, measured in degrees, where $\overline{\mathbf{R}}$ represents the \textit{ground truth}.

\noindent \textbf{Relative Translation Angular Error}: Defined over all image pairs as $\theta_{\mathrm{rel. trans. error}} = \cos^{-1}\Big( \frac{ {}^{j} \mathbf{\overline{t}}_{i} \cdot {}^{j} \mathbf{t}_{i} }{\| {}^{j} \mathbf{\overline{t}}_{i} \| \| {}^{j} \mathbf{t}_{i} \| } \Big)$. 

\noindent \textbf{Relative Pose Error Deg.}: Defined over all image pairs as $\max (\theta_{\mathrm{rel. rot. error}}, \theta_{\mathrm{rel. trans. error}})$, per \cite{Jin21ijcv_PaperToPractice}.

\noindent \textbf{\# Registered Cameras}: The number of cameras for which a global pose is estimated. 

\noindent \textbf{\# Tracks (Filtered)}: Defined as the total number of keypoint tracks $j$ over the entire dataset.

\noindent \textbf{3d Track Length (Filtered)}: Defined as the number of views in each keypoint track $j$, over all tracks.

\noindent \textbf{Track Avg. Reprojection Error (Filtered)}: Defined as $\| \Pi({}^i\mathbf{R}_w \mathbf{P}_j + {}^i\mathbf{t}_w ; \mathbf{C}_i)  - \mathbf{p}_u \|_2$, for 3d point $\mathbf{P}_j$ from track $j$, view $i$, keypoint detection $\mathbf{p}_u$, and optimized camera pose $({}^w \mathbf{R}_i, {}^w \mathbf{t}_i)$, in the notation of Eqn. \ref{eqn:ba} (measured in pixels).

\noindent \textbf{Global Rotation Angular Error}: Defined over all camera poses as $\theta_{\mathrm{global. rot. error}} = \| \log \Big( {}^w \mathbf{\overline{R}}_{i}^{\top} \circ {}^w \mathbf{R}_{i} \Big)^\vee \|_2$, in deg.

\noindent \textbf{Global Translation Angular Error}: Defined as $\theta_{\mathrm{global. trans. error}} = \cos^{-1}\Big( \frac{ {}^{w} \mathbf{\overline{t}}_{i} \cdot {}^{w} \mathbf{t}_{i} }{\| {}^{w} \mathbf{\overline{t}}_{i} \| \| {}^{w} \mathbf{t}_{i} \| } \Big)$, and measured in degrees, over all camera poses.

\noindent \textbf{Global Pose AUC} A high-quality SfM metric must jointly optimize both recall and precision, as either can be traded off for the other. Accordingly, we use a variant of Pose AUC   \cite{Sarlin20cvpr_SuperGlue, Sun21cvpr_LoFTR, Lindenberger23iccv_LightGlue}, but adapt it for global poses, instead of relative poses. We compute the pose error as the maximum angular error in rotation and translation and report its area under the curve (AUC) at $1^{\circ}$, $2.5^{\circ}$, $5^{\circ}$, $10^{\circ}$, and $20^{\circ}$.

\subsection{Discussion}

\noindent \textbf{What is the best existing front-end for global SfM?} Against the findings of many recent works on relative pose estimation \cite{Sun21cvpr_LoFTR,Sarlin20cvpr_SuperGlue,Lindenberger23iccv_LightGlue}, we find the most accurate for global SfM is SIFT (see Tab. \ref{tab:avg-front-end-perf}). However, LoFTR is a close second, and offers more point density than SIFT (see Fig. \ref{fig:qual-visual}).

\noindent \textbf{Sources of error in global SfM} As discussed in Section \ref{ss:outlier_rejection}, performance of our global SfM approach is sensitive to noise from two-view pose estimates produced by the front-end. A few failure cases can be see in Figure \ref{fig:qual-visual}, such as LoFTR on Palace of Fine Arts dataset, and SuperGlue and LightGlue on the South Building dataset, where building corners are clearly not at $90^\circ$ angles (see Figure \ref{fig:doppelganger-examples} for outlier 2-view pose estimates that lead to this failure).

\noindent \textbf{How does global SfM compare to SOTA incremental SfM?} We show that GTSfM can yield a solution close to the COLMAP solution for almost all datasets. On all datasets in our experiments, we compare GTSfM to COLMAP. 

\noindent \textbf{What metric captures global SfM quality?} Global Pose AUC. In general, catastrophic front-end average errors (i.e. the presence of significant outliers) indicates that a high quality solution will not be recoverable from back-end optimization. Among the many back-end optimization metrics we compute and analyze, we find that only two consistently correlate closely with the visual quality of the reconstruction -- the average global translation angular error (indicating the correct relative placement of cameras, as a measure of precision) and the number of cameras localized by global SfM (recall), both represented in Global Pose AUC. In practice, we find that reprojection error can be low after bundle adjustment even when the result is qualitatively poor; in addition, if only filtered tracks are evaluated, one can set an arbitrarily low algorithmic acceptance threshold for track reprojection error.\\
%
\begin{figure}[t!]
    \centering
    \includegraphics[width=\linewidth]{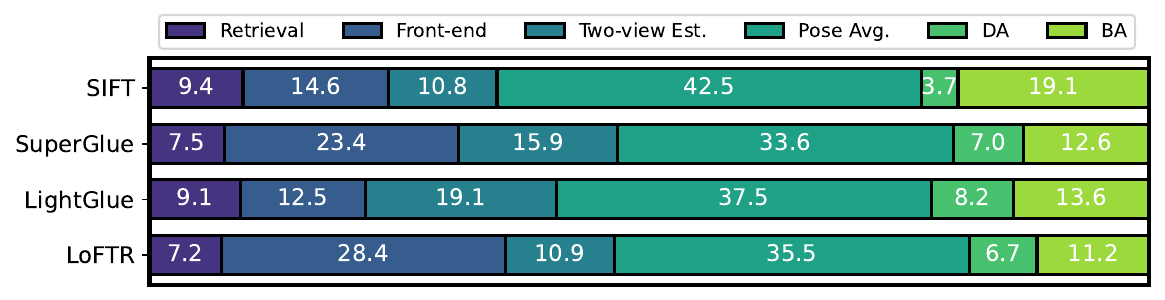}
    \caption{\textit{Relative} runtimes (as percentage of the total runtime) on the Crane Mast dataset for each module: Retrieval, Front-end, Two-view Estimation, Data Association (DA), Pose Averaging, and Global Bundle Adjustment (BA). Values are for 4 machines with 1 worker per machine.}
    \label{fig:relative-runtime}
\end{figure}
\begin{figure}[t!]
    \centering
    \includegraphics[width=0.8\linewidth]{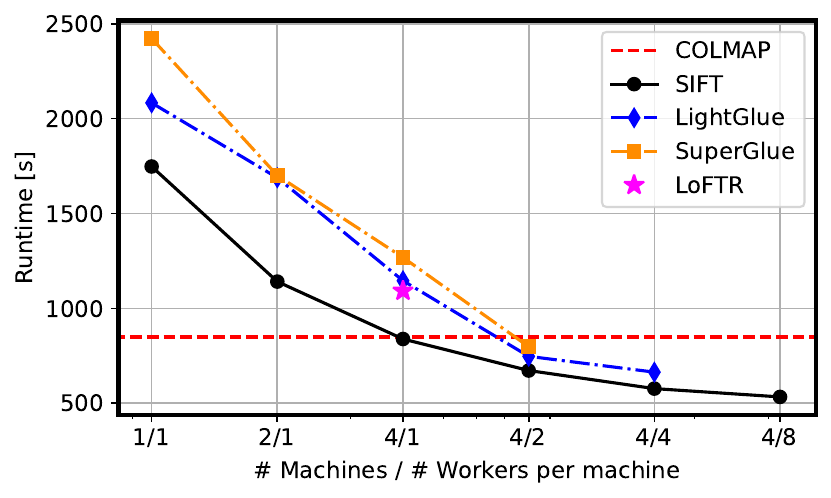}
    \caption{\textit{Absolute} runtimes on the Crane Mast dataset. Missing points indicate the front-end could not be run with more workers due to memory limitations (except LoFTR, which was only run with 4/1).}
    \label{fig:absolute-runtime}
\end{figure}
\noindent\textbf{Runtime Analysis} 
Since global SfM eliminates the need for repetitive local bundle adjustment, we expect it to be faster than its incremental counterpart. 
In Figures \ref{fig:relative-runtime} and \ref{fig:absolute-runtime}, we analyze the relative runtimes for each module and how absolute runtime scales with the number of machines and workers, respectively. 
COLMAP takes $\sim$$1.6\times$ as long on the Crane Mast dataset (533 sec. for GTSfM versus to 849 sec. for COLMAP) compared to GTSfM run on 4 machines with 8 workers per machine. 
Moreover, we are even able to demonstrate runtime improvement over COLMAP, which operates on \textit{handcrafted} SIFT features, with the more computationally expensive deep features (i.e., SuperGlue and LightGlue), as well as comparable runtimes using LoFTR, owing to the increased computational capacity offered by a cluster as opposed to a single machine. 
It is expected that the runtime savings would increase on larger datasets and by utilizing more machines.

\begin{table*}[!h]
    \centering
    \caption{Ablation experiments showing the effect of system modifications on Global Pose AUC (averaged over 9 datasets).}
    \begin{adjustbox}{width=\linewidth}
\begingroup
\begin{tabular}{llllllll}
\toprule
\rowcolorize            Module  & Ablation Description              & Front-End & @ 1 deg. & @ 2.5 deg. & @ 5 deg. & @ 10 deg. & @ 20 deg. \\
\midrule
All           & GTSfM Defaults                    & LoFTR     & 40.0      & 58.0        & 70.8    & 80.3     & 86.2             \\
              & (Full System)                     & SIFT      & 53.1    & 67.7      & 76.5    & 84.3     & 90.3      \\
\rowcolorize Front-End     & without 2-view                    & LoFTR     & 21.7 (\textcolor{red}{-45.8\%})   & 38.9 (\textcolor{red}{-32.9\%})     & 53.0 (\textcolor{red}{-25.2\%})   & 65.0 (\textcolor{red}{-19.1\% })    & 75.1 (\textcolor{red}{-12.9\%})       \\
\rowcolorize              & bundle adjustment                 & SIFT      & 45.9 (\textcolor{red}{-13.6\%})   & 60.2 (\textcolor{red}{-11.0\%})     & 71.8 (\textcolor{red}{ -6.2\%})   & 80.4 (\textcolor{red}{ -4.6\% })    & 85.3 (\textcolor{red}{ -5.6\%})        \\
              & without 2-view BA                 & LoFTR     & 31.4 (\textcolor{red}{-21.5\%})   & 45.2 (\textcolor{red}{-22.1\%})     & 57.4 (\textcolor{red}{-18.9\%})   & 69.6 (\textcolor{red}{-13.3\% })    & 78.8 (\textcolor{red}{ -8.6\%})        \\
              & ILS check                         & SIFT      & 47.2 (\textcolor{red}{-11.0\%})   & 66.3 (\textcolor{red}{ -2.1\%})     & 79.5 (\textcolor{forestgreen}{  +3.9\%})   & 88.1 (\textcolor{forestgreen}{  +4.4\% })    & 92.6 (\textcolor{forestgreen}{  +2.6\%})       \\
\rowcolorize View Graph /  & with min cycle                    & LoFTR     & 30.0 (\textcolor{red}{-25.0\%})   & 48.0 (\textcolor{red}{-17.2\%})     & 63.8 (\textcolor{red}{ -9.9\%})   & 76.4 (\textcolor{red}{ -4.8\% })    & 84.3 (\textcolor{red}{ -2.2\%})        \\
\rowcolorize Rotation Avg. & error filtering only              & SIFT      & 53.4 (\textcolor{forestgreen}{  +0.5\%})   & 67.2 (\textcolor{red}{ -0.7\%})     & 75.8 (\textcolor{red}{ -0.9\%})   & 83.3 (\textcolor{red}{ -1.2\% })    & 88.9 (\textcolor{red}{ -1.6\%})        \\
              & with median cycle                 & LoFTR     & 32.6 (\textcolor{red}{-18.6\%})   & 48.1 (\textcolor{red}{-17.0\%})     & 58.2 (\textcolor{red}{-17.7\%})   & 65.5 (\textcolor{red}{-18.4\% })    & 70.2 (\textcolor{red}{-18.5\%})       \\
              & error filtering only              & SIFT      & 52.8 (\textcolor{red}{ -0.6\%})   & 69.6 (\textcolor{forestgreen}{  +2.8\%})     & 80.8 (\textcolor{forestgreen}{  +5.6\%})   & 88.4 (\textcolor{forestgreen}{  +4.9\% })    & 92.5 (\textcolor{forestgreen}{  +2.5\%})       \\
\rowcolorize Translation   & without 1dsfm                     & LoFTR     & 40.2 (\textcolor{forestgreen}{  +0.5\%})   & 57.0 (\textcolor{red}{ -1.6\%})     & 70.6 (\textcolor{red}{ -0.2\%})   & 80.1 (\textcolor{red}{ -0.2\% })    & 85.8 (\textcolor{red}{ -0.5\%})        \\
\rowcolorize Averaging     & outlier rejection (MFAS)          & SIFT      & 54.9 (\textcolor{forestgreen}{  +3.4\%})   & 68.6 (\textcolor{forestgreen}{  +1.3\%})     & 77.0 (\textcolor{forestgreen}{  +0.7\%})   & 84.9 (\textcolor{forestgreen}{  +0.8\% })    & 90.5 (\textcolor{forestgreen}{  +0.2\%})       \\

 & without using tracks (only poses) & LoFTR & 26.7 (\textcolor{red}{-33.3\%}) & 38.0 (\textcolor{red}{-34.4\%}) & 44.7 (\textcolor{red}{-36.8\%}) & 51.6 (\textcolor{red}{-35.8\%}) & 57.2 (\textcolor{red}{-33.6\%})  \\
 &                                   & SIFT  & 23.7 (\textcolor{red}{-55.4\%}) & 36.0 (\textcolor{red}{-46.8\%}) & 48.0 (\textcolor{red}{-37.3\%}) & 60.8 (\textcolor{red}{-27.9\%}) & 70.4 (\textcolor{red}{-22.0\%})  \\

\rowcolorize              & without Huber loss                     & LoFTR     & 34.6 (\textcolor{red}{-13.6\%})   & 52.7 (\textcolor{red}{ -9.2\%})     & 64.1 (\textcolor{red}{ -9.5\%})   & 72.9 (\textcolor{red}{ -9.3\% })    & 79.9 (\textcolor{red}{ -7.3\%})        \\
\rowcolorize              &                                   & SIFT      & 44.8 (\textcolor{red}{-15.6\%})   & 59.2 (\textcolor{red}{-12.6\%})     & 67.5 (\textcolor{red}{-11.8\%})   & 75.3 (\textcolor{red}{-10.7\% })    & 81.4 (\textcolor{red}{ -9.9\%})        \\

 Bundle              & allow length-2 tracks             & LoFTR     & 30.6 (\textcolor{red}{-23.5\%})   & 51.1 (\textcolor{red}{-11.9\%})     & 66.2 (\textcolor{red}{ -6.4\%})   & 77.7 (\textcolor{red}{ -3.2\% })    & 84.8 (\textcolor{red}{ -1.6\%})        \\
 Adjustment              &                                   & SIFT      & 50.8 (\textcolor{red}{ -4.3\%})   & 66.7 (\textcolor{red}{ -1.5\%})     & 75.9 (\textcolor{red}{ -0.8\%})   & 82.0 (\textcolor{red}{ -2.7\% })    & 86.3 (\textcolor{red}{ -4.5\%})       \\
\rowcolorize        & without multi-stage BA     & LoFTR & 39.9 (\textcolor{red}{-0.1\%}) & 58.4 (\textcolor{forestgreen}{+0.6\% }) & 72.5 (\textcolor{forestgreen}{+2.4 }) & 81.5 (\textcolor{forestgreen}{+1.5\% }) & 87.2 (\textcolor{forestgreen}{+1.2\% }) \\
\rowcolorize & (use 1 stage) & SIFT  & 50.2 (\textcolor{red}{-5.5\%}) & 66.0 (\textcolor{red}{-2.6\%}) & 75.2 (\textcolor{red}{-1.8\%}) & 82.1 (\textcolor{red}{-2.6\%}) & 87.8 (\textcolor{red}{-2.8\%}) \\
              \bottomrule
\end{tabular}
\endgroup
\end{adjustbox}
\label{tab:ablations}
\end{table*}
\noindent \textbf{Ablation Experiments}
We examine the influence of each system component to determine which design decisions affect system performance most significantly (see Table \ref{tab:ablations}). For these ablations, we analyze only the two top-performing front-ends: SIFT \cite{Lowe04ijcv_SIFT} and LoFTR \cite{Sun21cvpr_LoFTR}. 

We use the average performance over 9 evaluation datasets to guide our analysis (with number of images provided in parentheses): \textit{Gerrard Hall} (100), \textit{South Building} (128), \textit{Skydio-Crane-Mast} (501),  Tanks and Temples \textit{Barn} (410), \textit{RC3} Astrovision (65) , Olsson's \textit{Palace of Fine Arts, San Francisco} (281), \textit{Ecole Superieure de Guerre} (35), \textit{King's College, Cambridge} (328), and \textit{Skansen-Kronan, Gothenburg, Sweden} (131).

We find that the use of 2-view bundle adjustment, enforcement of a minimum length of 3 or more for tracks in bundle adjustment, and utilization of camera-to-landmark (track) directions along with a Huber loss in translation averaging are most important. Use of the Minimum Feedback Arc Set (MFAS) outlier rejection from \cite{Wilson14eccv_1DSfM} appears to be least important.


\section{Conclusion}

In this work, we revisit the global SfM paradigm by leveraging both deep and classical front-ends.

\noindent \textbf{Current Limitations}: We observe lower accuracy on very large datasets with repetitive structures. SIFT performs best, and our ground truth is generated by SIFT-based methods, e.g. COLMAP; however, we still see qualitatively that improvements on our Pose AUC metric agree with observed quality. We assumed good camera intrinsics priors, which may not be readily available in all settings or applications. We tested only on a small-scale cluster (4 nodes in total).

\noindent \textbf{Acknowledgments}
We thank Yanwei Du, Ren Liu, Aditya Singh, Neha Upadhyay, Aishwarya Venkataramanan, Sushmita Warrier, Jon Womack, Jing Wu, Xiaolong Wu, and Kevin Fu for their code contributions. We are grateful to Fan Jiang and Varun Agrawal for their contributions to GTSAM which have been used in GTSfM. We thank James Hays for his financial support of the project.
This work was partially supported by a NASA Space Technology Graduate Research Opportunity.

{
    \small
    \bibliographystyle{ieeenat_fullname}
    \bibliography{main}

\begin{thebibliography}{76}
\providecommand{\natexlab}[1]{#1}
\providecommand{\url}[1]{\texttt{#1}}
\expandafter\ifx\csname urlstyle\endcsname\relax
  \providecommand{\doi}[1]{doi: #1}\else
  \providecommand{\doi}{doi: \begingroup \urlstyle{rm}\Url}\fi

\bibitem[col()]{colmap_dataset}
\url{https://colmap.github.io/datasets.html}.

\bibitem[Alayrac et~al.(2022)Alayrac, Donahue, Luc, Miech, Barr, Hasson, Lenc,
  Mensch, Millican, Reynolds, et~al.]{Alayrac22neurips_Flamingo}
Jean-Baptiste Alayrac, Jeff Donahue, Pauline Luc, Antoine Miech, Iain Barr,
  Yana Hasson, Karel Lenc, Arthur Mensch, Katherine Millican, Malcolm Reynolds,
  et~al.
\newblock Flamingo: a visual language model for few-shot learning.
\newblock \emph{Advances in Neural Information Processing Systems},
  35:\penalty0 23716--23736, 2022.

\bibitem[Arandjelovic et~al.(2016)Arandjelovic, Gronat, Torii, Pajdla, and
  Sivic]{Arandjelovic16cvpr_NetVLAD}
Relja Arandjelovic, Petr Gronat, Akihiko Torii, Tomas Pajdla, and Josef Sivic.
\newblock Netvlad: Cnn architecture for weakly supervised place recognition.
\newblock In \emph{CVPR}, 2016.

\bibitem[Barath and Matas(2018)]{Barath18cvpr_GCRANSAC}
Daniel Barath and Jiří Matas.
\newblock Graph-cut ransac.
\newblock In \emph{Proceedings of the IEEE Conference on Computer Vision and
  Pattern Recognition (CVPR)}, 2018.

\bibitem[Cai et~al.(2023)Cai, Tung, Wang, Averbuch-Elor, Hariharan, and
  Snavely]{Cai23iccv_Doppelgangers}
Ruojin Cai, Joseph Tung, Qianqian Wang, Hadar Averbuch-Elor, Bharath Hariharan,
  and Noah Snavely.
\newblock Doppelgangers: Learning to disambiguate images of similar structures.
\newblock In \emph{ICCV}, 2023.

\bibitem[Chen et~al.(2022)Chen, Wang, Changpinyo, Piergiovanni, Padlewski,
  Salz, Goodman, Grycner, Mustafa, Beyer, et~al.]{Chen22arxiv_PALI}
Xi Chen, Xiao Wang, Soravit Changpinyo, AJ Piergiovanni, Piotr Padlewski,
  Daniel Salz, Sebastian Goodman, Adam Grycner, Basil Mustafa, Lucas Beyer,
  et~al.
\newblock Pali: A jointly-scaled multilingual language-image model.
\newblock \emph{arXiv preprint arXiv:2209.06794}, 2022.

\bibitem[Chen et~al.(2020)Chen, Shen, Chen, and
  Wang]{Chen20pr_GraphBasedParallelSfM}
Yu Chen, Shuhan Shen, Yisong Chen, and Guoping Wang.
\newblock Graph-based parallel large scale structure from motion.
\newblock \emph{Pattern Recognition}, page 107537, 2020.

\bibitem[Cherti et~al.(2023)Cherti, Beaumont, Wightman, Wortsman, Ilharco,
  Gordon, Schuhmann, Schmidt, and Jitsev]{Cherti23cvpr_ReproducibleScalingLaws}
Mehdi Cherti, Romain Beaumont, Ross Wightman, Mitchell Wortsman, Gabriel
  Ilharco, Cade Gordon, Christoph Schuhmann, Ludwig Schmidt, and Jenia Jitsev.
\newblock Reproducible scaling laws for contrastive language-image learning.
\newblock In \emph{CVPR}, 2023.

\bibitem[Chirikjian(2011)]{chirikjian2011}
G.~S. Chirikjian.
\newblock \emph{Stochastic Models, Information Theory, and Lie Groups: Analytic
  Methods and Modern Applications}.
\newblock Springer Science+Business Media, 2011.

\bibitem[Chowdhery et~al.(2022)Chowdhery, Narang, Devlin, Bosma, Mishra,
  Roberts, Barham, Chung, Sutton, Gehrmann, et~al.]{Chowdhery22arxiv_Palm}
Aakanksha Chowdhery, Sharan Narang, Jacob Devlin, Maarten Bosma, Gaurav Mishra,
  Adam Roberts, Paul Barham, Hyung~Won Chung, Charles Sutton, Sebastian
  Gehrmann, et~al.
\newblock Palm: Scaling language modeling with pathways.
\newblock \emph{arXiv preprint arXiv:2204.02311}, 2022.

\bibitem[Crandall et~al.(2011)Crandall, Owens, Snavely, and
  Huttenlocher]{Crandall11cvpr_DISCO}
David Crandall, Andrew Owens, Noah Snavely, and Dan Huttenlocher.
\newblock Discrete-continuous optimization for large-scale structure from
  motion.
\newblock In \emph{CVPR 2011}, pages 3001--3008, 2011.

\bibitem[Cui et~al.(2015)Cui, Jiang, Tang, and
  Tan]{Cui15bmvc_LinearGlobTransEstFeatTracks}
Zhaopeng Cui, Nianjuan Jiang, Chengzhou Tang, and Ping Tan.
\newblock Linear global translation estimation with feature tracks.
\newblock In \emph{BMVC}, 2015.

\bibitem[Dellaert(2012)]{Dellaert12_GTSAM}
Frank Dellaert.
\newblock Factor graphs and {GTSAM}: A hands-on introduction.
\newblock Technical report, Georgia Institute of Technology, 2012.

\bibitem[Dellaert and Contributors(2022)]{Dellaert22software_GTSAM}
Frank Dellaert and GTSAM Contributors.
\newblock borglab/gtsam, 2022.

\bibitem[Dellaert et~al.(2020)Dellaert, Rosen, Wu, Mahony, and
  Carlone]{Dellaert20eccv_Shonan}
Frank Dellaert, David~M. Rosen, Jing Wu, Robert~E. Mahony, and Luca Carlone.
\newblock Shonan rotation averaging: Global optimality by surfing
  \({SO(p)}^{\mbox{n}}\).
\newblock In \emph{ECCV}, 2020.

\bibitem[DeTone et~al.(2018)DeTone, Malisiewicz, and
  Rabinovich]{Detone18cvprw_SuperPoint}
Daniel DeTone, Tomasz Malisiewicz, and Andrew Rabinovich.
\newblock Superpoint: Self-supervised interest point detection and description.
\newblock In \emph{The IEEE Conference on Computer Vision and Pattern
  Recognition (CVPR) Workshops}, 2018.

\bibitem[Driver et~al.(2023)Driver, Skinner, Dor, and
  Tsiotras]{driver2023astrovision}
Travis Driver, Katherine~A Skinner, Mehregan Dor, and Panagiotis Tsiotras.
\newblock {AstroVision}: Towards autonomous feature detection and description
  for missions to small bodies using deep learning.
\newblock \emph{Acta Astronautica, Special Issue on AI for Space}, 2023.

\bibitem[Edelsbrunner et~al.(1983)Edelsbrunner, Kirkpatrick, and
  Seidel]{Edelsbrunner83transit_AlphaShape}
Herbert Edelsbrunner, David Kirkpatrick, and Raimund Seidel.
\newblock On the shape of a set of points in the plane.
\newblock \emph{IEEE Transactions on information theory}, 29\penalty0
  (4):\penalty0 551--559, 1983.

\bibitem[Enqvist et~al.(2011{\natexlab{a}})Enqvist, Kahl, and
  Olsson]{Enqvist11iccvw_NonsequentialSfM}
Olof Enqvist, Fredrik Kahl, and Carl Olsson.
\newblock Non-sequential structure from motion.
\newblock In \emph{2011 IEEE International Conference on Computer Vision
  Workshops (ICCV Workshops)}, pages 264--271, 2011{\natexlab{a}}.

\bibitem[Enqvist et~al.(2011{\natexlab{b}})Enqvist, Olsson, and
  Kahl]{Enqvist11tr_StableSfMRotConsistency}
Olof Enqvist, Carl Olsson, and Fredrik Kahl.
\newblock Stable structure from motion using rotational consistency.
\newblock 2011{\natexlab{b}}.

\bibitem[Gargallo et~al.(2016)Gargallo, Kuang,
  et~al.]{Gargallo16github_OpenSfM}
P Gargallo, Y Kuang, et~al.
\newblock {OpenSfM}, 2016.

\bibitem[Goldstein et~al.(2016)Goldstein, Hand, Lee, Voroninski, and
  Soatto]{Goldstein16eccv_Shapefit}
Thomas Goldstein, Paul Hand, Choongbum Lee, Vladislav Voroninski, and Stefano
  Soatto.
\newblock Shapefit and shapekick for robust, scalable structure from motion.
\newblock In \emph{ECCV}, 2016.

\bibitem[Govindu(2001)]{Govindu01cvpr_TwoViewConstraintsMotionEstimation}
Venu~Madhav Govindu.
\newblock Combining two-view constraints for motion estimation.
\newblock In \emph{CVPR}, pages II--II. IEEE, 2001.

\bibitem[Govindu(2004)]{Govindu04cvpr_LieAlgebraicAveraging}
Venu~Madhav Govindu.
\newblock Lie-algebraic averaging for globally consistent motion estimation.
\newblock In \emph{CVPR}, 2004.

\bibitem[Govindu(2006)]{Govindu06accv_RobustnessInMotionAveraging}
Venu~Madhav Govindu.
\newblock Robustness in motion averaging.
\newblock In \emph{Asian Conference on Computer Vision}, pages 457--466.
  Springer, 2006.

\bibitem[Hartley et~al.(2013)Hartley, Trumpf, Dai, and
  Li]{Hartley13ijcv_RotationAveraging}
Richard~I. Hartley, Jochen Trumpf, Yuchao Dai, and Hongdong Li.
\newblock Rotation averaging.
\newblock \emph{Int. J. Comput. Vis.}, 103\penalty0 (3):\penalty0 267--305,
  2013.

\bibitem[He et~al.(2021)He, Wang, Sun, Shen, Bao, and
  Zhou]{He21tr_LoftrImwChallenge}
Xingyi He, Yuang Wang, Jiaming Sun, Zehong Shen, Hujun Bao, and Xiaowei Zhou.
\newblock Tech details for loftr in the imw challenge, 2021.

\bibitem[He et~al.(2023)He, Sun, Wang, Peng, Huang, Bao, and
  Zhou]{He23arxiv_DetectorFreeSfm}
Xingyi He, Jiaming Sun, Yifan Wang, Sida Peng, Qixing Huang, Hujun Bao, and
  Xiaowei Zhou.
\newblock Detector-free structure from motion.
\newblock In \emph{arxiv}, 2023.

\bibitem[Jiang et~al.(2013)Jiang, Cui, and
  Tan]{Jiang13iccv_GlobalLinearMethodCamPoseReg}
Nianjuan Jiang, Zhaopeng Cui, and Ping Tan.
\newblock A global linear method for camera pose registration.
\newblock In \emph{ICCV}, 2013.

\bibitem[Jin et~al.(2021)Jin, Mishkin, Mishchuk, Matas, Fua, Yi, and
  Trulls]{Jin21ijcv_PaperToPractice}
Yuhe Jin, Dmytro Mishkin, Anastasiia Mishchuk, Jiri Matas, Pascal Fua,
  Kwang~Moo Yi, and Eduard Trulls.
\newblock Image matching across wide baselines: From paper to practice.
\newblock \emph{International Journal of Computer Vision}, 129\penalty0
  (2):\penalty0 517--547, 2021.

\bibitem[Juli{\`a} and
  Monasse(2017)]{Julia17psivt_CriticalReviewTrifocalTensor}
Laura~F Juli{\`a} and Pascal Monasse.
\newblock A critical review of the trifocal tensor estimation.
\newblock In \emph{Pacific-Rim Symposium on Image and Video Technology}, pages
  337--349. Springer, 2017.

\bibitem[Kerbl et~al.(2023)Kerbl, Kopanas, Leimk{\"u}hler, and
  Drettakis]{Kerbl23acmtog_GaussianSplat}
Bernhard Kerbl, Georgios Kopanas, Thomas Leimk{\"u}hler, and George Drettakis.
\newblock {3D} gaussian splatting for real-time radiance field rendering.
\newblock \emph{ACM Transactions on Graphics (ToG)}, 42\penalty0 (4):\penalty0
  1--14, 2023.

\bibitem[Klingner et~al.(2013)Klingner, Martin, and
  Roseborough]{Klingner13iccv_StreetViewSfM}
Bryan Klingner, David Martin, and James Roseborough.
\newblock Street view motion-from-structure-from-motion.
\newblock In \emph{ICCV}, 2013.

\bibitem[Knapitsch et~al.(2017)Knapitsch, Park, Zhou, and
  Koltun]{Knapitsch17acmtg_TanksTemples}
Arno Knapitsch, Jaesik Park, Qian-Yi Zhou, and Vladlen Koltun.
\newblock Tanks and temples: Benchmarking large-scale scene reconstruction.
\newblock \emph{ACM Transactions on Graphics}, 36\penalty0 (4), 2017.

\bibitem[Lambert(2022)]{Lambert22thesis}
John Lambert.
\newblock \emph{Deep Learning for Building and Validating Geometric and
  Semantic Maps}.
\newblock PhD thesis, Georgia Institute of Technology, 2022.

\bibitem[Lambert et~al.(2022)Lambert, Li, Boyadzhiev, Wixson, Narayana,
  Hutchcroft, Hays, Dellaert, and Kang]{Lambert22eccv_salve}
John Lambert, Yuguang Li, Ivaylo Boyadzhiev, Lambert Wixson, Manjunath
  Narayana, Will Hutchcroft, James Hays, Frank Dellaert, and Sing~Bing Kang.
\newblock Salve: Semantic alignment verification for floorplan reconstruction
  from sparse panoramas.
\newblock In \emph{ECCV}, 2022.

\bibitem[Levenberg(1944)]{Levenberg44qam_MethodNonlinearLeastSquares}
Kenneth Levenberg.
\newblock A method for the solution of certain non-linear problems in least
  squares.
\newblock \emph{Quarterly of applied mathematics}, 2\penalty0 (2):\penalty0
  164--168, 1944.

\bibitem[Li and Snavely(2018)]{Li18cvpr_MegaDepth}
Zhengqi Li and Noah Snavely.
\newblock Megadepth: Learning single-view depth prediction from internet
  photos.
\newblock In \emph{CVPR}, 2018.

\bibitem[Li et~al.(2019)Li, Dekel, Cole, Tucker, Snavely, Liu, and
  Freeman]{Li19cvpr_DepthsFrozenPeople}
Zhengqi Li, Tali Dekel, Forrester Cole, Richard Tucker, Noah Snavely, Ce Liu,
  and William~T. Freeman.
\newblock Learning the depths of moving people by watching frozen people.
\newblock In \emph{CVPR}, 2019.

\bibitem[Lin et~al.(2014)Lin, Maire, Belongie, Hays, Perona, Ramanan,
  Doll{\'a}r, and Zitnick]{Lin14eccv_MSCOCO}
Tsung-Yi Lin, Michael Maire, Serge Belongie, James Hays, Pietro Perona, Deva
  Ramanan, Piotr Doll{\'a}r, and C~Lawrence Zitnick.
\newblock Microsoft coco: Common objects in context.
\newblock In \emph{ECCV}, 2014.

\bibitem[Lindenberger et~al.(2021)Lindenberger, Sarlin, Larsson, and
  Pollefeys]{Lindenberger21iccv_PixelPerfectSfM}
Philipp Lindenberger, Paul-Edouard Sarlin, Viktor Larsson, and Marc Pollefeys.
\newblock Pixel-perfect structure-from-motion with featuremetric refinement.
\newblock In \emph{ICCV}, 2021.

\bibitem[Lindenberger et~al.(2023)Lindenberger, Sarlin, and
  Pollefeys]{Lindenberger23iccv_LightGlue}
Philipp Lindenberger, Paul-Edouard Sarlin, and Marc Pollefeys.
\newblock {LightGlue: Local Feature Matching at Light Speed}.
\newblock In \emph{ICCV}, 2023.

\bibitem[Lowe(2004)]{Lowe04ijcv_SIFT}
David~G. Lowe.
\newblock Distinctive image features from scale-invariant keypoints.
\newblock \emph{International Journal of Computer Vision}, 60:\penalty0
  91--110, 2004.

\bibitem[Manam and Govindu(2022)]{Manam22eccv_CReTA}
Lalit Manam and Venu~Madhav Govindu.
\newblock Correspondence reweighted translation averaging.
\newblock In \emph{ECCV}, pages pp. 56--72, 2022.

\bibitem[Marquardt(1963)]{Marquardt63siam_AlgorithmLeastSquaresNonlinear}
Donald~W Marquardt.
\newblock An algorithm for least-squares estimation of nonlinear parameters.
\newblock \emph{Journal of the society for Industrial and Applied Mathematics},
  11\penalty0 (2):\penalty0 431--441, 1963.

\bibitem[Martinec and
  Pajdla(2007)]{Martinec07cvpr_RotationTranslationAveraging}
Daniel Martinec and Tomas Pajdla.
\newblock Robust rotation and translation estimation in multiview
  reconstruction.
\newblock In \emph{CVPR}, pages 1--8, 2007.

\bibitem[Mildenhall et~al.(2020)Mildenhall, Srinivasan, Tancik, Barron,
  Ramamoorthi, and Ng]{Mildenhall20eccv_NERF}
Ben Mildenhall, Pratul~P. Srinivasan, Matthew Tancik, Jonathan~T. Barron, Ravi
  Ramamoorthi, and Ren Ng.
\newblock Nerf: Representing scenes as neural radiance fields for view
  synthesis.
\newblock In \emph{ECCV}, 2020.

\bibitem[Moulon et~al.(2013)Moulon, Monasse, and
  Marlet]{Moulon13iccv_GlobalFusionSfM}
Pierre Moulon, Pascal Monasse, and Renaud Marlet.
\newblock Global fusion of relative motions for robust, accurate and scalable
  structure from motion.
\newblock In \emph{ICCV}, pages 3248--3255, 2013.

\bibitem[Moulon et~al.(2016)Moulon, Monasse, Perrot, and
  Marlet]{Moulon16iwrrpr_OpenMVG}
Pierre Moulon, Pascal Monasse, Romuald Perrot, and Renaud Marlet.
\newblock Openmvg: Open multiple view geometry.
\newblock In \emph{International Workshop on Reproducible Research in Pattern
  Recognition}, 2016.

\bibitem[Narayanan et~al.(2021)Narayanan, Shoeybi, Casper, LeGresley, Patwary,
  Korthikanti, Vainbrand, Kashinkunti, Bernauer, Catanzaro,
  et~al.]{Narayanan21ichpc_MegatronLM}
Deepak Narayanan, Mohammad Shoeybi, Jared Casper, Patrick LeGresley, Mostofa
  Patwary, Vijay Korthikanti, Dmitri Vainbrand, Prethvi Kashinkunti, Julie
  Bernauer, Bryan Catanzaro, et~al.
\newblock Efficient large-scale language model training on gpu clusters using
  megatron-lm.
\newblock In \emph{Proceedings of the International Conference for High
  Performance Computing, Networking, Storage and Analysis}, pages 1--15, 2021.

\bibitem[Nister(2004)]{Nister04tpami_FivePointRelativePose}
D. Nister.
\newblock An efficient solution to the five-point relative pose problem.
\newblock \emph{IEEE Transactions on Pattern Analysis and Machine
  Intelligence}, 26\penalty0 (6):\penalty0 756--770, 2004.

\bibitem[Olsson and Enqvist(2011)]{Olsson11_StableSfmUnordered}
Carl Olsson and Olof Enqvist.
\newblock Stable structure from motion for unordered image collections.
\newblock In \emph{Scandinavian Conference on Image Analysis}, pages 524--535.
  Springer, 2011.

\bibitem[Park et~al.(2021)Park, Sinha, Barron, Bouaziz, Goldman, Seitz, and
  Martin-Brualla]{Park21iccv_Nerfies}
Keunhong Park, Utkarsh Sinha, Jonathan~T Barron, Sofien Bouaziz, Dan~B Goldman,
  Steven~M Seitz, and Ricardo Martin-Brualla.
\newblock Nerfies: Deformable neural radiance fields.
\newblock In \emph{ICCV}, 2021.

\bibitem[Parra et~al.(2021)Parra, Chng, Chin, Eriksson, and
  Reid]{Parra21cvpr_RotationCoordinateDescent}
Alvaro Parra, Shin-Fang Chng, Tat-Jun Chin, Anders Eriksson, and Ian Reid.
\newblock Rotation coordinate descent for fast globally optimal rotation
  averaging.
\newblock In \emph{CVPR}, pages 4298--4307, 2021.

\bibitem[Phillips and
  Daniilidis(2019)]{Phillips19arxiv_GraphsLearningCycleConsistent}
Stephen Phillips and Kostas Daniilidis.
\newblock All graphs lead to {Rome}: Learning geometric and cycle-consistent
  representations with graph convolutional networks.
\newblock \emph{arXiv preprint arXiv:1901.02078}, 2019.

\bibitem[Pollefeys et~al.(2004)Pollefeys, Van~Gool, Vergauwen, Verbiest,
  Cornelis, Tops, and Koch]{Pollefeys04ijcv_VisualModeling}
Marc Pollefeys, Luc Van~Gool, Maarten Vergauwen, Frank Verbiest, Kurt Cornelis,
  Jan Tops, and Reinhard Koch.
\newblock Visual modeling with a hand-held camera.
\newblock \emph{International Journal of Computer Vision}, 59\penalty0
  (3):\penalty0 207--232, 2004.

\bibitem[Reizenstein et~al.(2021)Reizenstein, Shapovalov, Henzler, Sbordone,
  Labatut, and Novotny]{Reizenstein21iccv_CO3D}
Jeremy Reizenstein, Roman Shapovalov, Philipp Henzler, Luca Sbordone, Patrick
  Labatut, and David Novotny.
\newblock Common objects in {3D}: Large-scale learning and evaluation of
  real-life {3D} category reconstruction.
\newblock In \emph{ICCV}, 2021.

\bibitem[Rocklin et~al.(2015)]{Rocklin15scipy_Dask}
Matthew Rocklin et~al.
\newblock Dask: Parallel computation with blocked algorithms and task
  scheduling.
\newblock In \emph{Proceedings of the 14th python in science conference}, page
  136. SciPy Austin, TX, 2015.

\bibitem[Sarlin et~al.(2020)Sarlin, DeTone, Malisiewicz, and
  Rabinovich]{Sarlin20cvpr_SuperGlue}
Paul-Edouard Sarlin, Daniel DeTone, Tomasz Malisiewicz, and Andrew Rabinovich.
\newblock {SuperGlue}: Learning feature matching with graph neural networks.
\newblock In \emph{CVPR}, 2020.

\bibitem[Schonberger and Frahm(2016)]{Schonberger16cvpr_COLMAP}
Johannes~L. Schonberger and Jan-Michael Frahm.
\newblock Structure-from-motion revisited.
\newblock In \emph{{CVPR}}, 2016.

\bibitem[Sharp et~al.(2001)Sharp, Lee, and
  Wehe]{Sharp01icra_MultiviewCycleBasis}
G.C. Sharp, S.W. Lee, and D.K. Wehe.
\newblock Toward multiview registration in frame space.
\newblock In \emph{Proceedings 2001 ICRA. IEEE International Conference on
  Robotics and Automation}, pages 3542--3547 vol.4, 2001.

\bibitem[Snavely et~al.(2006)Snavely, Seitz, and
  Szeliski]{Snavely06siggraph_PhotoTourism}
Noah Snavely, Steven~M. Seitz, and Richard Szeliski.
\newblock Photo tourism: Exploring photo collections in {3D}.
\newblock In \emph{ACM SIGGRAPH 2006 Papers}, page 835–846, New York, NY,
  USA, 2006. Association for Computing Machinery.

\bibitem[Snavely et~al.(2008)Snavely, Seitz, and
  Szeliski]{Snavely08ijcv_ModelingTheWorld}
Noah Snavely, Steven~M. Seitz, and Richard Szeliski.
\newblock Modeling the world from internet photo collections.
\newblock \emph{Int. J. Comput. Vision}, 80\penalty0 (2):\penalty0 189–210,
  2008.

\bibitem[Sun et~al.(2021)Sun, Shen, Wang, Bao, and Zhou]{Sun21cvpr_LoFTR}
Jiaming Sun, Zehong Shen, Yuang Wang, Hujun Bao, and Xiaowei Zhou.
\newblock {LoFTR}: Detector-free local feature matching with transformers.
\newblock In \emph{CVPR}, pages 8922--8931, 2021.

\bibitem[Sweeney et~al.(2015{\natexlab{a}})Sweeney, Hollerer, and
  Turk]{Sweeney15acmicm_TheiaSfM}
Christopher Sweeney, Tobias Hollerer, and Matthew Turk.
\newblock Theia: A fast and scalable structure-from-motion library.
\newblock In \emph{Proceedings of the 23rd ACM international conference on
  Multimedia}, pages 693--696, 2015{\natexlab{a}}.

\bibitem[Sweeney et~al.(2015{\natexlab{b}})Sweeney, Sattler, Hollerer, Turk,
  and Pollefeys]{Sweeney15iccv_OptimizingViewingGraph}
Chris Sweeney, Torsten Sattler, Tobias Hollerer, Matthew Turk, and Marc
  Pollefeys.
\newblock Optimizing the viewing graph for structure-from-motion.
\newblock In \emph{ICCV}, 2015{\natexlab{b}}.

\bibitem[Touvron et~al.(2023)Touvron, Lavril, Izacard, Martinet, Lachaux,
  Lacroix, Rozi{\`e}re, Goyal, Hambro, Azhar, et~al.]{Touvron23arxiv_Llama}
Hugo Touvron, Thibaut Lavril, Gautier Izacard, Xavier Martinet, Marie-Anne
  Lachaux, Timoth{\'e}e Lacroix, Baptiste Rozi{\`e}re, Naman Goyal, Eric
  Hambro, Faisal Azhar, et~al.
\newblock Llama: Open and efficient foundation language models.
\newblock \emph{arXiv preprint arXiv:2302.13971}, 2023.

\bibitem[Tron et~al.(2016)Tron, Zhou, and
  Daniilidis]{Tron16cvprw_SurveyRotOptSfM}
Roberto Tron, Xiaowei Zhou, and Kostas Daniilidis.
\newblock A survey on rotation optimization in structure from motion.
\newblock In \emph{2016 IEEE Conference on Computer Vision and Pattern
  Recognition Workshops (CVPRW)}, 2016.

\bibitem[Vaswani et~al.(2017)Vaswani, Shazeer, Parmar, Uszkoreit, Jones, Gomez,
  Kaiser, and Polosukhin]{Vaswani17neurips_AttentionIsAllYouNeed}
Ashish Vaswani, Noam Shazeer, Niki Parmar, Jakob Uszkoreit, Llion Jones,
  Aidan~N Gomez, {\L}ukasz Kaiser, and Illia Polosukhin.
\newblock Attention is all you need.
\newblock \emph{Advances in neural information processing systems}, 30, 2017.

\bibitem[Wilson and Snavely(2014)]{Wilson14eccv_1DSfM}
Kyle Wilson and Noah Snavely.
\newblock Robust global translations with {1DSfM}.
\newblock In \emph{{ECCV}}, pages 61--75, 2014.

\bibitem[Wu(2013)]{Wu133dv_IncrementalSfM}
Changchang Wu.
\newblock Towards linear-time incremental structure from motion.
\newblock In \emph{International Conference on 3D Vision (3DV)}, pages
  127--134, 2013.

\bibitem[Zach et~al.(2010)Zach, Klopschitz, and
  Pollefeys]{Zach10cvpr_LoopConstraints}
Christopher Zach, Manfred Klopschitz, and Marc Pollefeys.
\newblock Disambiguating visual relations using loop constraints.
\newblock In \emph{CVPR}, pages 1426--1433, 2010.

\bibitem[Zhang et~al.(2023)Zhang, Larsson, and
  Barath]{Zhang23cvpr_RotAvgUncertaintyRobustLosses}
Ganlin Zhang, Viktor Larsson, and Daniel Barath.
\newblock Revisiting rotation averaging: Uncertainties and robust losses.
\newblock In \emph{CVPR}, 2023.

\bibitem[Zhou et~al.(2020)Zhou, Luo, Zhen, Shen, Li, Huang, Fang, and
  Quan]{Zhou20eccv_StochasticBundleAdjustment}
Lei Zhou, Zixin Luo, Mingmin Zhen, Tianwei Shen, Shiwei Li, Zhuofei Huang, Tian
  Fang, and Long Quan.
\newblock Stochastic bundle adjustment for efficient and scalable {3D}
  reconstruction.
\newblock In \emph{ECCV}, 2020.

\bibitem[Zhou et~al.(2018)Zhou, Park, and Koltun]{Zhou18arxiv_Open3D}
Qian-Yi Zhou, Jaesik Park, and Vladlen Koltun.
\newblock {Open3D}: {A} modern library for {3D} data processing.
\newblock \emph{arXiv:1801.09847}, 2018.

\bibitem[Zin{\ss}er et~al.(2005)Zin{\ss}er, Schmidt, and
  Niemann]{Zinsser05icprip_PointSetRegistration}
Timo Zin{\ss}er, Jochen Schmidt, and Heinrich Niemann.
\newblock Point set registration with integrated scale estimation.
\newblock In \emph{International conference on pattern recognition and image
  processing}, pages 116--119, 2005.

\end{thebibliography}
}

\vspace{5mm}
\appendix
\section{Appendix}

\textit{In this Appendix, we analyze system performance in a setting where an oracle `virtual' front-end provides noise-free correspondences and we provide additional implementation details.

    }

\subsection{Additional Implementation Details}
\noindent \textbf{Two-View Bundle Adjustment} We reject a two-view measurement if an indeterminate linear system is encountered during marginal covariance computation, as we find these are often tied to image pairs with no overlapping field of view.

\begin{figure}[!h]
    \begin{subfigure}[b]{0.5\textwidth}
        \centering
        \includegraphics[width=0.49\linewidth]{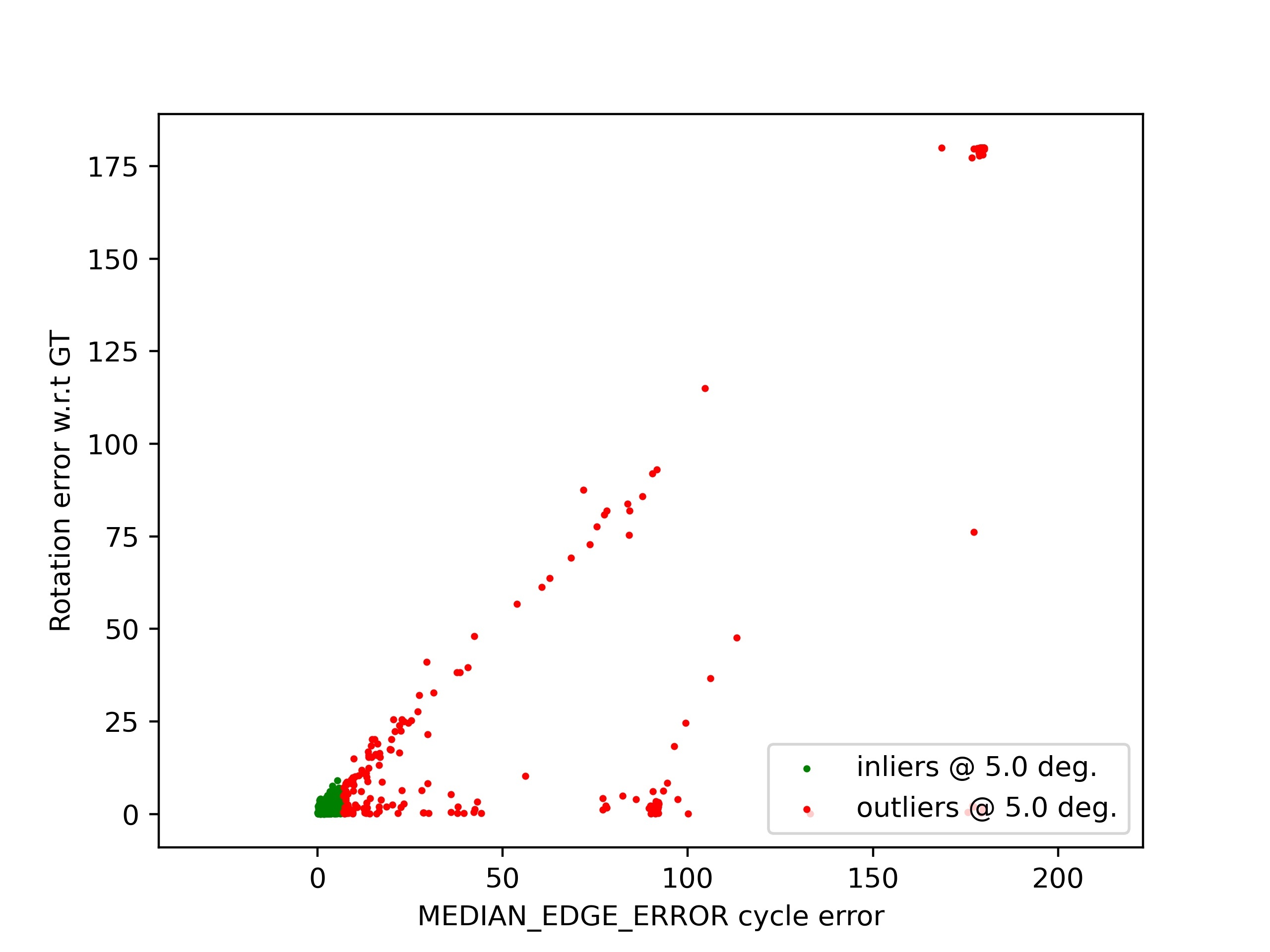}
        \caption{Single stage filtering}
    \end{subfigure}

    \vspace{5mm}

    \begin{subfigure}[b]{0.5\textwidth}
        \centering
        \includegraphics[width=0.49\linewidth]{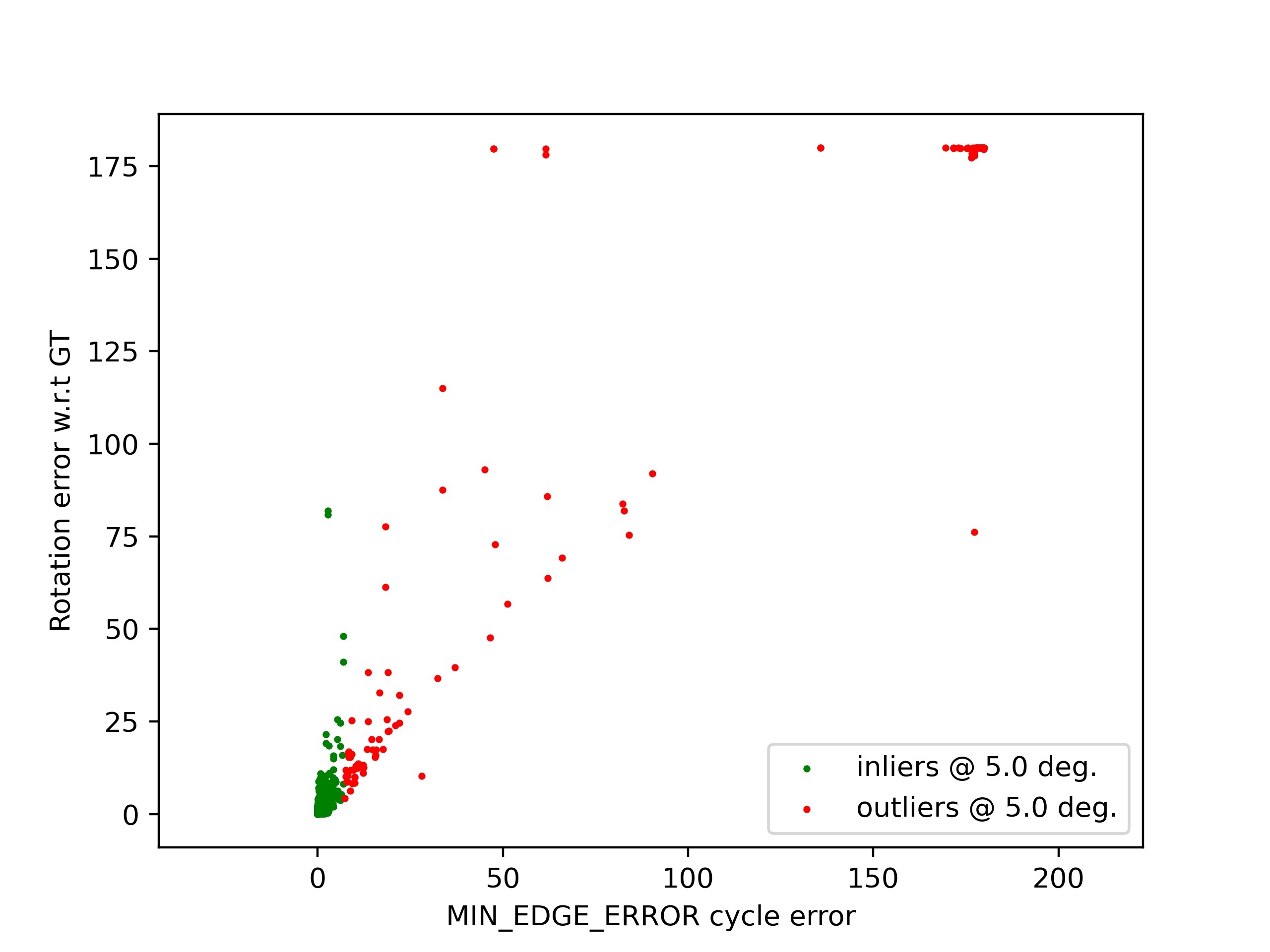}
        \includegraphics[width=0.49\linewidth]{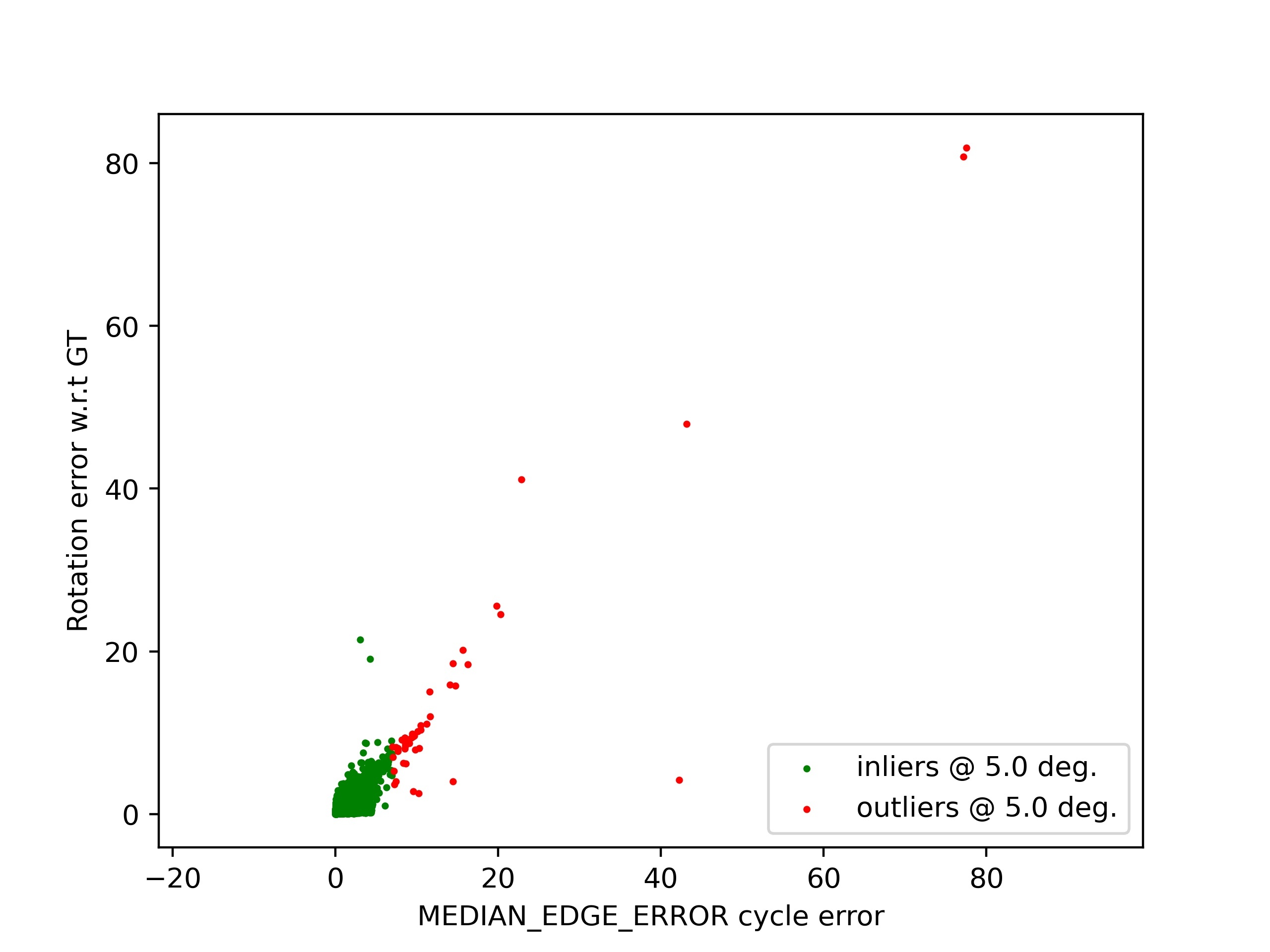}
        \caption{Two-stage filtering}
    \end{subfigure}
    \vspace{2mm}
    \caption{Visualization of edge $\mathcal{E}$ classification using cycle-consistency outlier rejection on the King's College, Cambridge dataset. \textbf{Top row:} Apply filtering by median cycle error  only. \textbf{Bottom two rows:} Use two stages, first filtering by minimum cycle error, and then by median. Utilizing two cascaded stages maximizes recall while retaining precision.}
    \label{fig:cycle-consistency-stages}
\end{figure}

\noindent\textbf{Cycle Consistency-based Outlier Rejection} We use a 7 degree  threshold $\epsilon_{cycle}$ when comparing against a triplet cycle error summary statistic (\texttt{min} or \texttt{median}). In Figure \ref{fig:cycle-consistency-stages}, for each edge we plot aggregated cycle errors versus ground truth rotation angular error for the corresponding relative rotation ${}^j\mathbf{R}_i$.

\noindent\textbf{Image Resolution} We run all experiments with 760 pixel resolution images, i.e. the shorter image side is resized to at most 760 pixels in length.

\begin{figure}
    \centering
    \begin{subfigure}[b]{0.5\textwidth}
        \centering
        \includegraphics[width=\linewidth]{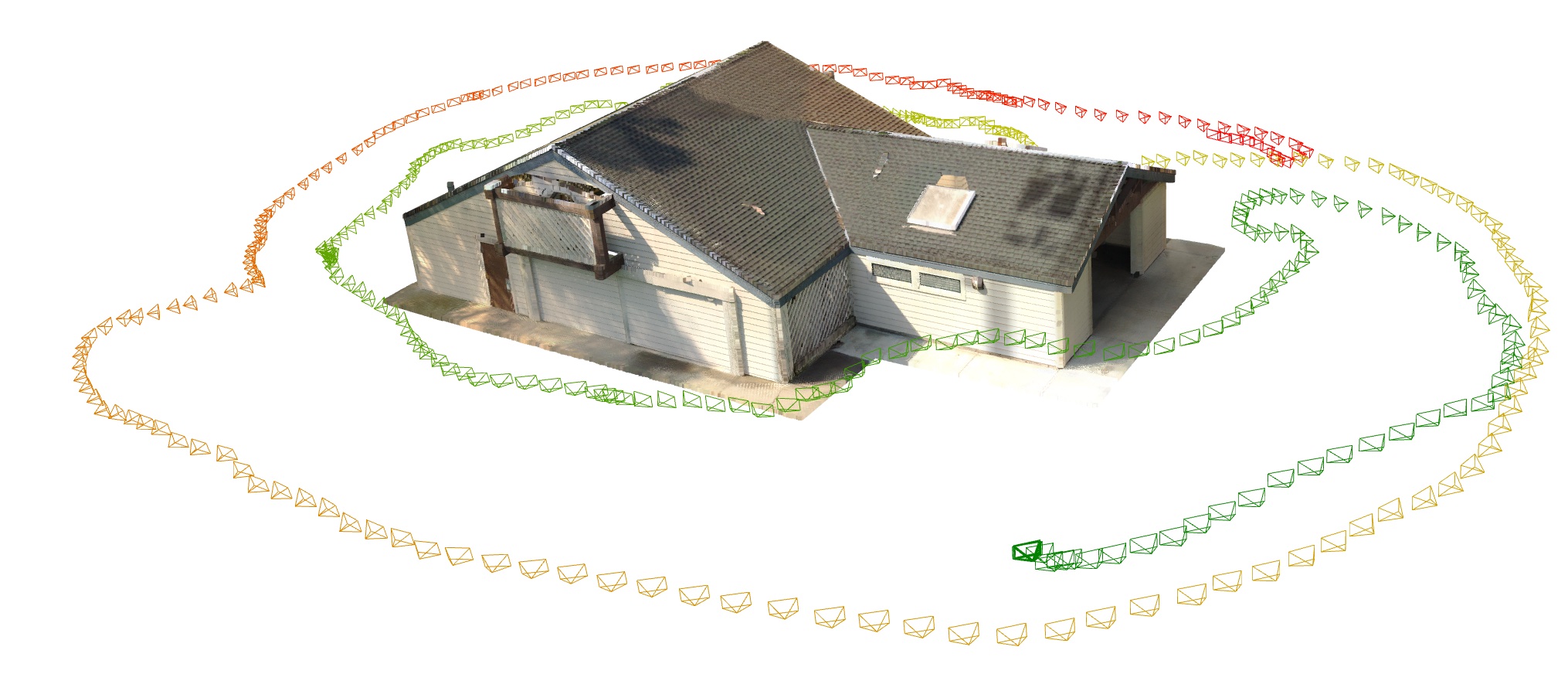}
        \caption{Ground truth laser scan data and COLMAP camera poses.}
    \end{subfigure}
    
    \vspace{10pt}
    
    \begin{subfigure}[b]{0.4\textwidth}
        \centering
        \includegraphics[width=\linewidth]{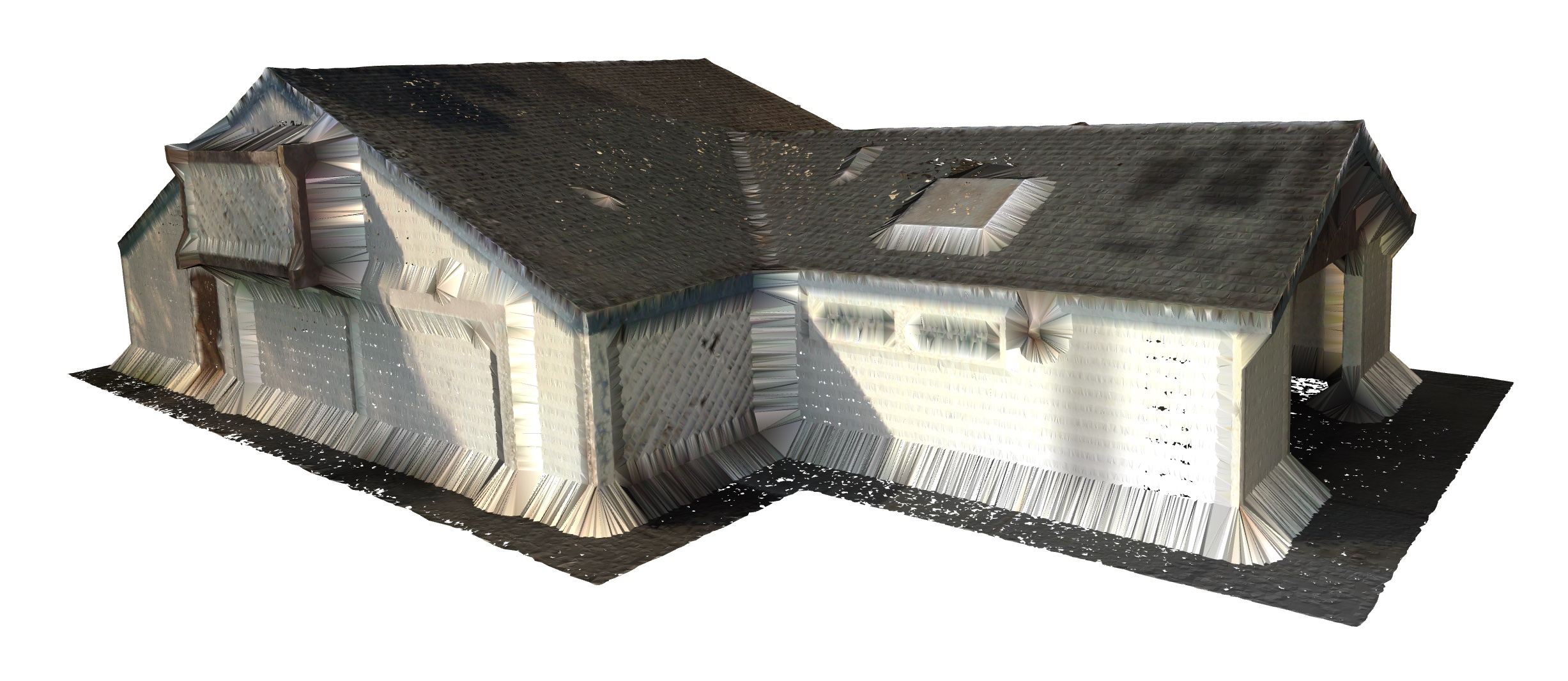}
        \caption{Reconstructed mesh surface.}
    \end{subfigure}
    
    \vspace{10pt}

    \begin{subfigure}[b]{0.4\textwidth}
        \centering
        \includegraphics[trim={0.5cm 0.5cm 0.5cm 0.5cm},clip,width=\linewidth]{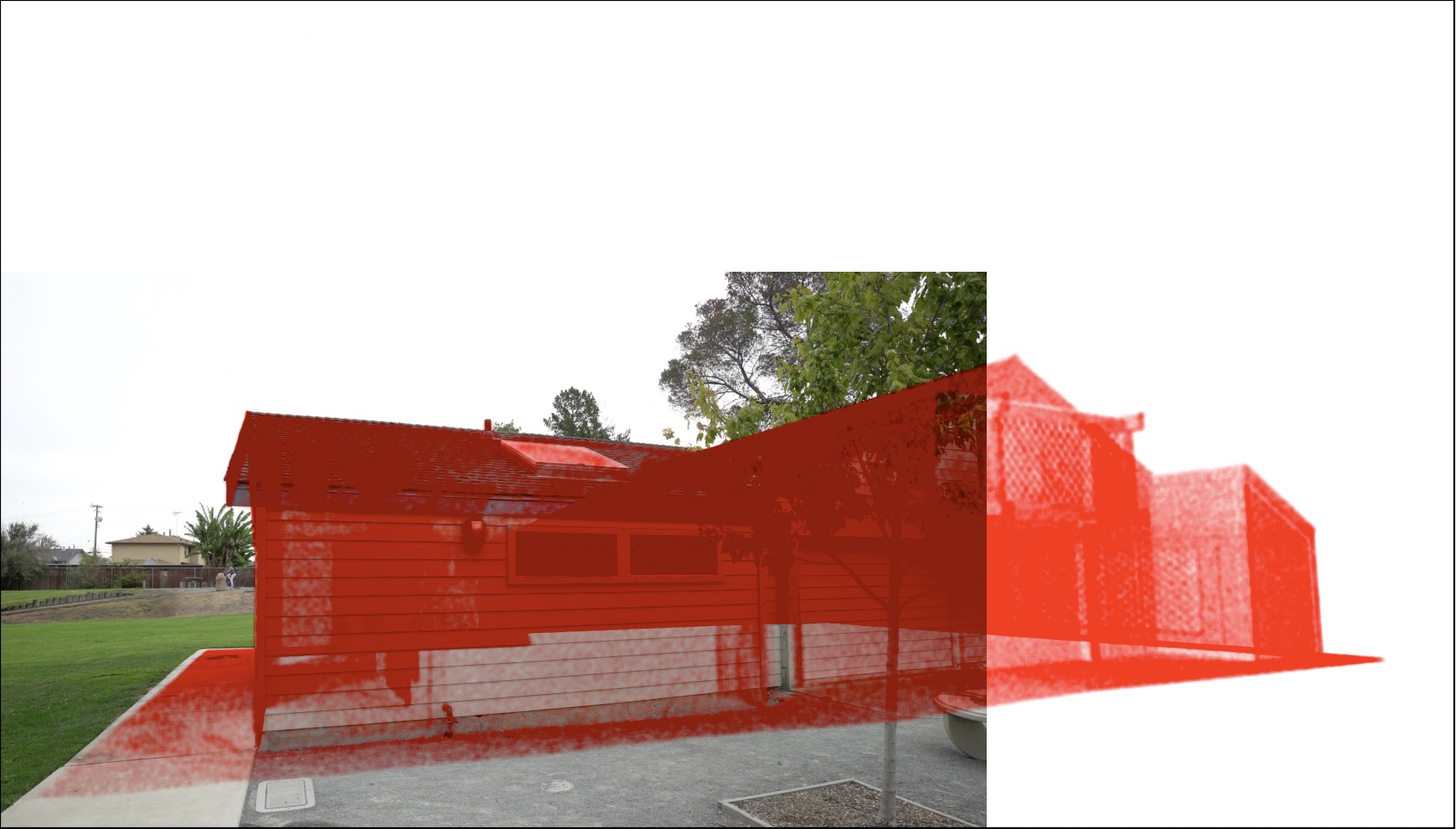}
        \caption{Projection of LiDAR scan points onto image.}
    \end{subfigure}
    
    \vspace{10pt}
    
    \begin{subfigure}[b]{0.33\textwidth}
        \centering
        \includegraphics[width=\linewidth]{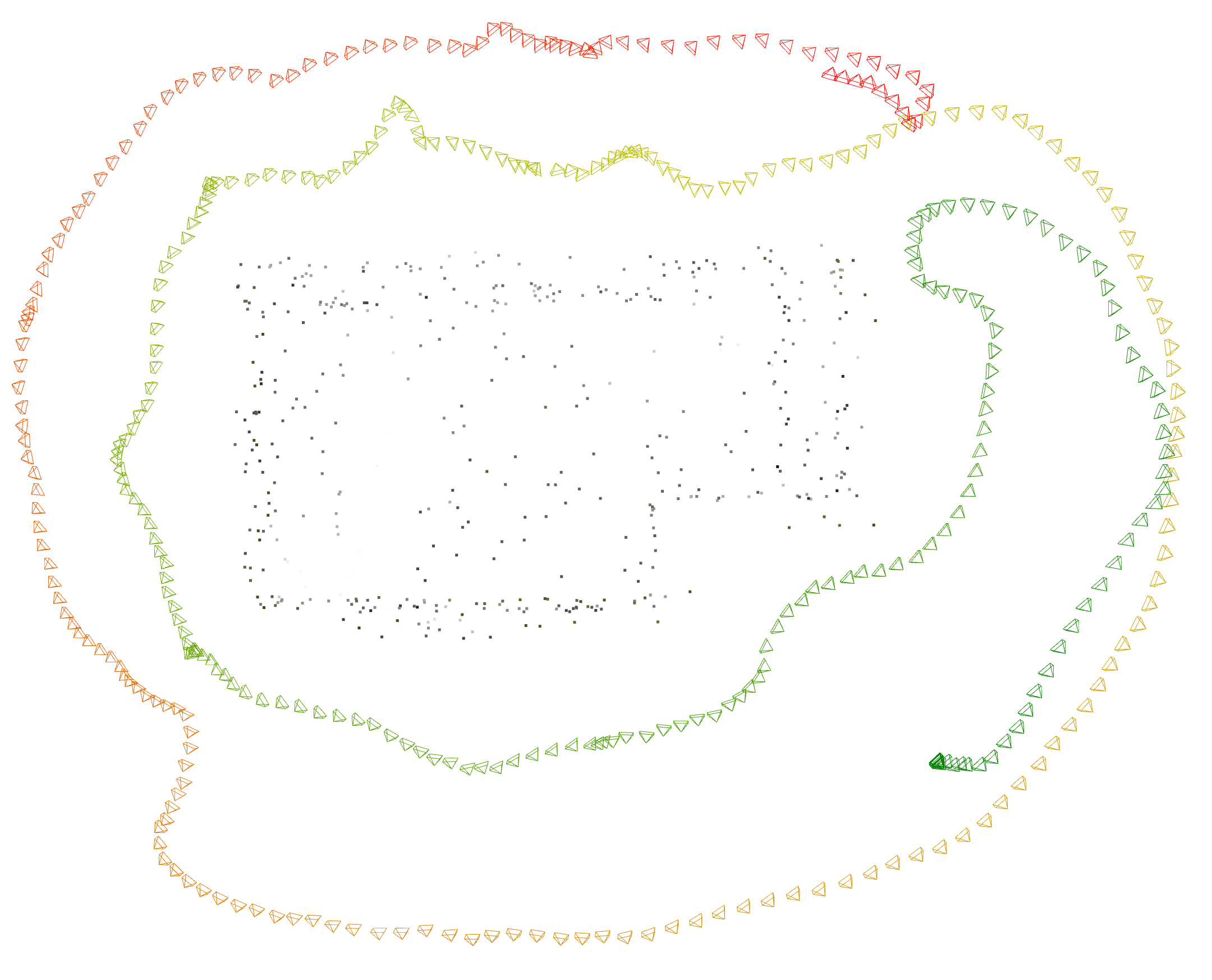}
        \caption{GTSfM reconstruction from selected virtual correspondences}
    \end{subfigure}
    
    \vspace{10pt}
    
    \begin{subfigure}[b]{0.33\textwidth}
        \centering
        \includegraphics[width=\linewidth]{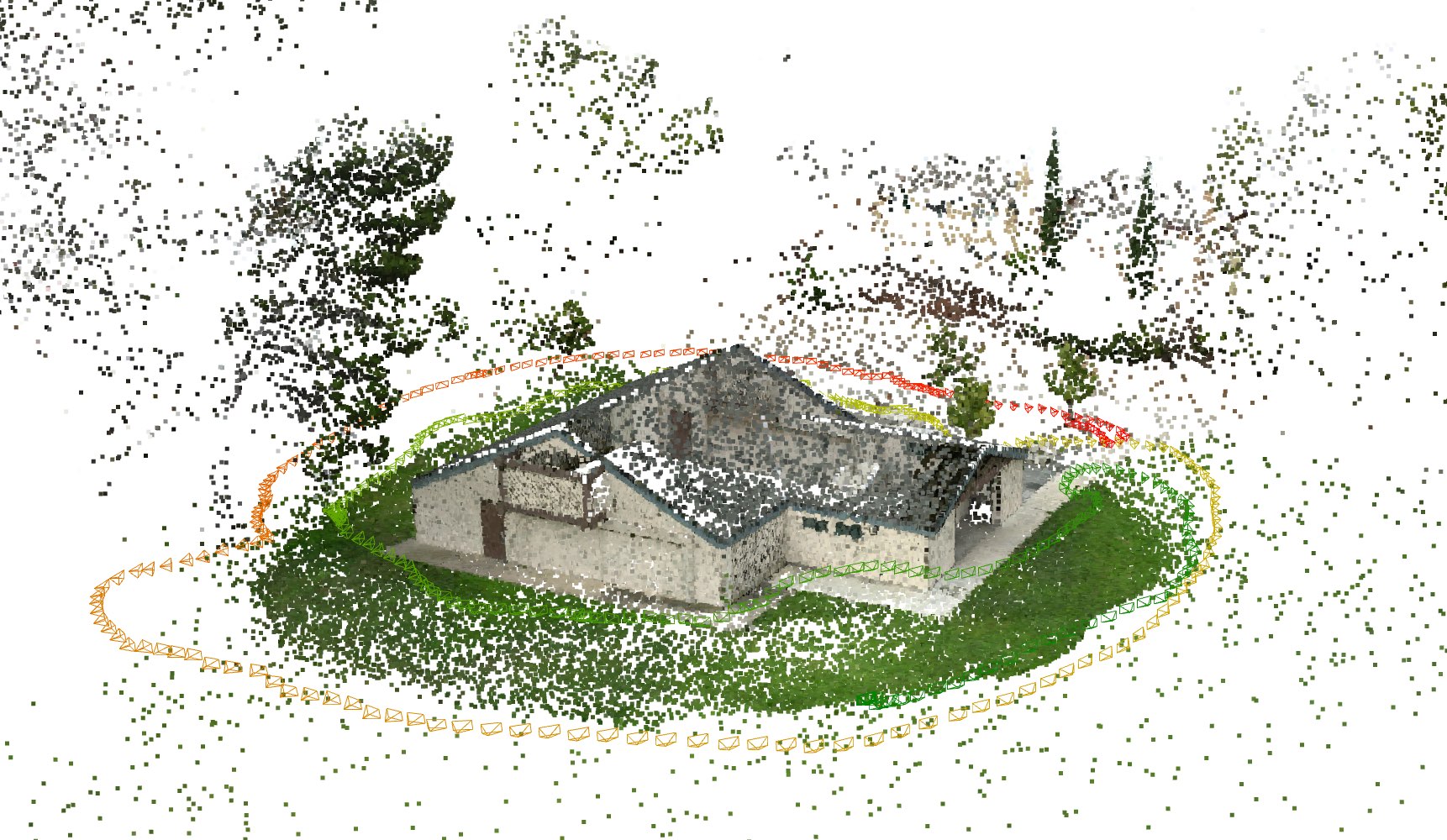}
        \caption{COLMAP `ground truth' reconstruction.}
    \end{subfigure}
    
    \vspace{10pt}
    \caption{Visualizations of inputs, outputs, and ground truth data for a `virtual' front-end on the Tanks and Temples \textit{Barn} sequence.}
    \vspace{1cm}
    \label{fig:virtual-front-end}
\end{figure}

\subsection{Ablation using Virtual Correspondences}
We examine the performance of our GTSfM system in a setting where an oracle `virtual' front-end provides noise-free correspondences, instead of using feature detection and matching. In this setting, any inaccuracy would arise from algorithms and optimization in the `back-end'. Such an experiment requires a high-fidelity 3d ground truth surface mesh that is registered to the global frame where ground truth cameras poses reside, and such datasets are rare. We use the \textit{Barn} sequence from the Tanks and Temples dataset \cite{Knapitsch17acmtg_TanksTemples}. We generate noise-free correspondences by reconstructing a  3d surface mesh from laser scans using the alpha shape algorithm \cite{Edelsbrunner83transit_AlphaShape}, as implemented in the Open3D library \cite{Zhou18arxiv_Open3D} (see Figure \ref{fig:virtual-front-end}). The alpha shape is a generalization of a convex hull, and we set $\alpha = 0.1$. 


We randomly sample 2000 3d points from the mesh surface, and for each image pair, attempt to project each 3d point into each image to see if it falls within the camera field of view (see Figure \ref{fig:virtual-front-end} for an example). If so, we then cast a ray from the camera center back through the image `keypoint' to see if the 3d point is occluded by another face of the 3d mesh. If occluded, we discard the `keypoint'. While intrinsics are not provided in the ground truth, the Tanks and Temples image EXIF data does contain the focal length.

We observe that GTSfM can achieve near-perfect camera localization when provided these noiseless correspondences, achieving a Global Pose AUC of over 99.5 at each threshold (see Table \ref{tab:virtual-front-end}).

We find that we must modify the noise model, decreasing the uncertainty $\Sigma$ on two-view measurements in Shonan rotation averaging from 1.0 to 0.1.

\begin{table*}
    \centering
        \caption{Results with `virtual correspondences' (i.e. noise-free correspondences synthetically generated from an oracle matcher), at 760p resolution. `Median / Mean' statistics are given for 1d distributions.  Abbreviations: View Graph Estimation (VG), Rotation Averaging (RA), Translation Averaging (TA), Bundle Adjustment (BA)}
    \begin{adjustbox}{width=\linewidth}
    \vspace{-10mm}
\begingroup
\begin{tabular}{lllllllllllllllllllllllll}
\toprule
\rowcolorize Dataset    & Front-End & \# Input & \# Retrieved & Front-End      & Front-End      & Front-End      & VG           & VG           & VG             & VG inlier      & VG inlier      & RA             & TA             & \# Reg.      & BA           & BA           & BA             & BA             & BA             & BA           & BA           & BA           & BA           & BA           \\
\rowcolorize           &           & Images   & Image        & Rot. angular   & Trans.         & Pose           & \# input     & \# inlier    & \# outlier     & Rot            & trans.         & rot.           & trans.         & Cameras      & \# Tracks    & track        & reproj         & rot.           & trans.         & Pose         & Pose         & Pose         & Pose         & Pose         \\
\rowcolorize           &           &          & Pairs        & errors         & Angular        & Errors         & edges        & edges        & edges          & angular        & angular        & angular        & angle          &              & (filtered)   & lengths      & errors         & angle          & angle          & AUC          & AUC          & AUC          & AUC          & AUC          \\
\rowcolorize           &           &          &              & (deg.)         & errors (deg).  & (deg.)         &              &              &                & errors (deg.)  & errors (deg.)  & error (deg.)   & error (deg.)   &              &              & (filtered)   & filtered (px)  & error (deg)    & error (deg.)   & @1 deg.      & @2.5 deg.    & @5 deg.      & @10 deg.     & @20 deg.     \\
\rowcolorize           &           &          &              & ($\downarrow$) & ($\downarrow$) & ($\downarrow$) & ($\uparrow$) & ($\uparrow$) & ($\downarrow$) & ($\downarrow$) & ($\downarrow$) & ($\downarrow$) & ($\downarrow$) & ($\uparrow$) & ($\uparrow$) & ($\uparrow$) & ($\downarrow$) & ($\downarrow$) & ($\downarrow$) & ($\uparrow$) & ($\uparrow$) & ($\uparrow$) & ($\uparrow$) & ($\uparrow$) \\
\midrule
Tanks \& Temples, Barn                                    & Virtual & 410  & 1247  & 0 / 0 & 0 / 0 & 0 / 0 &  1247 & 1234 &  13 &	0 / 0 & 0 / 0	& 0.9 / 1.1 &  0.9 / 1.0 & 408 & 	1613 & 14.0 / 21.8 & 0 / 0 & 	0 / 0 & 	0 / 0 & 	99.5 &	99.5 & 	99.5 & 	99.5 & 	99.5 \\
\bottomrule
\end{tabular}
\endgroup
\end{adjustbox}
\label{tab:virtual-front-end}
\end{table*}





\end{document}